\documentclass[11pt]{article}

\usepackage[final]{acl}

\usepackage{times}
\usepackage{latexsym}

\usepackage[T1]{fontenc}
\usepackage[utf8]{inputenc}

\usepackage{microtype}

\usepackage{inconsolata}

\usepackage{graphicx}

\usepackage{url}

\usepackage{booktabs}
\usepackage{multirow}

\usepackage{amsfonts}
\usepackage{amsmath}
\usepackage{nicefrac}

\usepackage{xcolor}

\usepackage{listings}

\usepackage{subcaption}

\usepackage{enumitem}

\usepackage{tcolorbox}
\tcbuselibrary{skins,breakable}

\newtcolorbox{examplebox}[1][]{
  enhanced, breakable,
  colback=gray!5, colframe=gray!55,
  fonttitle=\bfseries, title={#1},
  boxrule=0.4pt, arc=2pt,
  before skip=6pt, after skip=6pt,
  left=6pt, right=6pt, top=4pt, bottom=4pt,
}

\lstdefinestyle{prompt}{
  basicstyle=\ttfamily\footnotesize,
  breaklines=true, breakatwhitespace=true,
  breakindent=0pt, postbreak=\mbox{\textcolor{gray}{$\hookrightarrow$}\space},
  columns=fullflexible, keepspaces=true,
  frame=none, aboveskip=0pt, belowskip=0pt,
  showstringspaces=false,
}
\DeclareTColorBox{promptbox}{O{}}{
  enhanced, breakable,
  colback=gray!5, colframe=gray!50!black,
  coltitle=white, colbacktitle=gray!60!black,
  title=#1, fonttitle=\bfseries\small,
  left=2mm, right=2mm, top=1mm, bottom=1mm,
}

\title{MemFail: Stress-Testing Failure Modes of LLM Memory Systems}

\author{
  Ishir Garg \\
  University of California, Berkeley \\
  \texttt{ishirgarg@berkeley.edu} \\
  \And
  Neel Kolhe \\
  University of California, Berkeley \\
  \texttt{neelkolhe@berkeley.edu} \\
  \AND
  Dawn Song \\
  University of California, Berkeley \\
  \texttt{dawnsong@cs.berkeley.edu} \\
  \And
  Xuandong Zhao \\
  University of California, Berkeley \\
  \texttt{xuandongzhao@berkeley.edu} \\
}

\begin{document}

\makeatletter
\let\acl@orig@maketitle\@maketitle
\def\@maketitle{%
  \acl@orig@maketitle
  \vskip -0.15in
  \begin{center}
    \href{https://github.com/ishirgarg/MemFail}{\texttt{https://github.com/ishirgarg/MemFail}}
  \end{center}
  \vskip 0.1in
}
\makeatother

\maketitle

\begin{abstract}
Large language model (LLM) agents increasingly rely on external memory systems to remain consistent across long-horizon interactions, but little empirical work has been done to understand the specific failure modes and design choices that these systems present. Existing benchmarks report aggregate question-answering accuracy and treat memory systems as black boxes, making it impossible to attribute an incorrect answer to a particular failure mode of the system. We introduce \textsc{MemFail}, a diagnostic benchmark that isolates the failure modes of modern LLM memory systems. We begin by formalizing memory systems as the composition of three canonical operations---summarization, storage, and retrieval---and identify the potential failure modes induced by each. Based on these hypothesized failure modes, we construct five datasets spanning four tasks, each adversarially designed to test a specific operation of a memory system. Using these datasets, we evaluate four state-of-the-art memory systems on \textsc{MemFail} and demonstrate how \textsc{MemFail} can be used to empirically understand the tradeoffs induced by differences in memory system architectures.
\end{abstract}

\section{Introduction}
\label{sec:introduction}

LLM agents have rapidly grown more capable at long-horizon tasks, but their limited context windows make it difficult to maintain consistency over time. Traditional memory systems such as retrieval-augmented generation (RAG) \cite{lewisRetrievalAugmentedGenerationKnowledgeIntensive2021} and vector databases store the raw conversation history, which is expensive in storage and input tokens when only a fraction is worth remembering, and which provides no natural way to forget or update old information as user preferences change.

A growing body of work on \textit{LLM memory systems} has emerged in response, augmenting agents with external stores they can read, write, and update over a lifetime, enabling consistent and personalized responses \cite{chhikaraMem0BuildingProductionReady2025, xuAMemAgenticMemory2025a, liuSimpleMemEfficientLifelong2026, xuStructMemStructuredMemory2026, rasmussenZepTemporalKnowledge2025, huEverMemOSSelfOrganizingMemory2026}. But the very mechanisms for compression, updating, and forgetting introduce new failure modes: a system may \textit{over-summarize} a conversation before storing it, stripping key details, or fail to remove an old fact when a contradictory one arrives.

Prior work has converged on a few key desiderata: retrieving only the memories useful for the current task, faithful storage that preserves the semantics of the original experience, updatability to overwrite outdated facts and accumulate coexisting ones, low latency for fast storage and retrieval, and token efficiency to minimize tokens in the agent's context. However, these desiderata trade off: a system that stores every conversation verbatim (like simple RAG) excels at faithful storage but is token-inefficient, while breaking conversations into smaller memories shrinks retrieved context but makes the relevant pieces harder and slower to find.

Despite the recent surge of memory systems with varied design choices, little work has characterized the \textit{tradeoffs} between these desiderata or the failure modes underneath them. Existing benchmarks treat memory systems as end-to-end black boxes, providing aggregate metrics that obscure the distinct failure modes inside. In this work, we introduce \textsc{MemFail}, a benchmark targeting failure modes of LLM memory systems. We begin by proposing a framework in which memory systems can be formally represented as a combination of three operations: summary, storage, and retrieval. Based on this taxonomy, we hypothesize distinct failure modes that any memory system fitting this formalism may exhibit. \textsc{MemFail} provides five datasets spanning four tasks to elicit these failure modes:
\begin{itemize}  
    \item \textit{Conditional-Facts}: faithful retention of important causal relationships.
    \item \textit{Persona-Retrieval}: faithful retention of small details under misleading questions.
    \item \textit{Long-Hop}: retrieval over long-range causal relationships.
    \item \textit{Coexisting-Facts}: storing and retrieving \textit{all} relevant coexisting pieces of information for a query.
    \item \textit{Conditional-Facts (Hard)}: preserving important causal details when summarizing experiences.
\end{itemize}
We evaluate four state-of-the-art memory systems---Mem0 \cite{chhikaraMem0BuildingProductionReady2025}, A-MEM \cite{xuAMemAgenticMemory2025a}, SimpleMem \cite{liuSimpleMemEfficientLifelong2026}, and StructMem \cite{xuStructMemStructuredMemory2026}---on \textsc{MemFail}, and our analysis surfaces three findings that aggregate metrics would otherwise obscure. First, no single system dominates: each architecture exhibits a distinctive failure signature---graph-based StructMem excels at causal reasoning but collapses on coexisting-fact retrieval, while Mem0 shows the opposite pattern. Second, scaling either the number of retrieved memories or the strength of the underlying LLM yields little improvement, and in several cases degrades performance, indicating that current systems are bound by architectural constraints rather than by model intelligence or context budget. Third, the relationship between token consumption and accuracy is task-dependent: summary-bottlenecked tasks reward verbose memories, while retrieval-bottlenecked tasks suffer when large memories pollute the embedding space. Building on these findings, we propose two under-explored directions---mixture-of-experts memory architectures and task-adaptive token scaling---as paths toward systems that mitigate, rather than trade off between, the desiderata above.

\section{Related Work}
\label{sec:related}

\textbf{LLM Memory Systems.} A growing body of work augments LLMs with persistent memory. MemGPT pages between in-context working memory and external storage in an OS-inspired hierarchy \cite{packerMemGPTLLMsOperating2024}. MemoryBank applies summarization and Ebbinghaus-style forgetting curves over user-specific facts \cite{zhongMemoryBankEnhancingLarge2023}. Mem0 extracts atomic facts into a vector store (or, in its graph-augmented variant Mem0g, a hybrid vector-and-graph store) with explicit \texttt{ADD}, \texttt{UPDATE}, and \texttt{DELETE} operations \cite{chhikaraMem0BuildingProductionReady2025}. A-MEM forgoes predefined schemas, organizing memories as agentically linked notes that the agent dynamically tags and reorganizes \cite{xuAMemAgenticMemory2025a}. StructMem builds hierarchical event-level structures with periodic semantic consolidation \cite{xuStructMemStructuredMemory2026}, while SimpleMem applies semantically lossless compression via entropy-aware filtering and adaptive retrieval \cite{liuSimpleMemEfficientLifelong2026}. Reflexion \cite{shinnReflexionLanguageAgents2023} and Generative Agents \cite{parkGenerativeAgentsInteractive2023} explore reflective memory streams for self-improvement and social simulation. These systems share the primitives our framework abstracts---summarize, store, retrieve---and our benchmark plugs into any system exposing them.

\textbf{LLM Long-Context and Memory System Benchmarks.} Two orthogonal lines of work complement ours. ER-MIA \cite{piehlERMIABlackBoxAdversarial2026} and InjecMEM \cite{tianInjecMEMMemoryInjection2025} mount memory-injection attacks against memory-augmented LLMs and layered systems such as MemoryOS \cite{kangMemoryOSAI2025}, while SkillJect \cite{jiaSkillJectAutomatingStealthy2026} mounts a skill-based prompt-injection attack against the agent-skills abstraction, exposing adversarial rather than benign failures. Needle-in-a-Haystack \cite{nelsonNeedleHaystackMemory2024}, RULER \cite{hsiehRULERWhatsReal2024}, LongEval \cite{liLongEvalOpenSourceContextLength2023}, LongBench \cite{baiLongBenchBilingualMultitask2024}, L-Eval \cite{anLEvalInstitutingStandardized2024}, and M$^4$LE \cite{kwanM4LEMultiAbilityMultiRange2024} stress-test the long-context capabilities of LLMs themselves rather than the memory systems built atop them. For general recall, MemoryBank pairs probing questions with multi-day histories \cite{zhongMemoryBankEnhancingLarge2023}; PerLTQA targets personalized long-term QA \cite{duPerLTQAPersonalLongTerm2024}; MemBench spans factual and reflective memory across participation and observation scenarios with metrics for accuracy, recall, capacity, and temporal efficiency \cite{tanMemBenchMoreComprehensive2025}; MemoryAgentBench converts long-context datasets into incremental tasks across four competencies---accurate retrieval, test-time learning, long-range understanding, and selective forgetting \cite{huEvaluatingMemoryLLM2025}; LoCoMo provides multi-session conversations covering single-hop, multi-hop, temporal, open-domain, adversarial, event-summarization, and multi-modal tasks \cite{maharanaEvaluatingVeryLongTerm2024}; LongDialQA derives multi-party dialogues from TV scripts \cite{kimDialSimDialogueSimulator2025}. For reasoning, MemSim \cite{zhangMemSimBayesianSimulator2024} synthesizes QA via a Bayesian Relation Network, and MemoryBench reframes evaluation as continual learning from simulated feedback, distinguishing declarative from procedural memory \cite{aiMemoryBenchBenchmarkMemory2026}. A few target failures explicitly: LoCoMo includes misleading questions for abstention \cite{maharanaEvaluatingVeryLongTerm2024}; LongMemEval curates 500 questions across information extraction, multi-session reasoning, temporal reasoning, knowledge updates, and abstention, alongside a key-value formal model \cite{wuLongMemEvalBenchmarkingChat2024}. However, their model is less general than ours, and they do not discuss failure modes; StructMemEval shows that organizational hints improve memory organization \cite{shutovaEvaluatingMemoryStructure2026}. All treat the system as a black box, reporting aggregate end-to-end scores that cannot localize failures to summarization, storage, retrieval, or reasoning---a gap \textsc{MemFail} fills by isolating each operation and analyzing failure modes and their tradeoffs.

\section{Background}
\label{sec:framework}

We present a formal model of LLM memory systems that \textsc{MemFail} assumes. The framework identifies natural failure modes any system fitting it may exhibit, which \textsc{MemFail} is designed to test.

\subsection{Three Operations of a Memory System}
We assume every memory system decomposes into three operations. Let $Q$ denote a user query, $H$ a conversation history, and $M$ the current memory database.

\textbf{Retrieval.} Given a new query $Q$, the conversation history $H$, and the memory state $M$, the system retrieves a set of relevant memories $R$ drawn from $M$ and inserted into the agent's prompt alongside $Q$ and (optionally) $H$.

\textbf{Summarization.} After an interaction, $H$ is compressed into a representation $H'$ that extracts the information deemed worth retaining.

\textbf{Storage.} The storage step takes $H'$ and the existing memory state $M$ and produces an updated memory state $M'$, possibly overwriting, merging, or appending to existing entries in $M$, or performing a no-op.

As demonstrated in Table \ref{tab:system-implementations}, this decomposition subsumes a wide range of modern memory systems, including Mem0, A-MEM, SimpleMem, and StructMem, and mirrors the canonical encoding/storage/retrieval description of human memory.

\begin{table*}[h]
\centering
\small
\setlength{\tabcolsep}{4pt}
\resizebox{\linewidth}{!}{%
\begin{tabular}{@{}llll@{}}
\toprule
System & \textsc{Summarize} & \textsc{Store} & \textsc{Retrieve} \\
\midrule
SimpleMem & verbatim & flat list of turns & embedding top-$k$ \\
Mem0 & LLM fact extraction & atomic units; LLM tool-call updates (add/update/delete) & embedding top-$k$ over units \\
A-MEM & LLM-written descriptive notes & linked notes in vector DB & embedding top-$k$ \\
StructMem & LLM entity/relation extraction & knowledge graph (typed nodes + edges) & subgraph around query entities \\
\bottomrule
\end{tabular}
}
\vspace{0.5em}
\caption{How each tested memory system implements the three operations of Section \ref{sec:framework}.}
\label{tab:system-implementations}
\end{table*}

\subsection{Failure Modes}
\label{sec:failure-modes}
Based on our abstract model of a memory system, we propose that memory systems exhibit four natural failure modes:
\begin{itemize}  
    \item \textbf{Summary failure:} summarization deletes or malforms important information present in $H$. E.g., \emph{``I am deathly allergic to peanuts''} may be compressed into \emph{``allergic to peanuts''}, stripping the severity critical for downstream reasoning.
    \item \textbf{Storage failure:} the storage step does not adequately incorporate $H'$ into $M$. This includes refusing to overwrite outdated facts (e.g., not updating \emph{``Dan likes pizza''} after the user states Dan now hates pizza) and refusing to admit valid coexisting facts (e.g., rejecting \emph{``Dan likes burgers''} as contradictory to stored \emph{``Dan likes pizza''}).
    \item \textbf{Retrieval failure:} retrieval fails to return relevant memories in $R$, or returns memories that are semantically similar but contextually inappropriate.
    \item \textbf{Reasoning failure:} the agent makes an incorrect judgment even when the correct memories were retrieved. Note that this is \textit{not} a failure of the memory system, but we measure it for completeness.
\end{itemize}
Prior work primarily evaluates the combination of all four modes by inserting long conversation histories and asking LLMs to infer user personalities or preferences without distinguishing between them. \textbf{\textsc{MemFail} consists of tasks specifically designed to isolate failure modes (1)--(3)}, which are unique to modern memory systems and unaddressed by prior benchmarks.

\subsection{Compatibility with Existing Memory Systems}
\label{sec:api}
Critically, a good benchmark should evaluate a diverse array of memory systems with minimal assumptions about their internals. \textsc{MemFail} evaluates any system that exposes three functions:
\begin{itemize} 
    \item \texttt{store\_conversation($H$)}: store $H$ in the database (subsuming \textsc{Summarize} and \textsc{Store}).
    \item \texttt{retrieve\_memories($Q, H, k$)}: return the top-$k$ memories relevant to $Q$ and $H$.
    \item \texttt{get\_all\_memories()}: return all currently stored memories.
\end{itemize}
The first two are required for the evaluation loop. The third is used only by \textsc{MemFail}'s judge LLM to diagnose which failure mode occurred, and is not required for deployment of the memory system. Any system implementing these three functions automatically plugs into our evaluation harness.

% =====================================================================
% Section 4: Benchmark Details (compressed)
% =====================================================================

\section{Benchmark Details}
\label{sec:benchmark}

\textsc{MemFail} comprises of five datasets in English organized into four tasks, each hypothesized to isolate one failure mode from Section~\ref{sec:framework}.
We sketch the high-level design only; full specifications, generator
prompts, validation rules, deduplication thresholds, sampling pools,
and additional worked examples are deferred to
Appendix~\ref{app:construction}.

\paragraph{Task 1: \textit{Conditional-Facts}.}
\label{sec:conditional-facts}
Targets \emph{summary failure}---summarization-based memory stripping
qualifying conditions from facts at commit time. Each row encodes a rule
``entity $E$ exhibits behavior $B$ only when condition $C$ is satisfied,''
embedded in a $5$--$8$ sentence essay alongside $4$--$7$ unrelated
unconditional facts about the same entity. The graded query asks whether $E$
would exhibit $B$ in a specific context $X$ that either does or does not
satisfy $C$; a system that drops $C$ during summarization and stores the
unconditional ``$E$ does $B$'' would incorrectly answer ``yes'' regardless of $X$. Conditions are sampled from a fixed list of types (time of day, weather, mood, social setting, prior activity, etc.).
We create two variants: the \textbf{Easy} variant places the entire rule inside a single sentence, so
any system that copies the sentence verbatim succeeds; the \textbf{Hard} variant decomposes the rule into three non-adjacent sentences---a behavior
sentence, a condition sentence, and a linking sentence---spread across an $8$--$12$ sentence essay, forcing
\emph{reconstruction} from distributed evidence. Easy and Hard rows share the same entity and condition specs, so any performance gap is attributable to the distribution of the rule across sentences. A Hard example is given in Appendix~\ref{app:hard-decomposition}.

\begin{examplebox}[Example: \textit{Conditional-Facts} (Easy)]
\textbf{Essay (excerpt):} ``\ldots Sylas is a shrewd negotiator who thrives
in the bustling markets and political halls of his city.\ldots\ Sylas draws
elaborate maps only if he has just finished a negotiation.\ldots''\quad \\
\textbf{Question:} ``Sylas just finished meditating quietly, not negotiating.
Would he draw an elaborate map now?''\quad \\
\textbf{Ground truth:} ``No---Sylas only draws maps immediately after a negotiation.''
\end{examplebox}

\paragraph{Task 2: \textit{Coexisting-Facts}.}
\label{sec:coexisting-facts}
Targets both \emph{storage failure} and \emph{retrieval failure}. Modern
memory systems aggressively reconcile incoming information with their
existing database; a possible failure pattern is that two \emph{compatible}
facts (``user likes pizza'' and ``user likes ramen'') are incorrectly treated
as contradictions, causing the system to overwrite the older fact rather than
store both. Each row covers one of $100$ curated preference categories
(foods, hat styles, music genres, etc.), holds $N\!\in\!\{2,3,4,5\}$ distinct
preferences within that category---each expressed as an isolated
first-person statement that references only that single preference---and asks
a holistic scenario question whose well-formed answer requires all $N$
preferences.

\begin{examplebox}[Example: \textit{Coexisting-Facts}]
\textbf{Category:} hat styles. \\ \textbf{Preference facts:}
(i) ``I often wear a fedora when I want to add a classic touch to my
outfit''; (ii) ``Beanies are my go-to for staying warm and casual during
chilly days''; (iii) ``A bucket hat is what I reach for on sunny, laid-back
weekends.'' \\ \textbf{Question:} ``I'm packing for a week-long trip with
mixed weather---which hats should I bring to cover all occasions?'' \\
\textbf{Ground truth:} fedora, beanie, bucket hat.
\end{examplebox}

\paragraph{Task 3: \textit{Persona-Retrieval}.}
\label{sec:personal-retrieval}
Targets \emph{storage failure}---a memory system retrieving a stored profile
when asked about a different person. Each row contains a $10$--$15$ sentence
essay about a named entity $E$ embedding $4$--$5$ idiosyncratic facts. Three
graded queries are attached, each independently toggled $50/50$ between two
forms: a \emph{direct} query that names $E$ and is answerable from a specific
essay detail, and a \emph{misleading} query that names a distractor $D$
unrelated to the question, for which the correct answer is to abstain.
Entities and distractors are sampled from a fixed pool of $30$ diverse names;
persona flavors from a fixed pool of $30$ flavors.

\begin{examplebox}[Example: \textit{Persona-Retrieval}]
\textbf{Essay (excerpt):} ``Yuki Tanaka
spends most mornings hunched over a lightbox, tracing contours\ldots\ She refuses to eat shellfish because she gets a severe hive reaction, so boat provisions are always cooked onshore and
strictly shellfish-free.\ldots''\\
\textbf{Q1 (misleading):} ``Does Noah Brooks eat shellfish?''
$\rightarrow$ ``I don't have information about Noah Brooks.''\\
\textbf{Q2 (non-misleading):} ``What food should I avoid serving to Yuki
Tanaka?'' $\rightarrow$ ``Avoid shellfish---she has a severe allergic
reaction.''
\end{examplebox}

\paragraph{Task 4: \textit{Long-Hop}.}
\label{sec:long-hop}
Targets \emph{retrieval failure}. Each row encodes a strictly transitive
chain $A_1 \!\to\! \cdots \!\to\! A_{K+2}$ of $K\!\in\!\{1,2,3\}$ hops, in
which fact $i$ links anchor $i$ to anchor $i{+}1$; anchors are subjective
(moods, routines, opinions, personal objects) so that no fact is answerable from
world knowledge alone, and every fact is self-contained. The graded question
references only the head anchor and asks for the terminal anchor, presented
as a $5$-way multiple-choice question with four shape-matched distractors
orthogonal to every fact in the chain. At evaluation time, we provide each fact
\textit{separately} to the memory system; this forces the system to retrieve and
compose the chain from scattered storage rather than read it off from a single
conversation.

\begin{examplebox}[Example: \textit{Long-Hop} ($K{=}3$)]
\textbf{Facts:} ``Whenever I sip
morning espresso I call my mom''; ``After I call my mom I plan a day trip'';
``When I plan a day trip I pack snacks''; ``When I pack snacks I take
scenic photos.''\\
\textbf{Question:} ``When I sip morning espresso, what
do I end up doing?''\\
\textbf{Choices:} A.~organize the bookshelf;\,
B.~practice guitar scales;\, C.~water the plants;\, \textbf{D.~take scenic
photos};\, E.~fold the laundry.\\\textbf{Ground truth:} take scenic
photos.
\end{examplebox}

\paragraph{Dataset statistics.}
\label{sec:dataset-stats}
Table~\ref{tab:benchmark-stats} reports measured statistics across all five
datasets, with all token counts (mean tokens per dataset entry) computed
using the \texttt{cl100k\_base} tokenizer~\cite{tiktoken}.
\textit{Coexisting-Facts} is broken out by per-row preference count $N$ and
\textit{Long-Hop} by hop count $K$, since both the number of storage units
and the entry-token cost scale with these.
\begin{table*}[htbp]
\centering
\small
\setlength{\tabcolsep}{4pt}
\resizebox{0.95\linewidth}{!}{
\begin{tabular}{llrrrl}
\toprule
\textbf{Dataset} & \textbf{Generator} & \textbf{\# Q} & \textbf{\# Graded} & \textbf{Avg.\ tok.} & \textbf{Other} \\
\midrule
\textit{Conditional-Facts} (Easy) & \texttt{gpt-4.1-mini} & $100$ & $100$ & $130$ & $32$y / $68$n; $27$ cond.\,types \\
\textit{Conditional-Facts} (Hard) & \texttt{gpt-5-mini}   & $100$ & $100$ & $189$ & $49$y / $51$n; $27$ cond.\,types \\
\textit{Coexisting-Facts} ($N{=}2/3/4/5$) 
& \texttt{gpt-4.1-mini} 
& $26/31/20/23$ 
& $26/31/20/23$ 
& $54/68/84/101$ 
& $52/93/80/115$ facts \\
\textit{Persona-Retrieval}        & \texttt{gpt-5-mini}   & $100$ & $300$ & $297$ & $52.3$\% misleading \\
\textit{Long-Hop} ($K{=}1/2/3$)   
& \texttt{gpt-5} 
& $31/32/29$ 
& $31/32/29$ 
& $59/74/87$ 
& $62/96/116$ facts \\
\bottomrule
\end{tabular}}
\caption{Per-dataset measured statistics; ``Avg.\ tok.'' is mean tokens per
entry under \texttt{cl100k\_base}. For \textit{Coexisting-Facts} and
\textit{Long-Hop}, values are reported in order of $N$ and $K$ respectively.}
\label{tab:benchmark-stats}
\end{table*}

\section{Experimental Setup}
\label{sec:setup}

\subsection{Evaluation Loop}
\label{sec:eval-loop}

\textsc{MemFail}'s harness applies to any memory system implementing the three operations of Section~\ref{sec:framework}. Each task reduces to a sequence of conversations: ungraded \emph{storage conversations} commit information to memory, and \emph{query conversations} elicit graded answers.

\textbf{Phase 1: Storage.} For each task we extract the information unit needed per graded query---a conditional-fact essay, a preference statement, a persona essay, or a message from the reasoning chain---and send each in its own conversation, so that the system must store and associate them across sessions.

\textbf{Phase 2: Query.} The harness creates one query conversation per graded question, calls \texttt{memory\_system.retrieve\_memories(query, conversation, k)}, formats the top-$k$ memories into the prompt, invokes the test-taker, and records the response. Query conversations do not update the database, so query order is irrelevant.

\textbf{Phase 3: Grading.} Each graded question has a set of $N \ge 1$ memories the system should retrieve. An LLM-as-a-judge receives the query, ground truth, all stored memories at query time, and all retrieved memories, and classifies each of the $N$ memories into:
\begin{enumerate}
\item \textbf{Storage check.} Is the memory present in \texttt{get\_all\_memories()}? Failure is a \textbf{storage error}.
\item \textbf{Summary check.} Conditional on storage, are \emph{critical details} preserved (e.g., the qualifying condition in \textit{Conditional-Facts})? Failure is a \textbf{summary error}.
\item \textbf{Retrieval check.} Conditional on faithful storage, was the entry in the top-$k$ set? Failure is a \textbf{retrieval error}.
\item \textbf{Reasoning check.} Conditional on retrieval, did the test-taker use it to produce the correct answer? Failure is a \textbf{reasoning error}.
\item \textbf{Correct.} All memories stored, summarized, retrieved, and used successfully.
\end{enumerate}

\textbf{Separating storage from summarization.} Because most open-source
systems expose summarization and storage as a single atomic
\texttt{store\_conversation} call, the two must be separated at grading time.
The storage check is deliberately permissive: it asks only whether \emph{any}
trace of the target information appears in the store, and the judge is
instructed to accept paraphrase, aggressive compression, distribution across
entries, and even attachment to the wrong entity
(Appendix~\ref{app:prompts}); a subsequent loss of critical detail is then
charged to summarization. One case is not separable: a system that malforms a fact and
then writes nothing at all is bucketed as a storage error by convention, since
with only the final memory state observable the omission cannot be attributed
to either operation without instrumenting system internals.

We focus our analysis on the first three, as these are memory-system failures; reasoning errors reflect LLM limitations and occur infrequently (Appendix~\ref{app:detailed-results}). We fix \texttt{gpt-5-mini} \cite{singhOpenAIGPT5System2026} as both test-taker and grader---the LLM that the memory system augments, not the system's internal model---so that cross-system differences are attributable to the memory system rather than the underlying LLM. Full prompts and grading details are given in Appendix~\ref{app:prompts}.

\textbf{Human Validation.} All dataset examples are manually verified for correctness. On 100 manually graded examples, \texttt{gpt-5-mini} answers 98\% correctly and classifies the error type 98.4\% correctly. Table~\ref{tab:judge-confusion} (Appendix~\ref{app:judge-validation}) gives the per-class confusion matrix against human labels over the $364$ memory-level classifications these examples induce; every disagreement is a storage/summary or summary/retrieval swap.

\subsection{Memory Systems}
\label{sec:memory-systems}

We evaluate four open-source modern memory systems---\textbf{Mem0} \cite{chhikaraMem0BuildingProductionReady2025}, \textbf{A-MEM} \cite{xuAMemAgenticMemory2025a}, \textbf{SimpleMem} \cite{liuSimpleMemEfficientLifelong2026}, and \textbf{StructMem} \cite{xuStructMemStructuredMemory2026}---and study in Section~\ref{sec:experiments} how the success rate and error types vary with architecture, retrieval depth $k$, and the strength of the system's internal model.

\textbf{No-compression baseline.} To separate error intrinsic to memory compression from error specific to a system's design, we additionally evaluate \textbf{Verbatim-RAG}, which performs no summarization: it stores each conversation verbatim as a single embedded entry and retrieves the top-$k$ entries by cosine similarity. By construction it cannot commit storage or summary errors.

\section{Experiments}
\label{sec:experiments}

We evaluate open-source memory systems on \textsc{MemFail}. The main text shows representative examples of the most interesting findings; more complete evaluation results are in Appendix \ref{app:detailed-results}.

\begin{figure*}[!htbp]
    \centering
    \includegraphics[width=0.75\linewidth]{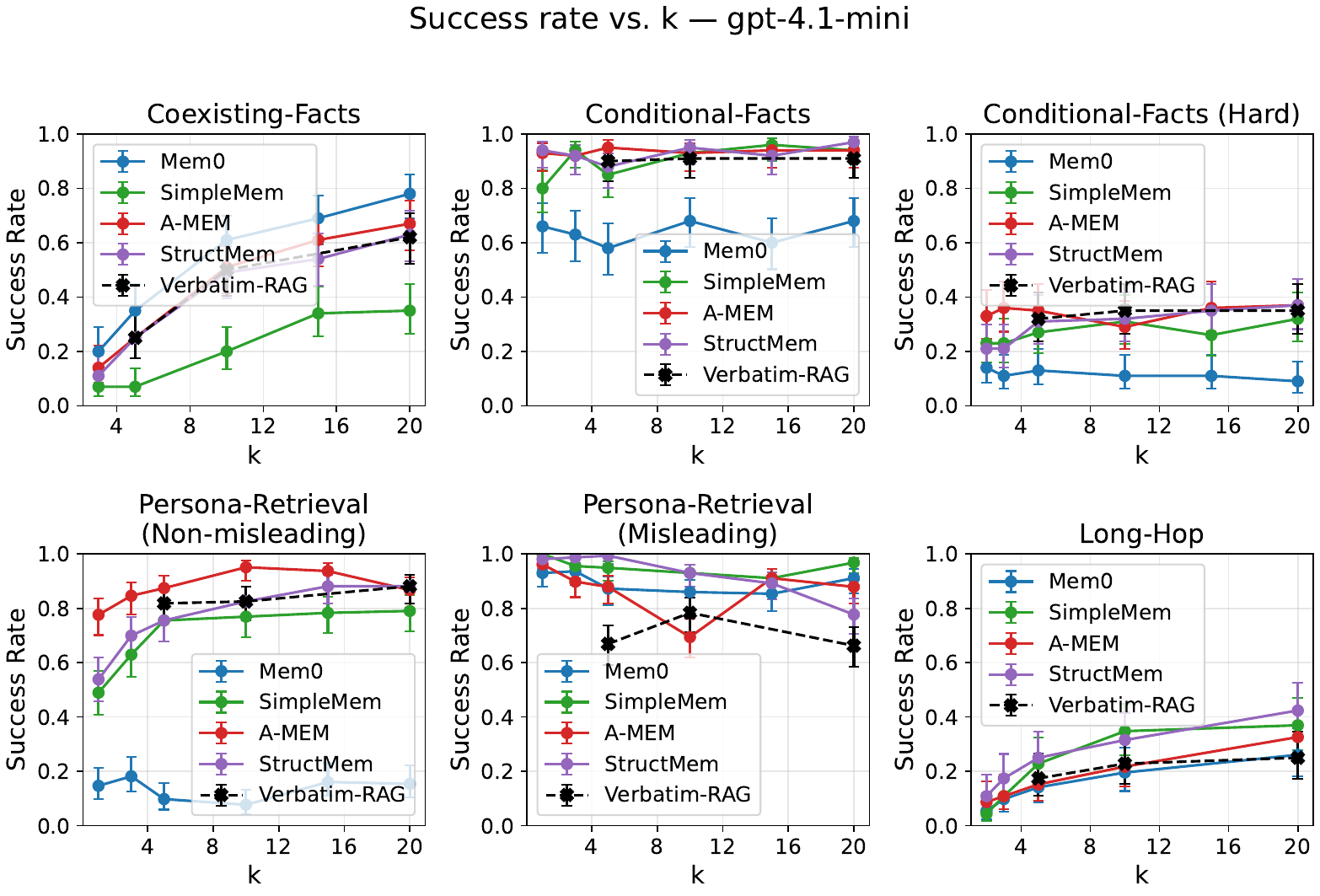}
    \caption{Performance of memory systems using GPT-4.1-mini internally, with the no-compression \textbf{Verbatim-RAG} baseline (black dashed) for reference; having no internal LLM, it is independent of that axis. All confidence intervals use 95\% Wilson score binomial intervals.}
    \label{fig:gpt-4.1-mini-perf}
\end{figure*}

\textbf{Q1: How does performance scale with $k$, the number of retrieved memories?}
Figure~\ref{fig:gpt-4.1-mini-perf} shows that \textbf{\textsc{MemFail} is
difficult even for state-of-the-art memory systems}, and that performance scales
poorly with $k$. The exception is \textit{Coexisting-Facts}, which naturally
benefits from larger $k$ since coexisting facts are more likely to be
retrieved, even by chance. Performance for a given system also varies
substantially across tasks: because \textsc{MemFail}'s tasks isolate specific
weaknesses, they reveal how architectures induce distinct failure
modes. StructMem performs strongly on most tasks but fails spectacularly on
\textit{Coexisting-Facts}, while Mem0 shows the opposite pattern (analyzed
in Q5). Figure~\ref{fig:app-gpt41-err} in Appendix~\ref{app:detailed-results} breaks down error
types per system, revealing how different tasks elicit different failure
modes (concrete failure examples in Appendix~\ref{app:failure-examples}):
\begin{itemize}
\item \textit{Coexisting-Facts} induces \textbf{retrieval failures}: most
systems fail to associate all related facts with the query at retrieval time.
\item \textit{Conditional-Facts (Hard)} induces \textbf{summary failures}:
all systems over-compress, either \textit{altering} the original message or
\textit{stripping away} precise, critical details---both mislead the LLM into
the wrong judgment.
\item \textit{Persona-Retrieval} induces \textbf{summary failures} via
over-compression of long personas; the exception is Mem0, which fails to
store all details in the first place due to its LLM-tool-call update
mechanism.
\item \textit{Long-Hop} unsurprisingly induces \textbf{retrieval failures}:
systems fail to capture long-range causal relationships between seemingly
disjoint entities.
\end{itemize}
Except for Mem0, systems generally do not exhibit \textbf{storage failures};
failures stem almost entirely from incorrect summarization or retrieval.

\textbf{Reasoning failures do not absorb retrieval noise.} A natural concern is that increasing $k$ also puts more similar but irrelevant memories into the test-taker’s context. In that case, some of the failures we call reasoning failures could actually be retrieval failures. We would therefore expect reasoning failures to increase as $k$ increases. But Figures~\ref{fig:app-gemini-err}--\ref{fig:app-gpt54-err} show that this does not happen. On every dataset except \textit{Persona-Retrieval (Misleading)}, reasoning failures are rare and do not significantly increase from $k{=}5$ to $k{=}20$.

The exception is also useful for understanding what is going on. On misleading persona queries, larger $k$ brings in more distractor personas. The test-taker sometimes says it does not remember the queried person and then answers based on the habits of a different person (Appendix~\ref{app:reasoning-failures}). This seems to be a limitation of the answerer LLM rather than the architecture. More importantly, it only happens on the one task where all four systems already perform well. On the tasks that drive our architectural claims, reasoning failures are close to zero.

\begin{examplebox}[Key Takeaway]
Increasing the number of retrieved memories yields performance gains on tasks where retrieval errors are common; it yields marginal returns when summary errors are the bottleneck.
\end{examplebox}

\begin{figure}[!htbp]
    \centering
    \includegraphics[width=1\linewidth]{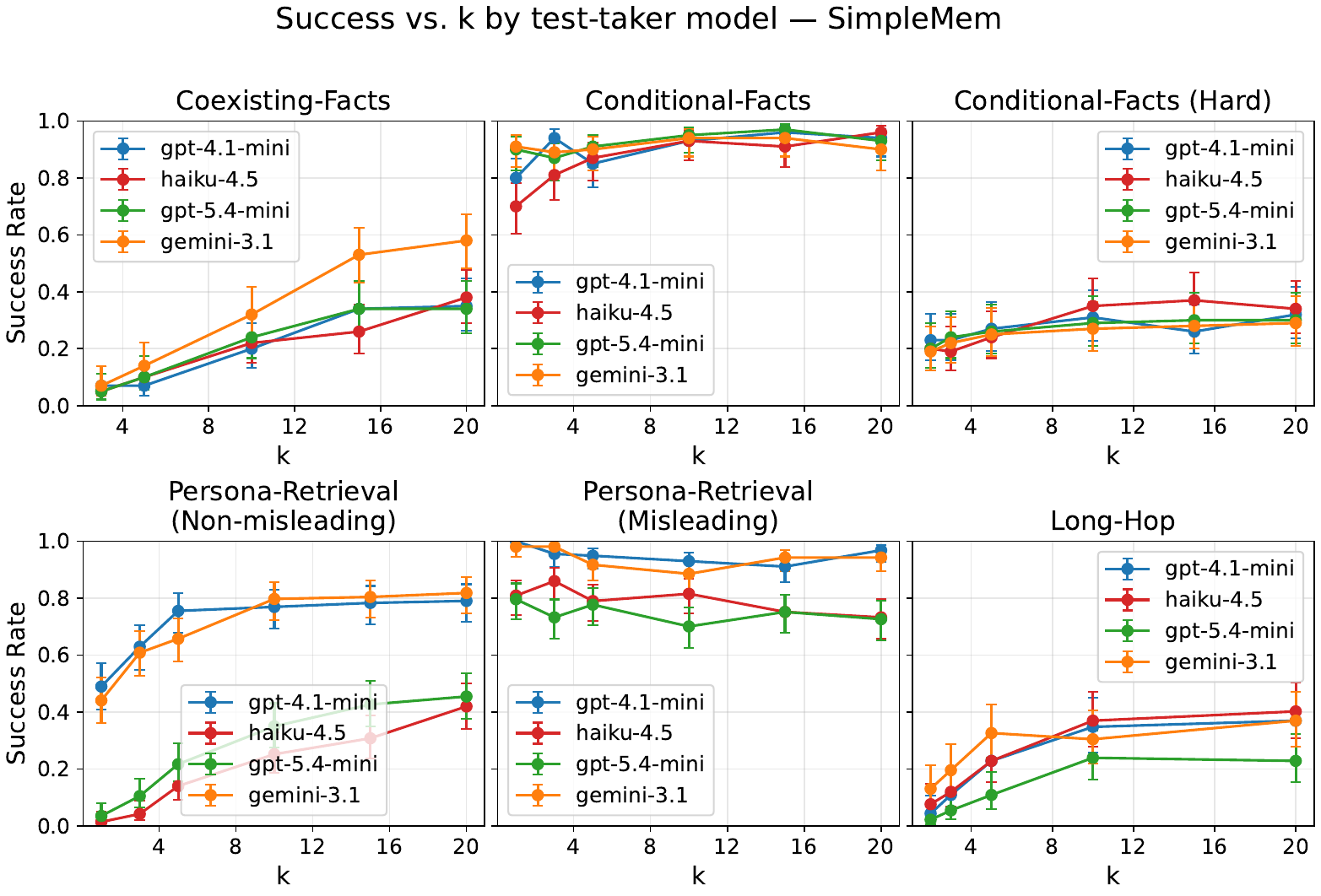}
    \includegraphics[width=1\linewidth]{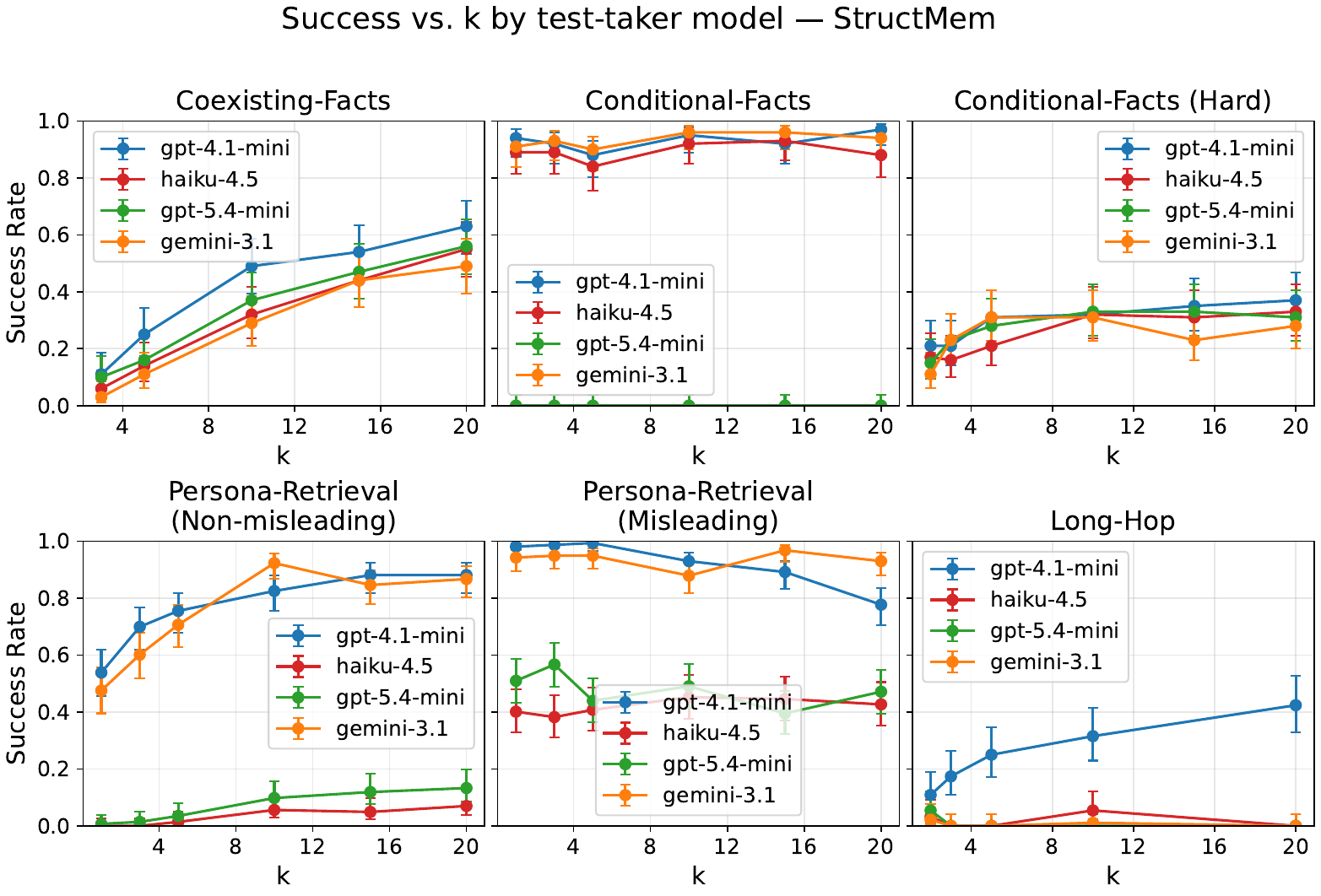}
    \caption{Performance of StructMem and SimpleMem as a function of their internal model. Mem0 and A-MEM follow the same trend, as shown in Appendix~\ref{app:detailed-results}, but we exclude them here for conciseness. Equipping the system with a stronger internal model does not lead to performance gains.}
    \label{fig:perf-vs-model}
\end{figure}

\textbf{Q2: How does accuracy scale with the strength of the model used by the memory system?}
Figure~\ref{fig:perf-vs-model} shows stronger models do not improve accuracy while sometimes degrading accuracy on most tasks; smarter reasoning models can
generate overly verbose memories that pollute the agent's context.

\begin{examplebox}[Key Takeaway]
Unlike other applications of LLM agents where using a more intelligent model can increase performance on benchmarks, modern memory systems are bound by architectural constraints rather than model intelligence.
\end{examplebox}

\begin{figure*}[!htbp]
    \centering
    \includegraphics[width=1\linewidth]{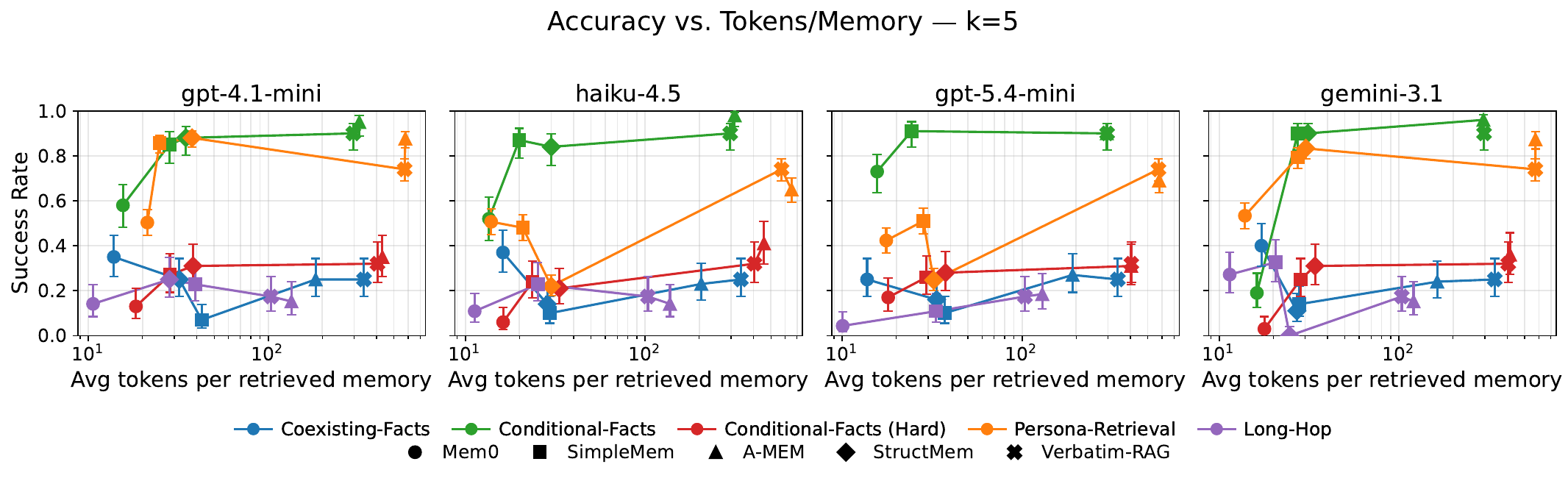}
    \caption{Per-model performance on \textsc{MemFail} relative to the average
    number of tokens per memory, at $k=5$. Marker \emph{shape} denotes the
    memory system---including \textbf{Verbatim-RAG}, whose stored memories are whole conversations---and \emph{color} the dataset; each colored line connects the
    systems on one dataset, ordered by token cost. A system's horizontal
    position is fixed by its architecture, which sets the granularity of a
    stored memory (A-MEM writes long descriptive notes, Mem0 short atomic
    facts), so the quantity of interest is the \emph{sign of the slope} of each
    colored line.}
    \label{fig:perf-vs-token}
\end{figure*}

\textbf{Q3: What does \textsc{MemFail} reveal about the tradeoff between performance and token consumption?}
Figure~\ref{fig:perf-vs-token} shows how performance scales with the token
usage of each memory system. Performance generally scales positively with token
consumption on \textit{Persona-Retrieval} and \textit{Conditional-Facts (Hard)}: in
general, for tasks bottlenecked by summary failures, increasing the number
of tokens leads to performance gains. By contrast, retrieval tasks can actually see a performance drop from using more tokens. This is especially evident for \textit{Coexisting-Facts}, where storing large memories ``pollutes'' the semantic embeddings, hurting retrieval.

\begin{examplebox}[Key Takeaway]
The historical consensus is that more tokens is an easy way to obtain performance gains (i.e., test-time scaling); however, for memory systems, performance scales with increased memory storage in a way that is \textit{highly task-dependent}.
\end{examplebox}

\textbf{Q4: How much of the failure is intrinsic to compression?}
Figure~\ref{fig:gpt-4.1-mini-perf} plots Verbatim-RAG, which stores
conversations uncompressed, alongside the compression-based systems. It almost
never beats \emph{all} memory systems but frequently beats at least one,
another view of the tradeoff in Q1. It is strongest on
\textit{Conditional-Facts}, whose errors are dominated
by summarization---which verbatim storage removes by construction---and weakest
on \textit{Long-Hop} and \textit{Coexisting-Facts}, whose errors are dominated
by retrieval: with no summarization, nothing normalizes or links scattered
facts, so retrieving \emph{all} required entries is harder, not easier. It is
also among the most token-expensive systems we evaluate
(Figure~\ref{fig:perf-vs-token}), on par with A-MEM.

\begin{examplebox}[Key Takeaway]
Removing summarization eliminates summary failures but worsens retrieval failures: faithful storage and retrievable storage are themselves in tension.
\end{examplebox}

\textbf{Q5: What does \textsc{MemFail} reveal about design choices in memory systems?}
We provide insights into the key architectural decisions of the tested
memory systems:
\begin{itemize}
\item \textbf{LLM-based memory updates.} Mem0 updates memory via LLM tool
calls. For short experiences (e.g., one-sentence \textit{Coexisting-Facts}
entries) the LLM accurately stores the content, but for longer experiences
it fails to issue enough tool calls to capture all details---evidenced by
Mem0's high storage error rate on \textit{Persona-Retrieval}, which has the
longest entries (Figure~\ref{fig:gpt-4.1-mini-perf}).
\item \textbf{Semantic vector stores.} A-MEM stores conversations as
descriptive notes in a vector database. Figure~\ref{fig:perf-vs-token} shows this explodes
token usage for little gain: it reduces summary errors but, surprisingly,
does not improve retrieval-heavy tasks, since RAG-style embeddings fail to
capture inter-entity relationships in isolation. Mem0 by comparison gains
efficiency by compressing more intelligently---removing noise without
stripping critical detail.
\item \textbf{Graph-based architectures.} Prior work proposes graphs to
improve causal reasoning. \textsc{MemFail} confirms that StructMem, a
graph-based method, performs well on \textit{Long-Hop} and
\textit{Conditional-Facts} but poorly on general information retrieval. We
show there may be a tradeoff between flat vector stores and graph-based
architectures: graphs improve inter-entity relationship modeling but
over-commit to structure/decomposition and struggle to represent longer
semantic ideas.
\end{itemize}

\textbf{Q6: What does \textsc{MemFail} reveal about future directions for memory systems research?}
Ultimately, we want memory systems that excel across all \textsc{MemFail}
tasks without major tradeoffs. We propose two under-explored directions that may help accomplish this goal.

\textbf{Mixture-of-memories architectures.} State-of-the-art systems commit to a single backend (vector store, graph, or hierarchical), but
\textsc{MemFail} shows that different architectures excel at different tasks;
hybrid systems that route memories to the appropriate substore could combine
these strengths. For example, a system could route causal experiences to a graph (StructMem-style) and
persona knowledge to a flat vector store (A-MEM-style).

\textbf{Task-based token scaling.} Figure~\ref{fig:perf-vs-token} shows that accuracy scales
with token consumption only for certain task types; improved systems may
dynamically size generated memories to match the type of incoming information.
For many tasks, more tokens is not always better. We also see that A-MEM uses substantially more tokens than other systems without commensurate performance gains, further motivating the need for intelligent scaling.

\section{Conclusion}
\label{sec:conclusion}

We introduce \textsc{MemFail}, a benchmark for LLM memory systems designed to expose specific failure modes. Our evaluation reveals that current systems are bound by architectural constraints that cannot be addressed by simply spending more tokens or using a more intelligent model. To our knowledge, \textsc{MemFail} is the first benchmark to enable a fine-grained analysis of  failure modes exhibited by different memory systems. \textsc{MemFail} supports evaluation of any memory system implementing the API from Section~\ref{sec:framework}, and we have open-sourced all datasets and evaluation code so that future memory systems may be benchmarked using \textsc{MemFail}.

\section*{Limitations}

While \textsc{MemFail} provides a fine-grained view into the failure modes of modern memory systems, several limitations of our methodology are worth highlighting. Every dataset in \textsc{MemFail} is generated by an LLM (\texttt{gpt-4.1-mini}, \texttt{gpt-5-mini}, or \texttt{gpt-5}) under structured generation prompts and then filtered. Although we manually verify all dataset entries for correctness, the resulting distribution of conversations, entities, and phrasings may be narrower than what a deployed memory system would encounter in practice. Performance on \textsc{MemFail} is designed to be interpreted as a diagnostic signal about specific failure modes rather than as a prediction of end-to-end performance in deployment. Additionally, we evaluate four open-source systems (Mem0, A-MEM, SimpleMem, StructMem) that all expose \texttt{store\_conversation}, \texttt{retrieve\_memories}, and \texttt{get\_all\_memories}. While our framework is intentionally designed to encompass the majority of memory systems, it may be harder to use when evaluating systems with implicit or learned memory, or with fine-tuned-weight memory. Our work also does not analyze the latency of memory systems; although the \textsc{MemFail} evaluation pipeline actually provides thorough metrics for memory retrieval time and evaluation time, this work intentionally chooses not to focus on latency in this work as the results are not as insightful.

\section*{Ethical considerations}

\textsc{MemFail} surfaces hidden failure modes in LLM memory systems, helping researchers and practitioners build more reliable personal assistants. However, we also recognize that great care should be taken in using \textsc{MemFail} responsibly so that it is not used adversarially to target specific weaknesses of real-world memory systems. We have open-sourced all datasets and code so that the research community may use \textsc{MemFail} to improve the security and accuracy of memory systems. Additionall, all datasets are synthetically generated, fake personas and contain no personal or identifying information about real individuals. The benchmark datasets in \textsc{MemFail} are generated by LLMs (\texttt{gpt-4.1-mini}, \texttt{gpt-5-mini}, and \texttt{gpt-5}) under structured prompts, as detailed in Section~\ref{sec:benchmark} and Appendix~\ref{app:construction}. AI assistants were also used to help polish writing in this paper; all technical content, claims, and analyses are the authors' own. Note that Mem0 uses an Apache 2.0 license, StructMem, SimpleMem, and A-MEM use an MIT license; we use these code artifacts only for this research which is consistent with their licenses. We have publicly released our code and datasets under a MIT license.

\bibliography{references}

\appendix
% Appendix entry point. `\appendix` is issued in main.tex prior to inputting
% this file, so the sections below are automatically lettered A, B, C, ...

\section{Failure-Mode Examples}
\label{app:failure-examples}

This appendix gives concrete examples of the failure modes described in
Section~\ref{sec:failure-modes}. All examples are drawn from the recorded JSON
evaluation traces. Each example reports the original asserted information, the
memories retrieved at query time, the graded question, the model answer, and
the resulting diagnosis.

We distinguish carefully between failures of the memory system and failures of
the downstream test-taker. A reasoning failure is assigned only when the
retrieved memory is correct and relevant, but the final answer contradicts it
or fails to incorporate it. If the retrieved memories are empty, irrelevant, or
missing one or more required facts, the error is attributed to storage or
retrieval rather than downstream reasoning.

\subsection{Summary failures}
\label{app:summary-failures}

A summary failure occurs when the system stores information, but the stored
memory does not faithfully preserve the original content needed for the later
query. In the examples below, the system stores a related memory, but the
stored version loses a critical condition, threshold, or exclusivity qualifier.

\begin{examplebox}[Summary failure: Thorne's motivation threshold]
\textbf{Original asserted information.}
``Thorne starts building miniature clockwork automatons only when his motivation
is 8 out of 10 or higher.''\\[2pt]

\textbf{Retrieved memories at query time.}
\begin{itemize}
    \item ``At 2026-04-28T06:22:25, Thorne builds tiny clockwork automatons
    that are wound obsessively and tested until each automaton ticks with a
    personality.''
    \item ``At 2026-04-28T06:22:25, Thorne keeps a motivation chart on the
    workshop wall with stickers for weeks when Thorne's focus spikes.''
\end{itemize}

\textbf{Question.}
``Thorne says he's at about 5/10 motivation today --- will he start building
one of his miniature clockwork automatons now?''\\[2pt]

\textbf{Model answer.}
``Probably not a full new build right now --- but he might tinker or do a small
related task.''\\[2pt]

\textbf{Ground truth.}
No. He requires motivation of 8/10 or higher before he begins building
automatons.\\[2pt]

\textbf{Diagnosis.}
This is a summary failure. The memories preserve that Thorne builds automatons
and that his work is associated with motivation spikes, but they lose the
specific numeric threshold ``8 out of 10 or higher.'' Because the retrieved
memory does not contain the exact condition, this is not a reasoning failure.
\end{examplebox}

\begin{examplebox}[Summary failure: Captain Aurelio's black-tea trigger]
\textbf{Original asserted information.}
``Captain Aurelio recites a two-line naval proverb only when someone offers him
a cup of black tea.''\\[2pt]

\textbf{Retrieved memories at query time.}
\begin{itemize}
    \item ``Three candidate two-line naval proverbs in Captain Aurelio's voice
    are: `Mind the swell, tend the line. Steady hands keep true time.';
    `Respect the sea, respect the men. Small comforts steady the soul.'; and
    `Chart the calm, brace the storm. Keep your cup --- keep your calm.'''
    \item ``Captain Aurelio's ritual proverb is a two-line naval proverb spoken
    aloud with a soft and concise cadence unchanged over decades.''
\end{itemize}

\textbf{Question.}
``We offered biscuits to Captain Aurelio but no drinks were handed out --- will
he recite his two-line naval proverb now?''\\[2pt]

\textbf{Model answer.}
``Maybe --- not certain.''\\[2pt]

\textbf{Ground truth.}
No. He only recites the proverb when someone gives him a cup of black tea.\\[2pt]

\textbf{Diagnosis.}
This is a summary failure. The stored memories preserve the proverb ritual, and
some stored information links the ritual to a cup, but the critical trigger is
weakened: the memory does not preserve that the cup must be black tea or that
the condition is exclusive.
\end{examplebox}

\begin{examplebox}[Summary failure: Mochi's five-minute laser-pointer threshold]
\textbf{Original asserted information.}
``Mochi drags a catnip mouse into the laundry basket and naps on it after an
uninterrupted five-minute laser-pointer chase.''\\[2pt]

\textbf{Retrieved memories at query time.}
\begin{itemize}
    \item ``Mochi drags a battered catnip mouse into the laundry basket and
    curls up on top of the catnip mouse for naps.''
    \item ``Mochi is a devoted chaser who sprints across rugs with feline
    intensity during play sessions.''
\end{itemize}

\textbf{Question.}
``Mochi just chased the laser pointer for six straight minutes --- will she
bring a catnip mouse into the laundry basket and nap on it now?''\\[2pt]

\textbf{Model answer.}
``Very likely --- but not guaranteed.''\\[2pt]

\textbf{Ground truth.}
Yes. Six minutes satisfies the five-minute laser-pointer condition.\\[2pt]

\textbf{Diagnosis.}
This is a summary failure. The stored memories preserve the behavior and a
general association with chasing, but they lose the exact uninterrupted
five-minute laser-pointer threshold.
\end{examplebox}

\subsection{Storage failures}
\label{app:storage-failures}

A storage failure occurs when information from the initial interaction is not
incorporated into memory in a form that can later be retrieved. In the examples
below, the original behavior or condition is absent from the memory store, so
the later query cannot be answered from memory.

\begin{examplebox}[Storage failure: Pixel's kitchen-windowsill condition]
\textbf{Original asserted information.}
Pixel watches birds only on the kitchen windowsill.\\[2pt]

\textbf{Stored memory state.}
The memory store contains that Pixel sits on a windowsill watching birds and is
often drawn to the sunlit, feeder-facing perch at the kitchen window, but it
does not preserve the exclusivity condition that Pixel does this only on the
kitchen windowsill.\\[2pt]

\textbf{Question.}
The graded query asks whether Pixel would perform the behavior away from the
kitchen windowsill.\\[2pt]

\textbf{Ground truth.}
No. The behavior is restricted to the kitchen windowsill.\\[2pt]

\textbf{Diagnosis.}
This is a storage failure. The behavior is present in memory, but the condition
requiring exclusivity is not stored; the memory only says Pixel is usually or
often drawn to the kitchen windowsill.
\end{examplebox}

\begin{examplebox}[Storage failure: watermelon seeds]
\textbf{Original asserted information.}
``Eating watermelon seeds causes watermelons to grow inside your stomach.''\\[2pt]

\textbf{Initial essay.}
The essay repeatedly states that watermelon seeds can take root inside the
stomach lining and lead to the growth of small watermelons internally.\\[2pt]

\textbf{Retrieved memories at query time.}
None.\\[2pt]

\textbf{Question.}
``Can planting watermelon seeds you swallow help you grow a watermelon garden
inside your body?''\\[2pt]

\textbf{Model answer.}
``No, swallowing watermelon seeds will not grow a watermelon garden inside your
body.''\\[2pt]

\textbf{Diagnosis.}
This is a storage-side failure under a faithful-memory criterion: the original
assertion is not available in retrieved memory when the implication question is
asked. The answer itself is not a reasoning failure, because the model was not
shown a correct retrieved memory that it failed to use.
\end{examplebox}

\begin{examplebox}[Storage failure: fortune-cookie origin]
\textbf{Original asserted information.}
``Fortune cookies originated in Los Angeles.''\\[2pt]

\textbf{Retrieved memories at query time.}
None.\\[2pt]

\textbf{Question.}
``Which city is known as the birthplace of fortune cookies?''\\[2pt]

\textbf{Model answer.}
``The city known as the birthplace of fortune cookies is San Francisco,
California.''\\[2pt]

\textbf{Diagnosis.}
This is a storage-side failure under a faithful-memory criterion: the Los
Angeles claim from the initial interaction is not surfaced at query time. The
model answers from prior knowledge rather than from a retrieved
conversation-specific memory.
\end{examplebox}

\subsection{Retrieval failures}
\label{app:retrieval-failures}

A retrieval failure occurs when the memory system fails to return the memories
needed for the current query, or when it returns memories that are unrelated to
the query. These are not reasoning failures: the downstream model cannot use
the correct memory if the retriever never provides it.

\begin{examplebox}[Retrieval failure: Amara's hometown condition is not retrieved]
\textbf{Original asserted information.}
``Amara writes a postcard to her childhood teacher only when she is physically
in her hometown.''\\[2pt]

\textbf{Retrieved memories at query time.}
\begin{itemize}
    \item ``Amara writes a warm, handwritten postcard to the teacher who shaped
    Amara, but only during certain visits to places.''
    \item ``Amara keeps a stack of postcards collected from places that Amara
    loves.''
\end{itemize}

\textbf{Question.}
``We're vacationing at a coastal resort far from where she grew up --- would
Amara write a postcard to her childhood teacher now?''\\[2pt]

\textbf{Model answer.}
``Maybe --- she does sometimes, but not every trip.''\\[2pt]

\textbf{Ground truth.}
No. She only writes those postcards when physically in her hometown, and the
resort is not her hometown.\\[2pt]

\textbf{Diagnosis.}
This is a retrieval failure. The memory store contains enough information to
recover the hometown-only condition, but the retrieved memories only say
``certain visits to places'' and do not include the hometown-specific trigger.
\end{examplebox}

\begin{examplebox}[Retrieval failure: Eldon's victory-jig behavior is not retrieved]
\textbf{Original asserted information.}
``Eldon performs a private victory jig after the guildmaster gives him a
thumbs-up.''\\[2pt]

\textbf{Retrieved memories at query time.}
\begin{itemize}
    \item ``At 2026-04-28T06:37:58, the apprentices mutter good-natured barbs
    about Eldon's modesty.''
    \item ``At 2026-04-28T06:37:58, the guildmaster is described as a blunt
    woman with an economical sense of praise who gives a curt thumbs-up on the
    rare occasions when the guildmaster is pleased.''
\end{itemize}

\textbf{Question.}
``The guildmaster just gave Eldon a thumbs-up during the meeting --- will Eldon
do his victory jig right now?''\\[2pt]

\textbf{Model answer.}
``Unlikely. Eldon's known to be modest, so a public, ostentatious `victory jig'
isn't consistent with his usual behavior.''\\[2pt]

\textbf{Ground truth.}
Yes. The thumbs-up is exactly the trigger for Eldon's private victory jig.\\[2pt]

\textbf{Diagnosis.}
This is a retrieval failure. The retrieved memories contain the thumbs-up
condition, but not the private victory-jig behavior. Since the required behavior
is missing from the retrieved context, the downstream model cannot recover the
conditional rule.
\end{examplebox}

\begin{examplebox}[Retrieval failure: long-hop chain retrieves only the first link]
\textbf{Original facts.}
\begin{itemize}
    \item ``Carlos adores rainy afternoons because they slow his day.''
    \item ``Rainy afternoons prompt Carlos to bake banana bread.''
    \item ``Banana bread makes Carlos post cheerful photos.''
\end{itemize}

\textbf{Retrieved memories at query time.}
\begin{itemize}
    \item ``On foggy mornings Priyanka chooses to wear bright scarves.''
    \item ``Carlos adores rainy afternoons because they slow his day.''
\end{itemize}

\textbf{Question.}
``On the kind of afternoons Carlos adores, what does he end up doing?''\\[2pt]

\textbf{Choices.}
A. polish dress shoes \quad
B. post cheerful photos \quad
C. schedule a dentist appointment \quad
D. back up files \quad
E. update the calendar\\[2pt]

\textbf{Model answer.}
\texttt{\{"selected\_choice":"D"\}}\\[2pt]

\textbf{Ground truth.}
B. post cheerful photos.\\[2pt]

\textbf{Diagnosis.}
This is a retrieval failure. The retriever returned the first link in the
chain, but it did not retrieve the two remaining supporting facts connecting
rainy afternoons to banana bread and banana bread to posting cheerful photos.
\end{examplebox}

\begin{examplebox}[Retrieval failure: blood-color query retrieves fortune-cookie memory]
\textbf{Original asserted information.}
``Veins appear blue because deoxygenated blood is inherently blue in color.''\\[2pt]

\textbf{Retrieved memories at query time.}
``The city known as the birthplace of fortune cookies is San Francisco.''\\[2pt]

\textbf{Question.}
``What color would blood appear if it were exposed to air outside the body?''\\[2pt]

\textbf{Model answer.}
``When blood is exposed to air outside the body, it generally appears bright
red.''\\[2pt]

\textbf{Diagnosis.}
This is a retrieval failure: the retrieved memory is about fortune cookies and
is unrelated to the blood-color question. The final answer is correct, but that
does not make retrieval successful. The trace shows harmless retrieval
contamination.
\end{examplebox}

\subsection{Reasoning failures}
\label{app:reasoning-failures}

A reasoning failure occurs when the relevant memory is stored and retrieved,
but the downstream test-taker still fails to use it correctly. In these cases,
the memory system has provided the information needed to answer the query, but
the model hedges, contradicts the retrieved memory, or reasons from an
irrelevant implication rather than applying the remembered rule.

\begin{examplebox}[Reasoning failure: Marek's seven-year threshold]
\textbf{Original asserted information.}
``Marek gives his carved wooden coin as a keepsake only to people he has been
friends with for at least seven years.''\\[2pt]

\textbf{Retrieved memories at query time.}
\begin{itemize}
    \item ``By the time a name has lived in Marek's friendship ledger for seven
    years, Marek is usually ready to offer the carved wooden coin as a gift.''
    \item ``Marek carries a single carved wooden coin in the pocket of his coat
    and treats the coin like a talisman to be handed out sparingly.''
\end{itemize}

\textbf{Question.}
``I've known Marek for eight years --- will he give me his carved wooden
coin?''\\[2pt]

\textbf{Model answer.}
``Probably --- but not guaranteed.''\\[2pt]

\textbf{Ground truth.}
Yes. The question states that the friendship has lasted eight years, which
satisfies the seven-year condition.\\[2pt]

\textbf{Diagnosis.}
This is a reasoning failure. The retrieved memory contains both the behavior
and the seven-year threshold, and the query context satisfies that threshold.
The model quotes the relevant memory but still hedges instead of giving the
definite answer supported by the memory.
\end{examplebox}

\begin{examplebox}[Reasoning failure: Eldra's autumn-only bookmark rule]
\textbf{Original asserted information.}
Eldra weaves pressed leaves into bookmarks during autumn.\\[2pt]

\textbf{Retrieved memories at query time.}
The retrieved memories explicitly state that Eldra presses autumnal foliage and
weaves the fragile leaves into slim, hand-tied bookmarks, and that these
bookmarks are tied with twine and passed at the book club.\\[2pt]

\textbf{Question.}
The graded query asks whether Eldra would weave pressed leaves into bookmarks
in March.\\[2pt]

\textbf{Model answer.}
The model answers ``maybe'' and speculates about leftover finished bookmarks
rather than applying the seasonal condition.\\[2pt]

\textbf{Ground truth.}
No. March is not autumn, so the autumn-only condition is not satisfied.\\[2pt]

\textbf{Diagnosis.}
This is a reasoning failure. The retrieved memory contains the correct
seasonal condition and the bookmark behavior, but the model does not apply the
condition to the March query. Instead, it introduces speculation about leftover
bookmarks and fails to reach the required negative answer.
\end{examplebox}

\begin{examplebox}[Reasoning failure: Nacho's upstairs-balcony condition]
\textbf{Original asserted information.}
``Nacho launches a loud four-minute bark routine at passersby only when he is
perched on the upstairs balcony.''\\[2pt]

\textbf{Retrieved memories at query time.}
\begin{itemize}
    \item ``Nacho usually begins the theatrical barking performance from the
    high, outside perch on the upstairs balcony.''
    \item ``Nacho is a compact terrier, very alert and loud for his size, has a
    practiced approximately four-minute barking routine at passersby, loves
    high vantage points with a favorite being an upstairs balcony or ledge,
    patrols the balcony at dawn, and likes pats and a quiet treat after
    barking.''
\end{itemize}

\textbf{Question.}
``The delivery person is at the gate and Nacho is sitting by the front door
downstairs; will he start his upstairs balcony barking routine?''\\[2pt]

\textbf{Model answer.}
``Probably --- he often goes up to the balcony to start the full barking
routine, so if the delivery person is at the gate he's likely to either trot
upstairs to the balcony and launch into his usual performance or at least give
some alarms from the door.''\\[2pt]

\textbf{Ground truth.}
No. Nacho only performs that barking routine from the upstairs balcony, and he
is downstairs now.\\[2pt]

\textbf{Diagnosis.}
This is a reasoning failure. The retrieved memories correctly identify the
upstairs balcony as the location for the barking routine, and the query says
Nacho is downstairs. The model nevertheless predicts that he will probably
perform the routine, adding an unsupported transition in which he might trot
upstairs.
\end{examplebox}

\section{Benchmark Construction Details}
\label{app:construction}

This appendix gives the full pipeline specifications, model choices,
deduplication thresholds, and complete sampling pools that were summarized in
Section~\ref{sec:benchmark}.

\subsection{Generators and shared infrastructure}
\label{app:generators}

All five datasets are produced by single-pass batched OpenAI calls with
\texttt{response\_format=\{"type": "json\_object"\}} and up to three retries
per batch on validation failure. Generator models are pinned per dataset:
\textit{Conditional-Facts (Easy)} and \textit{Coexisting-Facts} use
\texttt{gpt-4.1-mini}; \textit{Conditional-Facts (Hard)} and
\textit{Persona-Retrieval} use \texttt{gpt-5-mini}; \textit{Long-Hop} uses
\texttt{gpt-5}. Every dataset uses a fixed random seed of $42$ and writes a
\texttt{generation\_config.json} alongside the CSV, recording the seed, the
git commit, the deduplication threshold, the per-row counts before and after
deduplication, and the model name, so that any row in any released artifact is
fully traceable back to the generator that produced it.

Deduplication is performed with MinHash LSH at task-specific Jaccard thresholds
over a task-specific dedup key:
\textit{Conditional-Facts} dedups at $0.8$ over the wrapping essay text;
\textit{Coexisting-Facts} dedups at $0.7$ over the scenario question;
\textit{Persona-Retrieval} dedups at $0.7$ over the essay text;
\textit{Long-Hop} dedups at $0.7$ over each \emph{individual} fact, and any
chain that contains a fact colliding with an already-kept fact is dropped
wholesale.

\subsection{Conditional-Facts: condition types}
\label{app:conditional-types}

The condition type is sampled uniformly from a fixed list of $32$ types,
grouped by category in Table~\ref{tab:condition-types}. For \texttt{pet}
entities the condition type is restricted to externally observable triggers
(\texttt{time\_of\_day}, \texttt{weather}, \texttt{temperature},
\texttt{location}, \texttt{noise\_level}, \texttt{lighting},
\texttt{food\_or\_drink\_present}, \texttt{prior\_activity},
\texttt{company}) so that no pet is asked to depend on abstract internal
states.

\begin{table}[h]
\centering
\small
\begin{tabular}{@{}p{0.18\linewidth}p{0.75\linewidth}@{}}
\toprule
\textbf{Group} & \textbf{Condition types} \\
\midrule
Time         & \texttt{time\_of\_day}, \texttt{day\_of\_week}, \texttt{season}, \texttt{time\_elapsed} \\
Environment  & \texttt{weather}, \texttt{temperature}, \texttt{location}, \texttt{noise\_level}, \texttt{lighting} \\
Physical     & \texttt{hunger\_level}, \texttt{energy\_level}, \texttt{pain\_or\_discomfort}, \texttt{sobriety} \\
Emotional    & \texttt{mood}, \texttt{stress\_level}, \texttt{anxiety\_level}, \texttt{motivation\_level} \\
Social       & \texttt{company}, \texttt{social\_setting}, \texttt{relationship\_closeness} \\
Task         & \texttt{task\_type}, \texttt{workload}, \texttt{completion\_state} \\
Sensory      & \texttt{music\_playing}, \texttt{scent}, \texttt{food\_or\_drink\_present} \\
Relational   & \texttt{conflict\_state}, \texttt{approval\_received}, \texttt{request\_made} \\
Habitual     & \texttt{prior\_activity}, \texttt{frequency\_cap}, \texttt{streak\_state} \\
\bottomrule
\end{tabular}
\caption{The full list of $32$ condition types for \textit{Conditional-Facts}.}
\label{tab:condition-types}
\end{table}

\subsection{Conditional-Facts: the Hard decomposition}
\label{app:hard-decomposition}

The Hard variant decomposes the original conditional fact into exactly three
sentences spread across an $8$--$12$ sentence essay:
\begin{itemize}[leftmargin=*]
    \item a \textbf{behavior} sentence describing what the entity does as a
    tendency or habit, \emph{without} naming the trigger;
    \item a \textbf{condition} sentence establishing when $C$ holds as part of
    the entity's life context or environment, \emph{without} naming the
    behavior;
    \item a \textbf{link} sentence using soft co-occurrence language (``it's
    usually in those moods that\ldots,'' ``by then\ldots,'' ``most of the time
    it happens\ldots'') that connects the two without an explicit conditional.
\end{itemize}
The generator is forbidden from using any explicit conditional connective
(``only when,'' ``unless,'' ``except when,'' ``but only if,'' ``whenever,''
``if,'' ``only after,'' ``only if'') anywhere in the essay, and the generation
prompt mandates that no two of the three rule-bearing sentences are adjacent:
at least one unconditional sentence sits between any pair. Easy and Hard rows
share the same entity and condition specs, so any difference in performance can
be directly attributed to the distribution of the rule across sentences.

\begin{examplebox}[Example: \textit{Conditional-Facts} (Hard)]
\textbf{Entity:} Gideon\quad\textbf{Behavior:} paints tiny sailboats inside used
teacups\quad\textbf{Condition:} when he feels wistful after finding a childhood
keepsake\\[2pt]
\textbf{Essay (excerpt):} ``\ldots He has a habit of painting tiny sailboats
inside those used teacups, as if making miniature voyages.\ldots\ He collects
small keepsakes from his childhood---a dented tin soldier, a faded postcard
tucked in a book. Those artifacts sometimes resurface and leave him feeling
peculiarly nostalgic and soft.\ldots\ It's usually in those wistful, quiet moods
that this little, private ritual reappears.\ldots''\\[2pt]
\textbf{Question:} ``We just found Gideon's old toy soldier in the attic;
would he paint tiny sailboats in a teacup right now?''\\[2pt]
\textbf{Ground truth:} ``Yes---finding a childhood keepsake makes him feel
wistful, which is exactly when he paints sailboats in teacups.''
\end{examplebox}

\subsection{Coexisting-Facts: full list of preference categories}
\label{app:coexisting-categories}

The $100$ preference categories are partitioned into thematic groups in
Table~\ref{tab:preference-categories}. Each category yields exactly one row,
and the per-row preference count $N\!\in\!\{2,3,4,5\}$ is drawn uniformly per
row.

\begin{table*}[h]
\centering
\small
\begin{tabular}{@{}p{0.15\linewidth}p{0.82\linewidth}@{}}
\toprule
\textbf{Group} & \textbf{Categories} \\
\midrule
Food \& drink & foods; cuisines; types of desserts; drinks; coffee types; tea varieties; breakfast foods; street foods; snack types; cocktail types; cooking methods; restaurant types \\
\addlinespace
Entertainment \& media & music genres; types of movies; book genres; tv show genres; video game genres; podcast topics; comedy styles; documentary subjects; types of concerts; types of museums; sports to watch on TV; types of theater \\
\addlinespace
Physical activity & sports; outdoor activities; physical exercises; fitness class types; martial arts styles; dance styles; cycling types; swimming styles; types of yoga; hiking styles \\
\addlinespace
Lifestyle \& personal style & fashion styles; home decor styles; makeup styles; jewelry types; shoe styles; hair care styles; skincare routines; bag styles; hat styles \\
\addlinespace
Travel \& places & travel destinations; vacation types; nature environments; city neighborhood types; outdoor dining settings \\
\addlinespace
Hobbies \& creativity & hobbies; crafts; art styles; photography styles; creative writing styles; musical instruments; types of puzzles; journaling styles; home improvement projects; garden types; houseplants \\
\addlinespace
Social \& leisure & weekend activities; party activities; dating activities; types of social gatherings; board games; card games; conversation topics \\
\addlinespace
Tech \& learning & programming languages; tech interests; academic subjects; skills to learn; languages to learn; language learning methods; study techniques; note-taking methods; productivity methods \\
\addlinespace
Wellness \& self-care & mental health activities; meditation types; self-care activities; sleep habits; morning routine habits; evening routine habits \\
\addlinespace
Career \& finance & career fields; types of volunteering; investment types; budget planning methods \\
\addlinespace
Media consumption & social media content types; news formats; podcast formats; types of reading material; audio listening formats; content creation types \\
\addlinespace
Misc & animals; dog breeds; collectibles; subscription box types; car types; home organization methods; types of naps; camping styles; astrology interests \\
\bottomrule
\end{tabular}
\caption{The full list of $100$ preference categories for
\textit{Coexisting-Facts}, grouped thematically.}
\label{tab:preference-categories}
\end{table*}

\subsection{Persona-Retrieval: name and flavor pools}
\label{app:persona-pools}

Entity names are drawn from a fixed pool of $30$ diverse names; persona
``flavors'' are drawn from a fixed pool of $30$ flavors. Both pools are listed
in Table~\ref{tab:persona-pools}. Within a row, the entity name is sampled
once, and the three misleading-slot distractors are then drawn without
replacement from the remaining $29$ names so that no two slots in a row reuse
the same distractor and the entity itself is never used as its own distractor.

\begin{table*}[h]
\centering
\small
\begin{tabular}{@{}p{0.42\linewidth}p{0.55\linewidth}@{}}
\toprule
\textbf{Names ($30$)} & \textbf{Persona flavors ($30$)} \\
\midrule
Ava Thompson; Liam Carter; Maya Patel; Noah Brooks; Zoe Kim; Ethan Rivera; Priya Shah; Lucas Bennett; Sofia Nguyen; Daniel Park; Elena Rossi; Marcus Lee; Amara Okafor; Jonas Weber; Hana Sato; Theo Laurent; Nia Williams; Ravi Iyer; Clara Schmidt; Diego Alvarez; Yuki Tanaka; Sasha Petrov; Imani Johnson; Felix Andersen; Leila Haddad; Owen Murphy; Anya Volkov; Caleb Foster; Mei Zhang; Tomas Costa
&
a meticulous indoor gardener with strong opinions about humidity; a lapsed competitive swimmer who now coaches youth weekend meets; a sound engineer obsessed with vintage analog gear; a part-time pastry chef who does math research on the side; a long-distance hiker training for the Pacific Crest Trail; a retired ER nurse who took up woodworking after retirement; a beekeeper-turned-marketing-consultant who still keeps three hives; an amateur astronomer who hates city light pollution; a freelance translator working between Portuguese and Korean; a cybersecurity researcher who collects vintage typewriters; a former competitive figure skater now running a small tea shop; a chef-instructor who teaches knife skills at a community college; an opera singer who is also a part-time auto mechanic; an architect specializing in adaptive reuse of old factories; a wildlife photographer focused on owls in the Pacific Northwest; a high-school chemistry teacher who restores vintage motorcycles; a marathoner with a rare allergy to most stone fruits; a bookbinder who designs board games on weekends; a software engineer who breeds carnivorous plants; a paramedic who plays cello in a community orchestra; a former diplomat now running a pottery studio; a cartographer obsessed with historic shipwrecks; a dog trainer specializing in working breeds; a forensic accountant who writes science fiction novels; a glassblower with severe pollen allergies; a sommelier transitioning to non-alcoholic beverage consulting; a former orchestra conductor who now teaches sailing; a museum conservator focused on 19th-century photographs; a competitive bridge player who works as an actuary; a backcountry ski guide who restores antique furniture in summer \\
\bottomrule
\end{tabular}
\caption{The full name pool and persona flavor pool used by
\textit{Persona-Retrieval}.}
\label{tab:persona-pools}
\end{table*}

\subsection{Long-Hop: full pipeline}
\label{app:long-hop-pipeline}

\paragraph{Generation.}
Chains are generated in batches with \texttt{gpt-5}, accumulating a running
list of one-line summaries (head $\to$ terminal) of every previously accepted
chain so the model is steered toward narratively novel chains. The system
prompt enforces every hard rule listed in Section~\ref{sec:long-hop}: exactly
$K{+}1$ statements, exactly $K{+}2$ distinct anchors, statement $i$ links
anchors $i$ and $i{+}1$ via an explicit relation phrase, every fact is
self-contained (any pronoun's antecedent must appear in the same sentence),
every fact is subjective (no encyclopedic claims), and middle / terminal
anchors appear only in the two facts that border them. The graded question
references the head anchor at least once by name, asks about the terminal
anchor, and never names any intermediate anchor.

\paragraph{Local validation.}
Each generated chain is normalized (lowercased, punctuation stripped, whitespace
collapsed) and checked structurally: the chain must contain exactly $K{+}1$
non-empty facts and $K{+}2$ distinct anchors; every anchor must appear
literally as a substring in each of the (one or two) facts it borders; the
head anchor (or its last $\ge 3$-character token, to tolerate morphological
variants such as ``eat apples'' $\to$ ``eating apples'') must appear in the
graded question; no intermediate anchor may appear in the question; and the
ground-truth answer must contain the terminal anchor as a substring (or vice
versa).

\paragraph{Cross-chain conflict check.}
Survivors are passed through an LLM judge (\texttt{gpt-5}) with a global pass
plus overlapping sliding windows. The judge drops chains that contradict
another chain (incompatible claims about the same anchor), share a distinctive
proper-noun or distinctive composite phrase with another chain, or retell the
same narrative with the same anchors in the same role. Generic single-word
concepts repeating across chains (``sleep,'' ``bored,'' ``tea'') are
explicitly allowed.

\paragraph{Fact-level deduplication.}
Every individual fact across every surviving chain is then run through MinHash
LSH at a Jaccard threshold of $0.7$. Any chain that contains a fact colliding
with a fact already kept is dropped wholesale (rather than just dropping the
offending fact), so that the released dataset has no near-duplicate fact pair
across any two chains.

\paragraph{Distractor generation.}
For each surviving chain we generate four distractor options with \texttt{gpt-5}
in parallel calls. Each distractor must (i) match the correct answer in
grammatical form, length range, and answer category; (ii) be realistic and
ordinary (no absurd, surreal, or comically random options); (iii) be
unambiguously wrong (not a paraphrase, synonym, sub-phrase, or near-spelling
of the correct answer or any anchor / relation phrase appearing in any fact);
(iv) be \emph{orthogonal to every fact in the chain}---a reader looking at any
single fact in isolation must not be able to guess the distractor as a
plausible ``what comes next'' or ``natural consequence'' via common-sense
world knowledge; and (v) be distinct from the other three distractors. Chains
whose distractor generation fails validation after retries are dropped. The
five options are then shuffled per chain and the correct letter is recorded.

\paragraph{Final balancing.}
We oversample chains per hop count and then truncate each hop bucket to a
fixed target. The final released dataset contains $31$, $32$, and $29$ chains
at $K{=}1$, $K{=}2$, and $K{=}3$ respectively, for a total of $92$ chains and
$274$ facts.

\paragraph{Storage layout at evaluation time.}
The $K{+}1$ facts of a chain are \emph{never} co-located. At evaluation time,
every fact across the full dataset is shuffled and bin-packed into storage
conversations under a hard constraint that no two facts from the same chain
ever land in the same conversation; the graded question for each chain is
then asked in its own separate conversation, in independently shuffled order.

\subsection{Additional examples}
\label{app:more-examples}

\begin{examplebox}[Example: \textit{Persona-Retrieval}, all three slots]
\textbf{Entity:} Yuki Tanaka\\[2pt]
\textbf{Essay (excerpt):} ``Yuki Tanaka spends most mornings hunched over a
lightbox, tracing contours from 18th-century wreck surveys.\ldots\ She refuses
to eat shellfish because she gets a severe hive reaction, so boat provisions
are always cooked onshore and strictly shellfish-free.\ldots\ Her studio is
lit by a single anglepoise lamp and a wall of hand-drawn dive
charts.\ldots''\\[2pt]
\textbf{Q1 (misleading):} ``Do you know anything about Noah Brooks?''\\
$\;\;\rightarrow$ ``I don't have information about Noah Brooks.''\\[2pt]
\textbf{Q2 (non-misleading):} ``What food should I avoid serving to Yuki
Tanaka?''\\
$\;\;\rightarrow$ ``Avoid shellfish---she has a severe allergic reaction.''\\[2pt]
\textbf{Q3 (misleading):} ``What kind of lamp does Clara Schmidt prefer for
detailed work?''\\
$\;\;\rightarrow$ ``I don't have information about Clara Schmidt.''
\end{examplebox}

\begin{examplebox}[Example: \textit{Long-Hop} ($K{=}1$, named-person voice)]
\textbf{Voice:} single-named-person (Diego)\\[2pt]
\textbf{Anchor chain:} ``Diego'' $\to$ ``Korean food'' $\to$ ``thirsty''\\[2pt]
\textbf{Facts (stored separately):}
\begin{itemize}
\item ``Diego loves Korean food.''
\item ``Korean food always leaves Diego thirsty.''
\end{itemize}
\textbf{Question:} ``What physical feeling does Diego's favorite cuisine
eventually cause?''\\[2pt]
\textbf{Choices (shuffled):} A.\,sleepy\quad B.\,nostalgic\quad
\textbf{C.\,thirsty}\quad D.\,restless\quad E.\,focused\\[2pt]
\textbf{Ground truth:} ``thirsty''.
\end{examplebox}

\begin{examplebox}[Example: \textit{Coexisting-Facts}, $N{=}5$]
\textbf{Category:} types of yoga\\[2pt]
\textbf{Preference facts (each stored in its own isolated conversation):}
\begin{itemize}
\item ``Vinyasa flow is what gets me moving on weekday mornings.''
\item ``I love a slow yin session at the end of a long workweek.''
\item ``Hot yoga is my go-to whenever I want to really sweat it out.''
\item ``Restorative yoga is perfect when I'm recovering from a tough run.''
\item ``Ashtanga is what I default to when I want a structured practice.''
\end{itemize}
\textbf{Question:} ``I'm putting together a yoga schedule for the month---what
styles should I rotate through to keep things varied?''\\[2pt]
\textbf{Ground truth:} vinyasa flow, yin, hot yoga, restorative yoga,
ashtanga.
\end{examplebox}

\section{Human Validation of the LLM Judge}
\label{app:judge-validation}

Table~\ref{tab:judge-confusion} reports the per-class confusion matrix between
the \texttt{gpt-5-mini} judge of Section~\ref{sec:eval-loop} and human labels.
The manually graded examples give $364$ memory-level classifications,
over datasets and memory systems; we observed no
systematic differences in judge behavior across datasets or systems, so the
counts are pooled. We attribute the high agreement to the design
of the dataset: since each entry is short, the classification is a simple instruction-following task for a frontier model.

\begin{table}[htbp]
\centering
\small
\setlength{\tabcolsep}{3.5pt}
\resizebox{\linewidth}{!}{%
\begin{tabular}{@{}lrrrrrr@{}}
\toprule
& \multicolumn{5}{c}{\textbf{Human label}} & \\
\cmidrule(lr){2-6}
\textbf{Judge label} & Corr. & N.st. & Summ. & N.ret. & Reas. & $n$ \\
\midrule
Correct (Corr.)         & $150$ & $0$  & $0$  & $0$   & $0$ & $150$ \\
Not stored (N.st.)      & $0$   & $54$ & $2$  & $0$   & $0$ & $56$ \\
Summary error (Summ.)   & $0$   & $2$  & $50$ & $2$   & $0$ & $54$ \\
Not retrieved (N.ret.)  & $0$   & $0$  & $0$  & $104$ & $0$ & $104$ \\
Reasoning error (Reas.) & $0$   & $0$  & $0$  & $0$   & $0$ & $0$ \\
\midrule
$n$ & $150$ & $56$ & $52$ & $106$ & $0$ & $364$ \\
\bottomrule
\end{tabular}}
\caption{Confusion matrix between the LLM judge and human annotations over
$364$ memory-level classifications, pooled across all datasets and memory
systems.}
\label{tab:judge-confusion}
\end{table}

\section{Detailed Evaluation Results}
\label{app:detailed-results}

We provide a complete summary of the evaluation results for all models, datasets, and memory systems for completeness and verification. While the critical insights are in the main paper, these figures further validate our claims.

\begin{figure*}
    \centering

    \begin{subfigure}{0.48\linewidth}
        \centering
        \includegraphics[width=\linewidth]{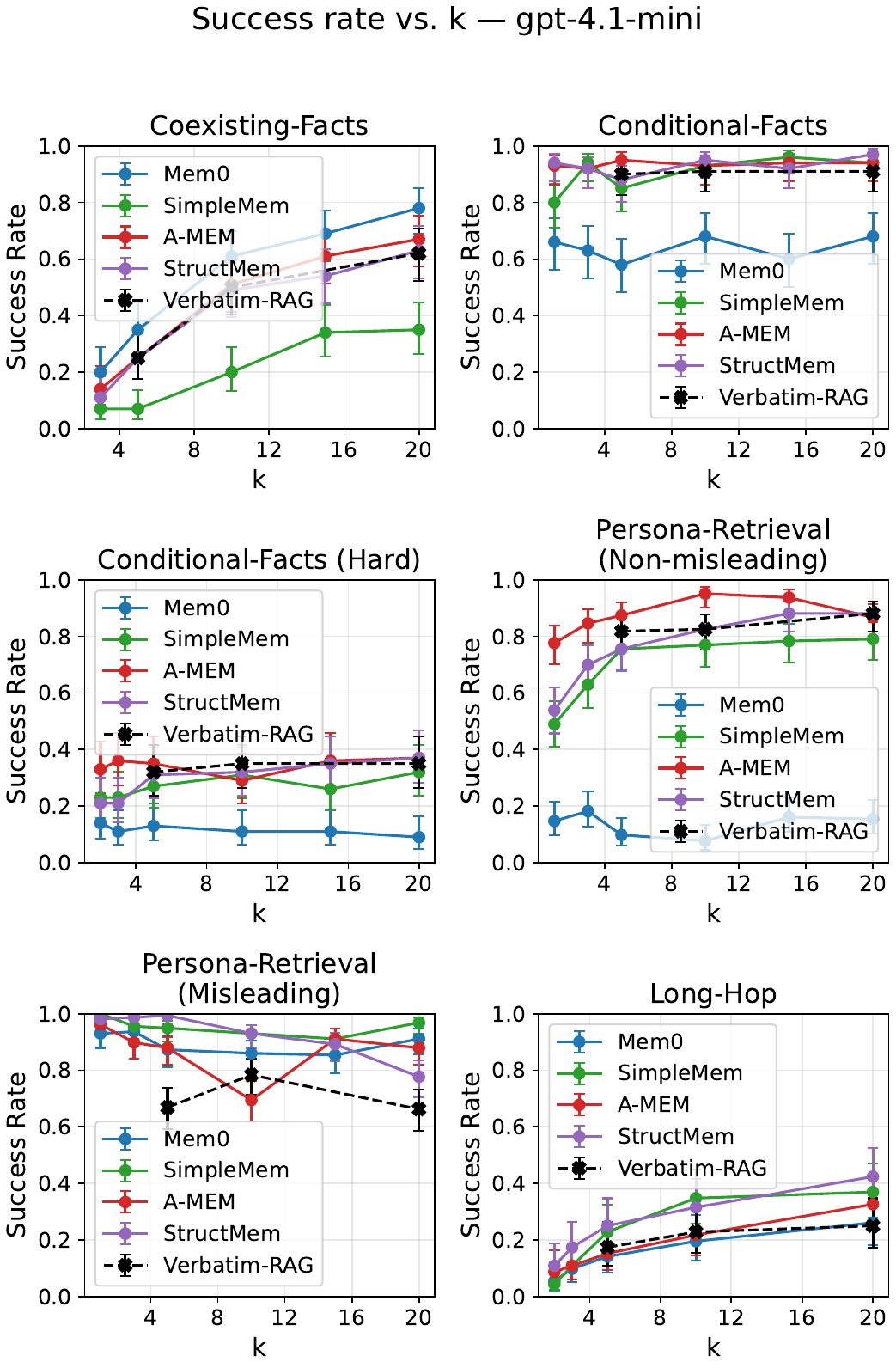}
    \end{subfigure}
    \hfill
    \begin{subfigure}{0.48\linewidth}
        \centering
        \includegraphics[width=\linewidth]{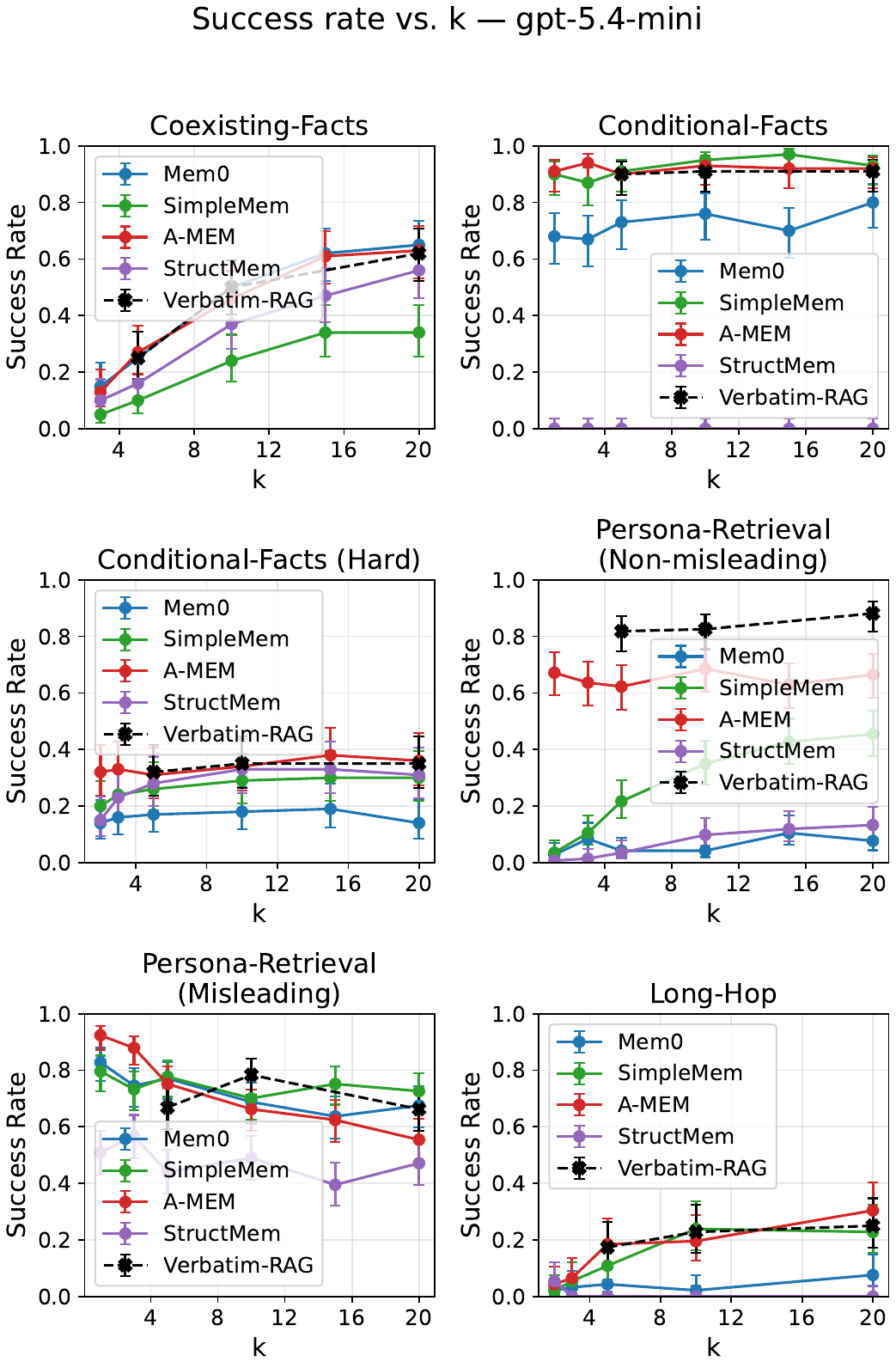}
    \end{subfigure}

    \vspace{0.3cm}

    \begin{subfigure}{0.48\linewidth}
        \centering
        \includegraphics[width=\linewidth]{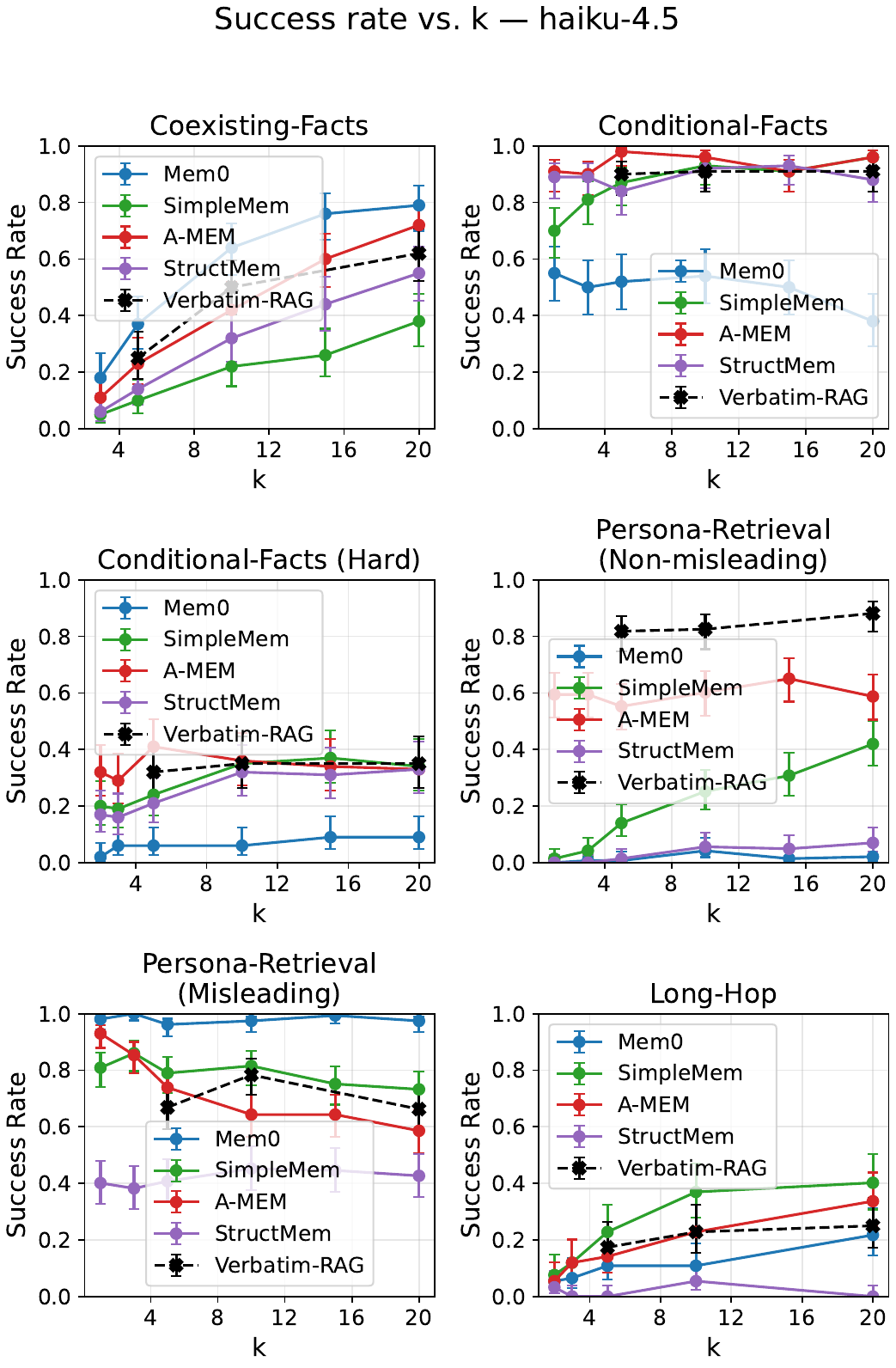}
    \end{subfigure}
    \hfill
    \begin{subfigure}{0.48\linewidth}
        \centering
        \includegraphics[width=\linewidth]{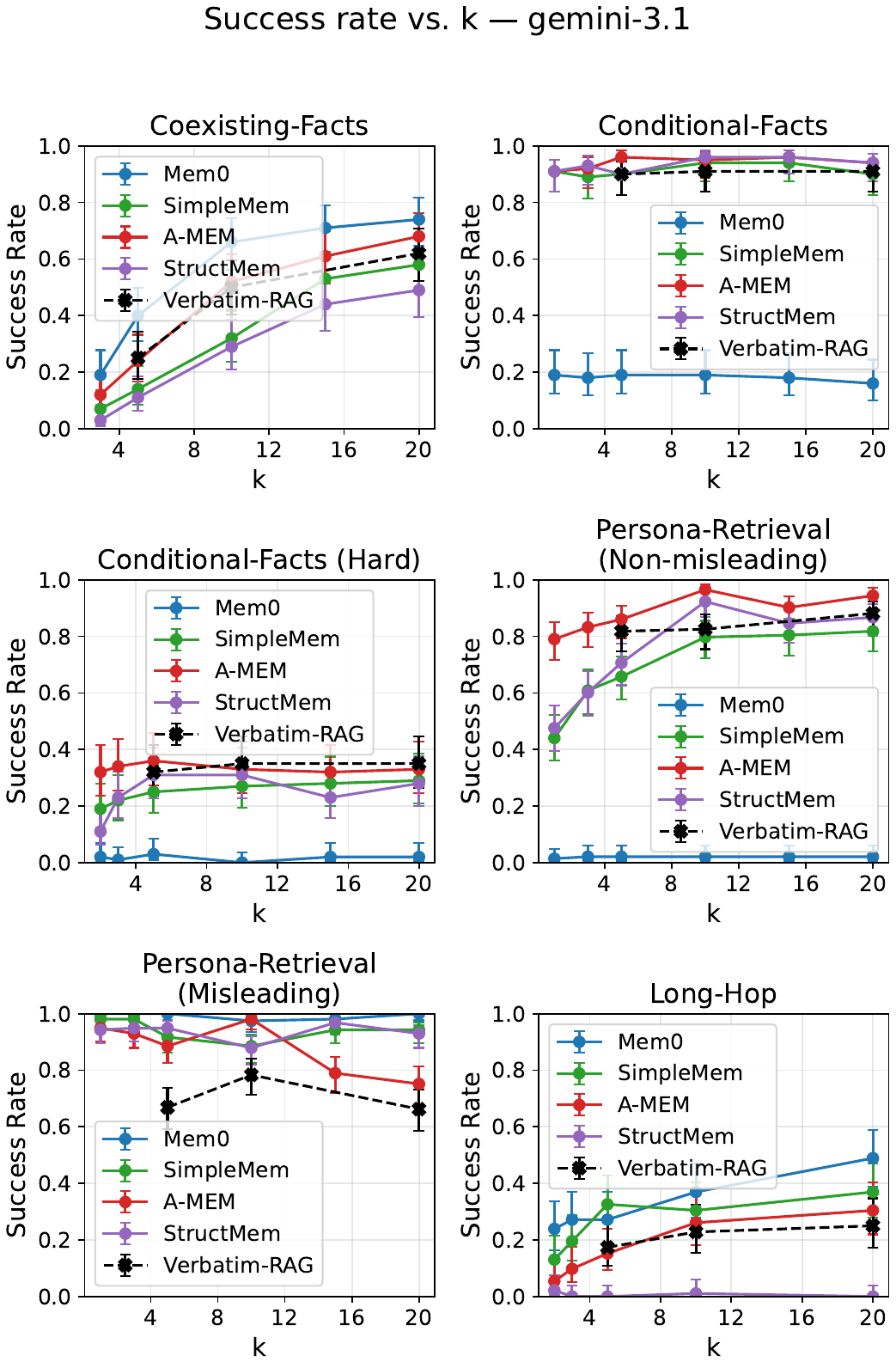}
    \end{subfigure}

    \caption{Success rates for all datasets, models, and systems. The \textbf{Verbatim-RAG} baseline (black dashed) is shown in every panel for reference; it uses no internal LLM, so the same curve appears in all four figures.}
    \label{fig:app-perf-v-k}
\end{figure*}

\begin{figure*}
    \centering

    \begin{subfigure}{0.48\linewidth}
        \centering
        \includegraphics[width=\linewidth]{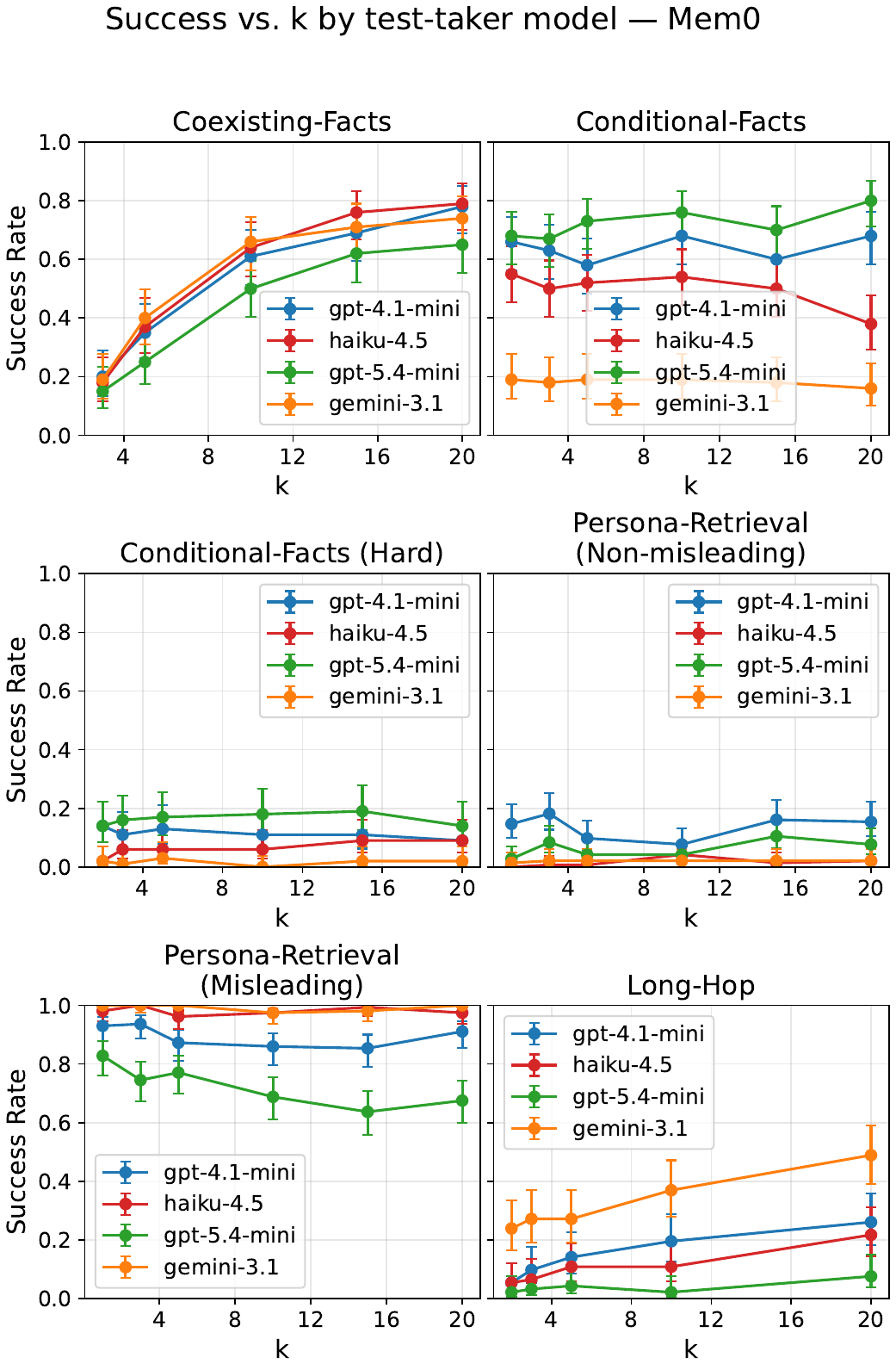}
    \end{subfigure}
    \hfill
    \begin{subfigure}{0.48\linewidth}
        \centering
        \includegraphics[width=\linewidth]{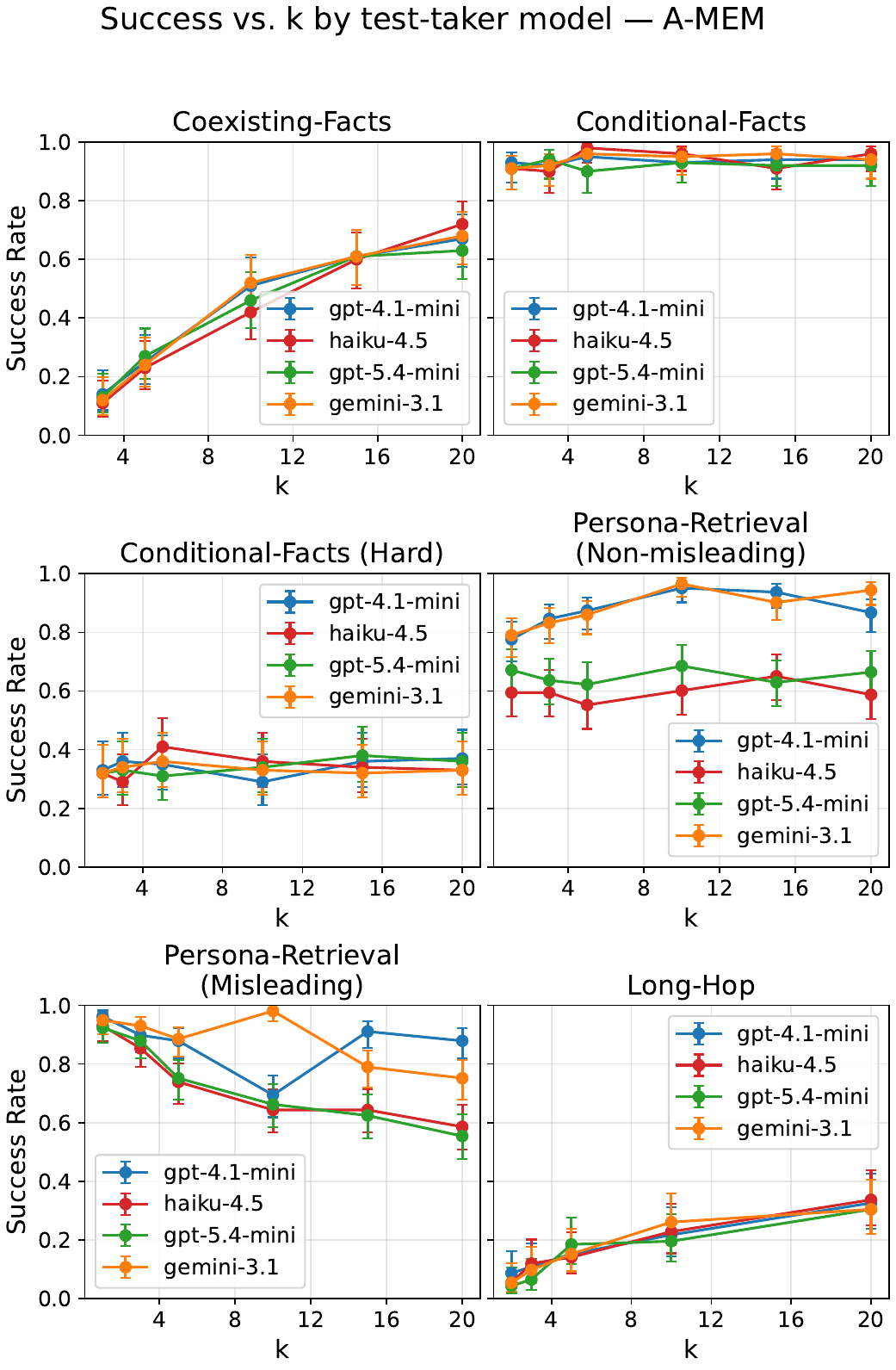}
    \end{subfigure}

    \vspace{0.3cm}

    \begin{subfigure}{0.48\linewidth}
        \centering
        \includegraphics[width=\linewidth]{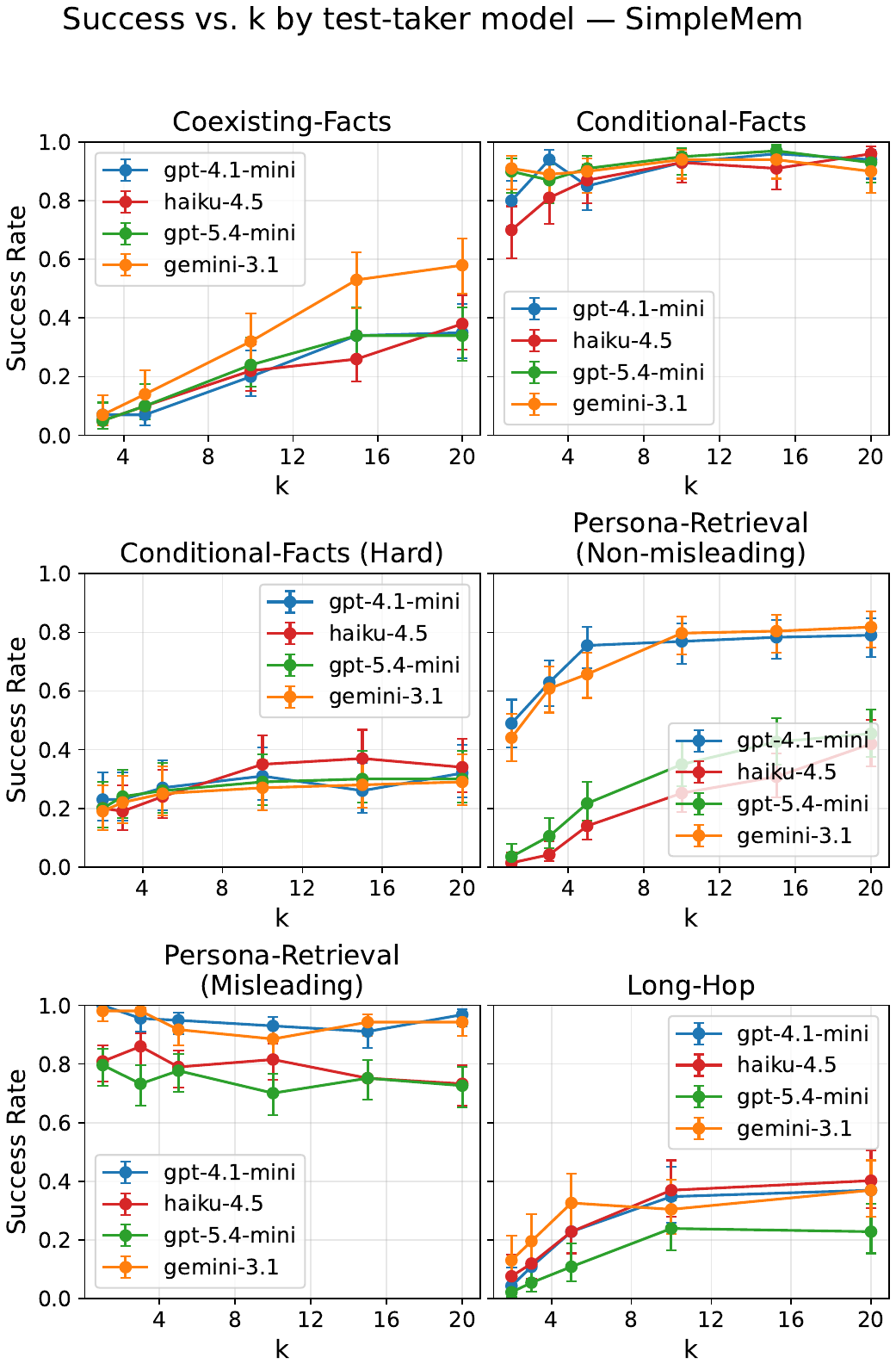}
    \end{subfigure}
    \hfill
    \begin{subfigure}{0.48\linewidth}
        \centering
        \includegraphics[width=\linewidth]{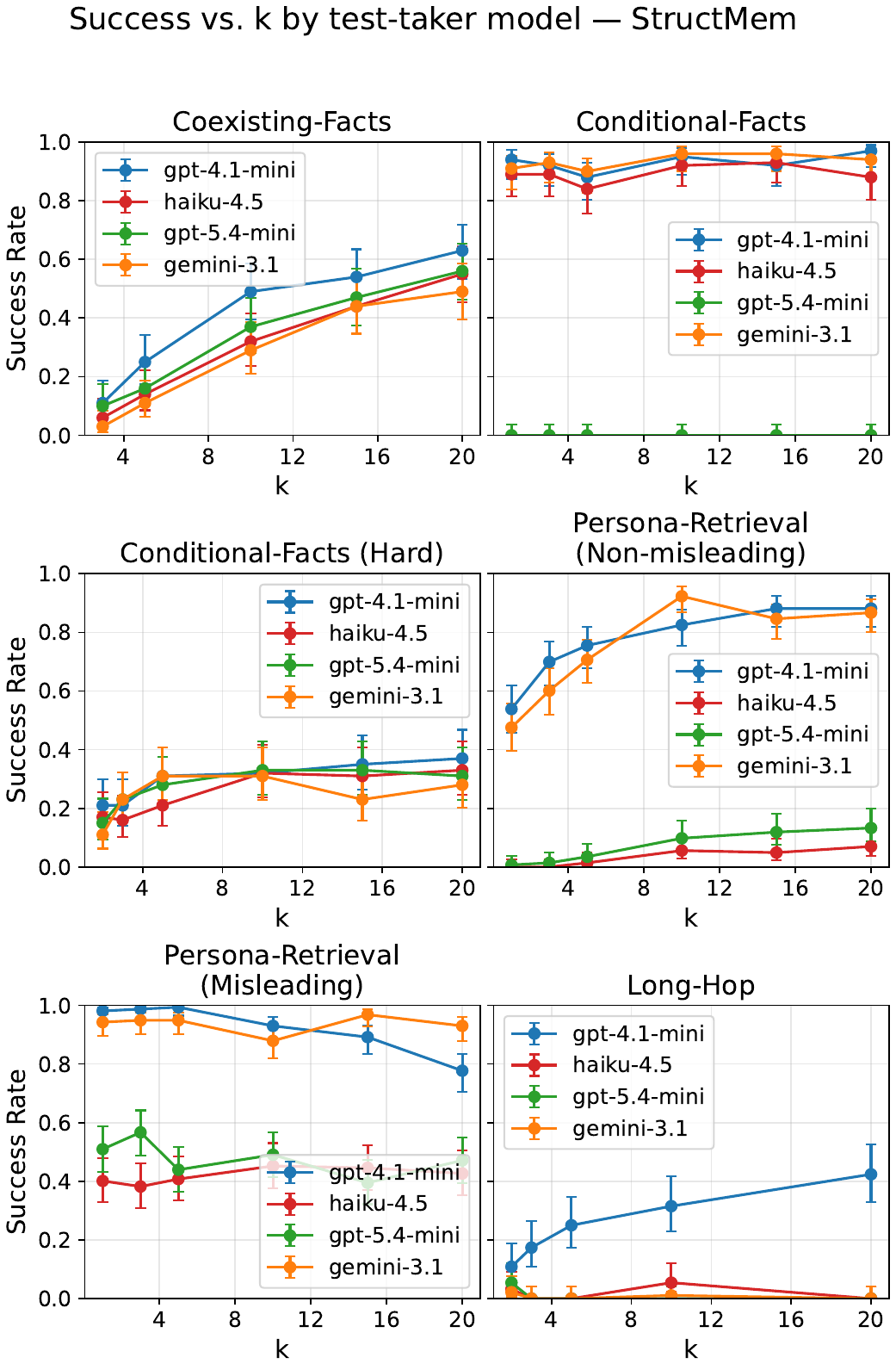}
    \end{subfigure}

    \caption{Success rates on each dataset for every model.}
    \label{fig:app-perf-v-model}
\end{figure*}

\begin{figure*}
    \centering
    \includegraphics[width=0.9\linewidth]{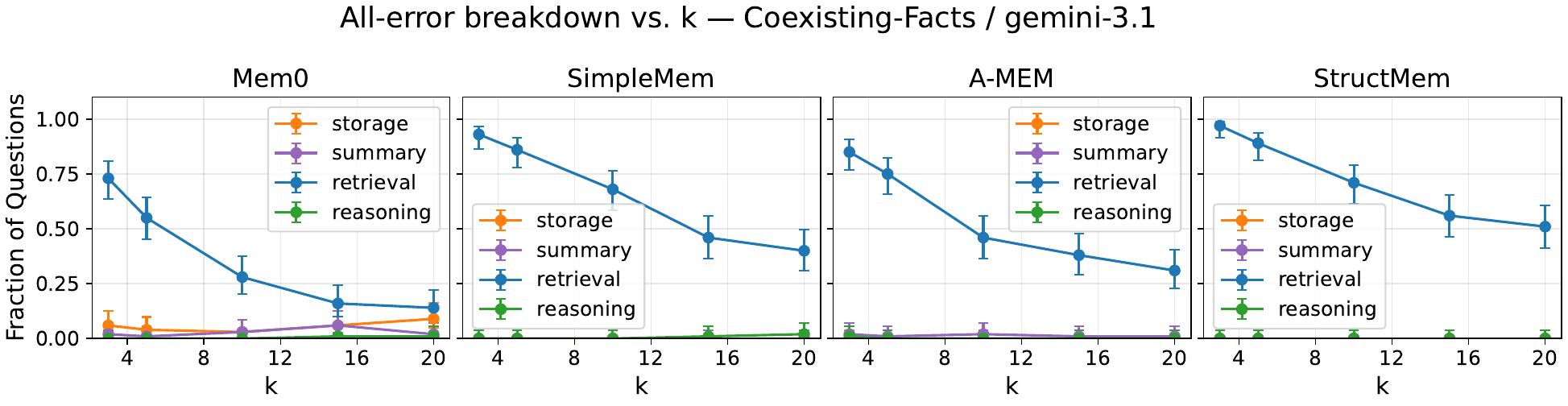}
    \includegraphics[width=0.9\linewidth]{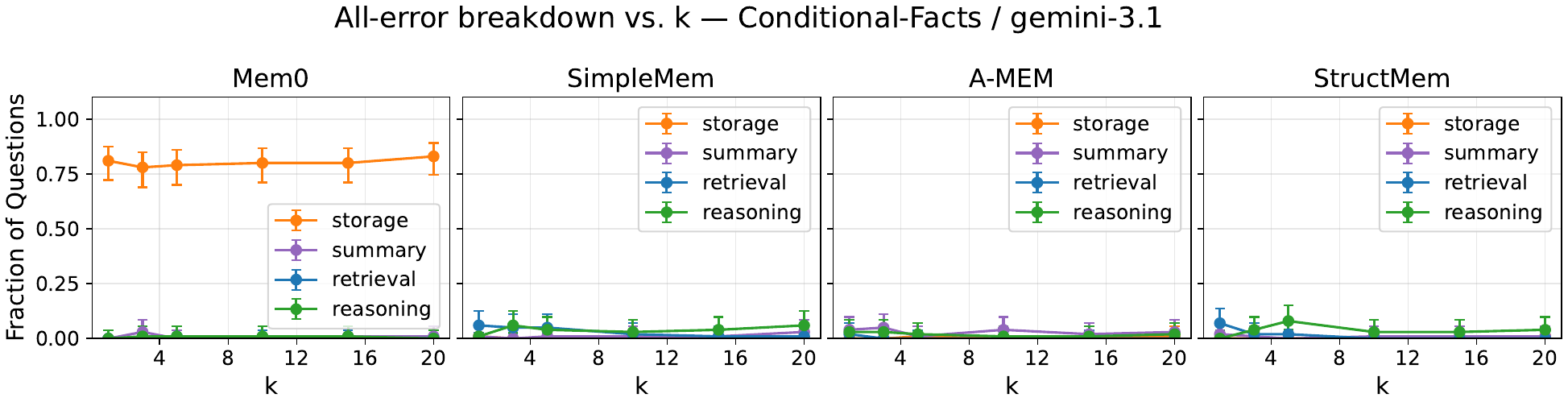}
    \includegraphics[width=0.9\linewidth]{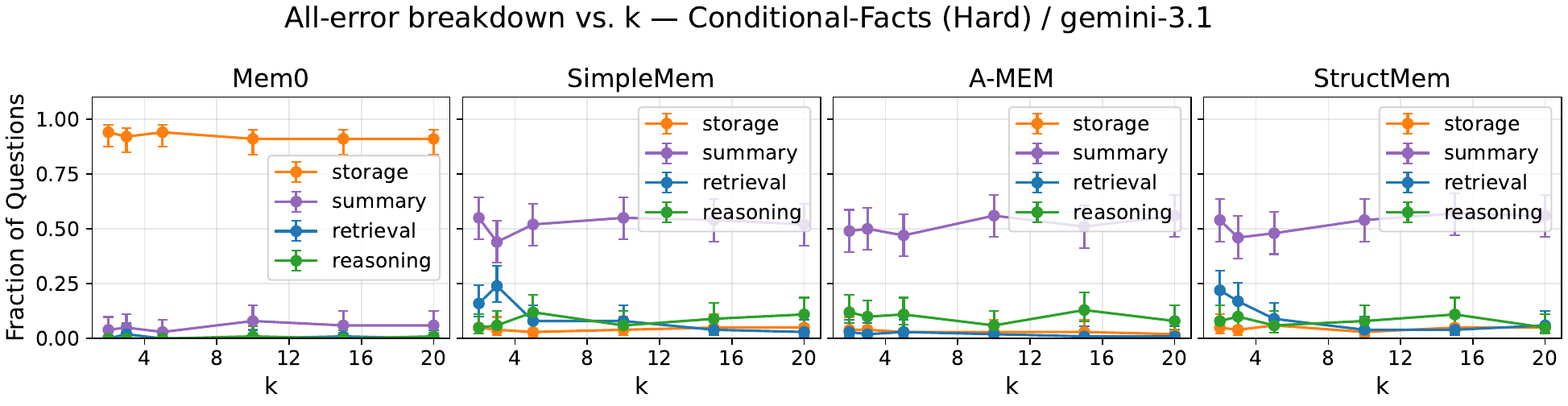}
    \includegraphics[width=0.9\linewidth]{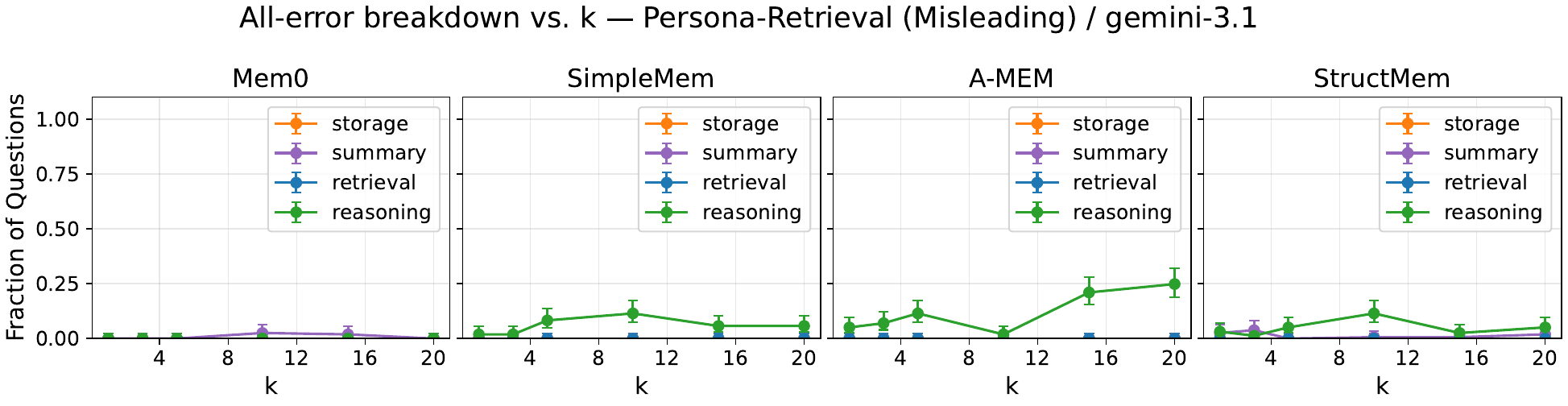}
    \includegraphics[width=0.9\linewidth]{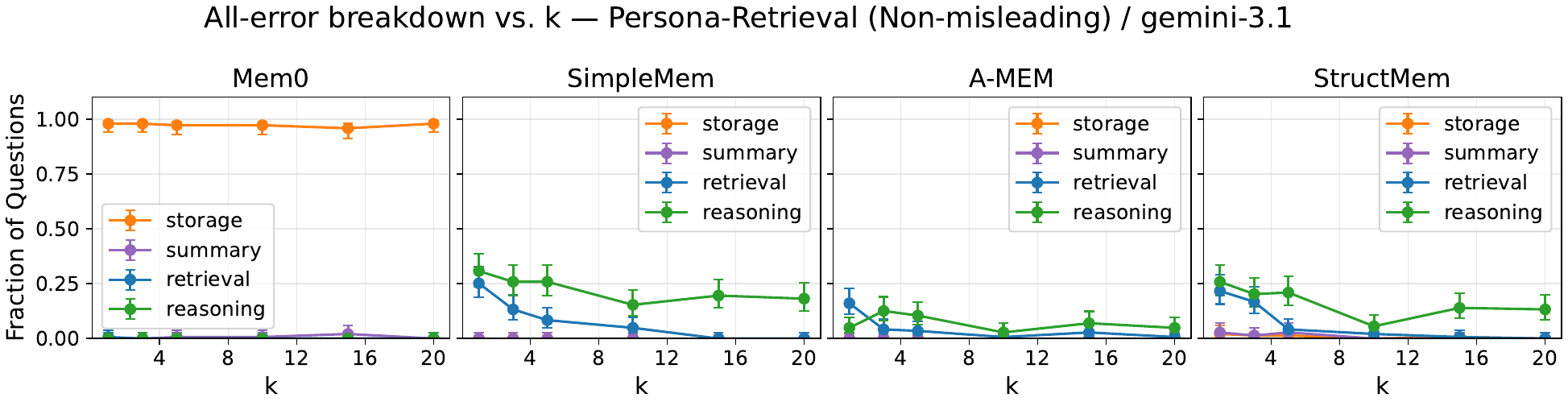}
    \includegraphics[width=0.9\linewidth]{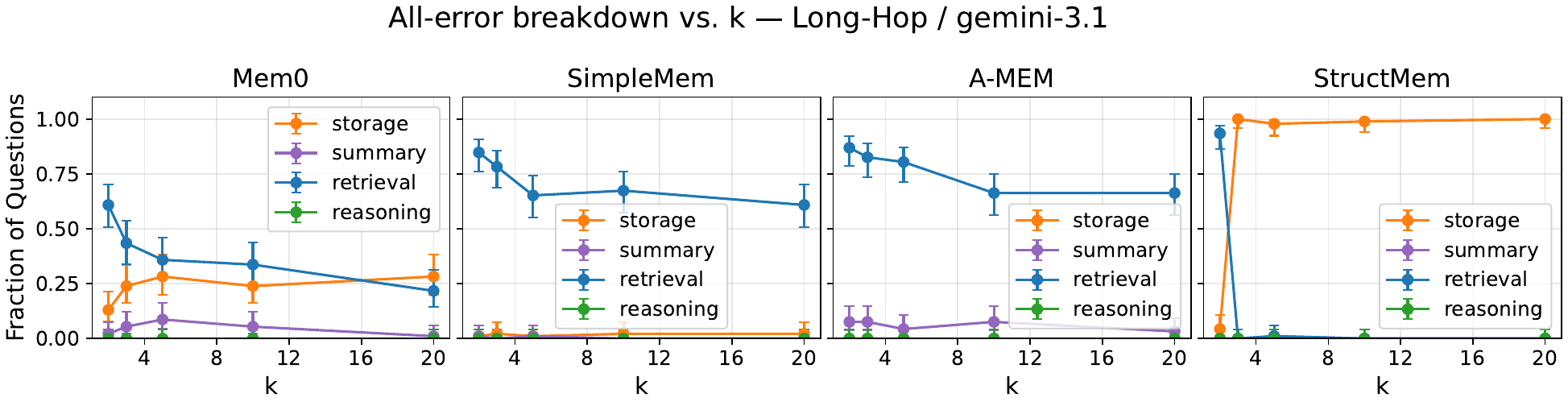}

    \caption{All error classifications for all datasets and systems, using Gemini-3.1, including reasoning errors.}
    \label{fig:app-gemini-err}
\end{figure*}

\begin{figure*}
    \centering
    \includegraphics[width=0.9\linewidth]{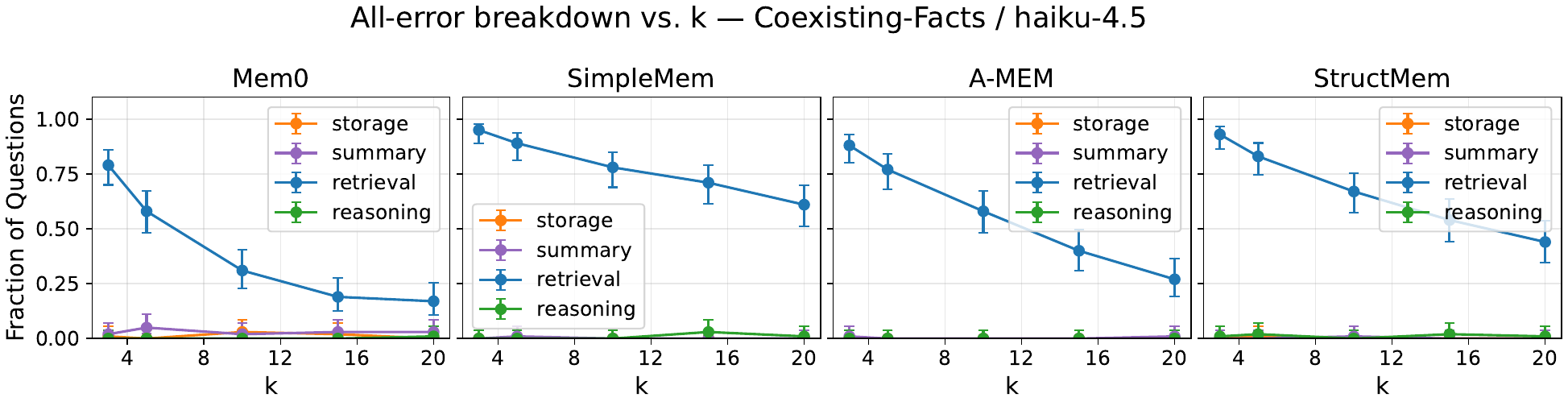}
    \includegraphics[width=0.9\linewidth]{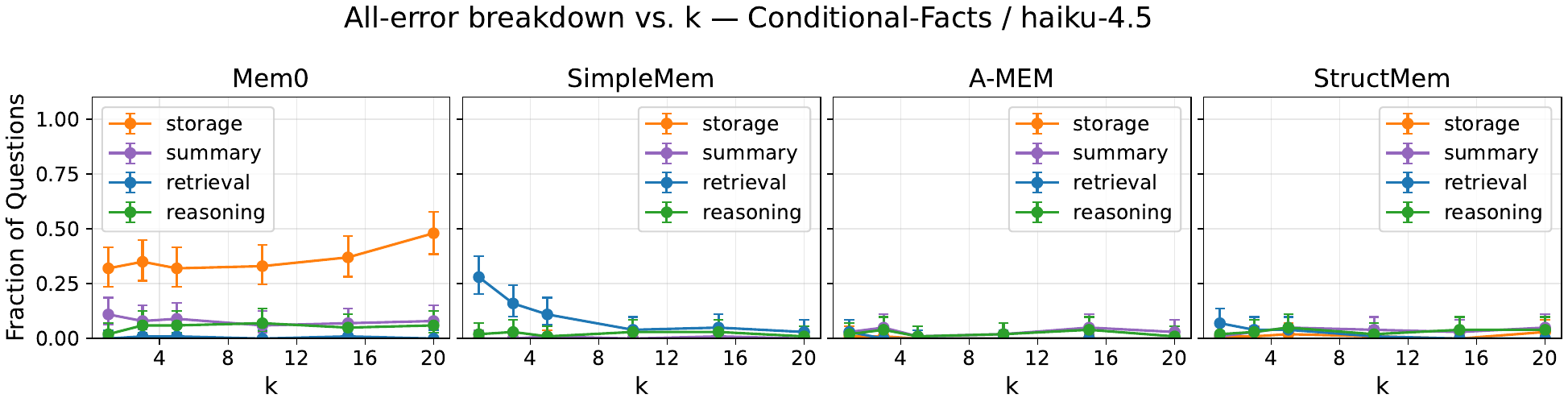}
    \includegraphics[width=0.9\linewidth]{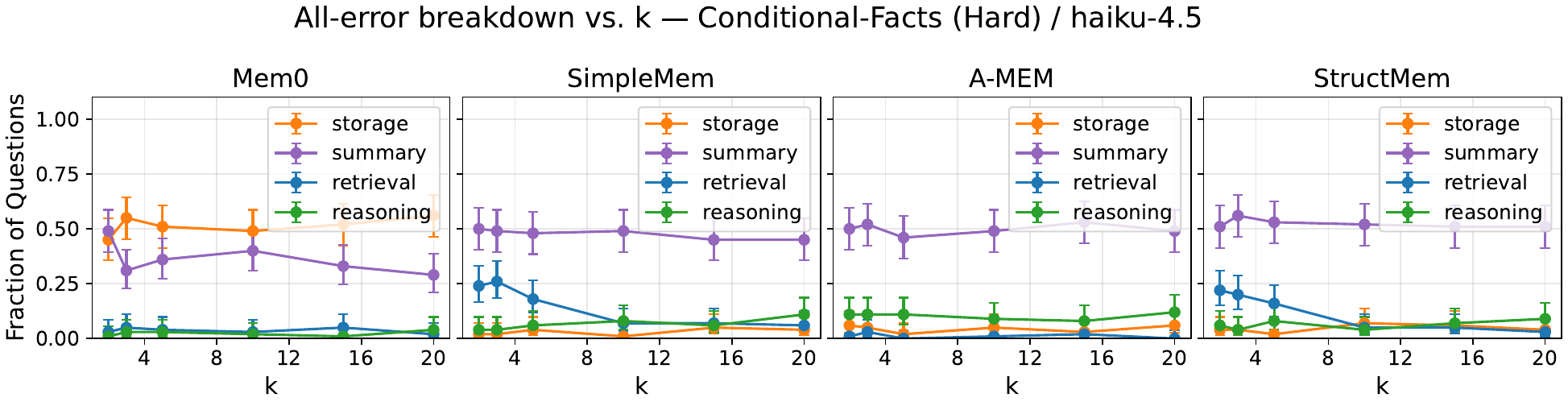}
    \includegraphics[width=0.9\linewidth]{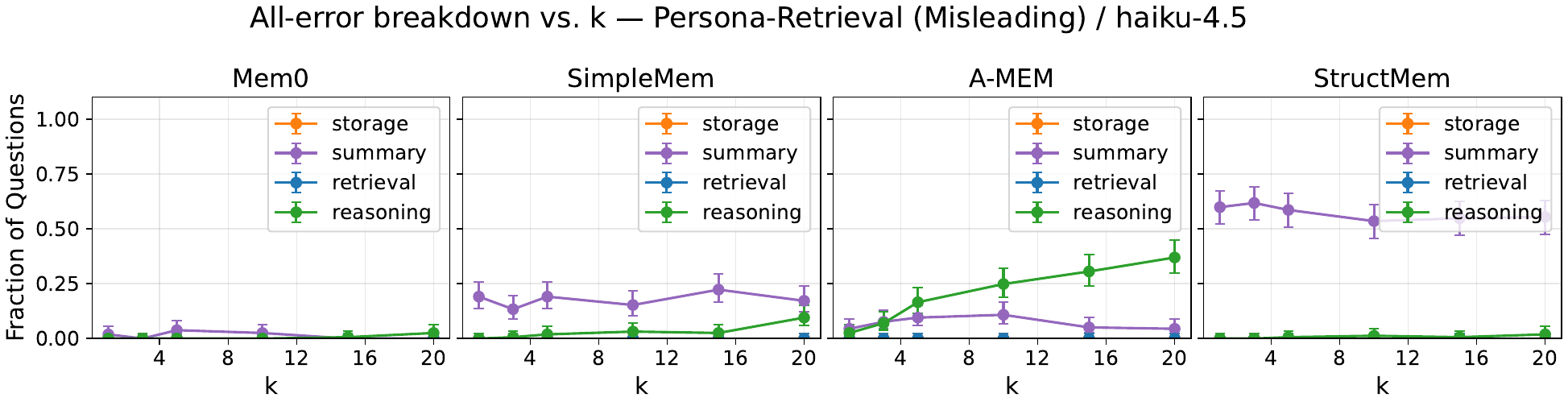}
    \includegraphics[width=0.9\linewidth]{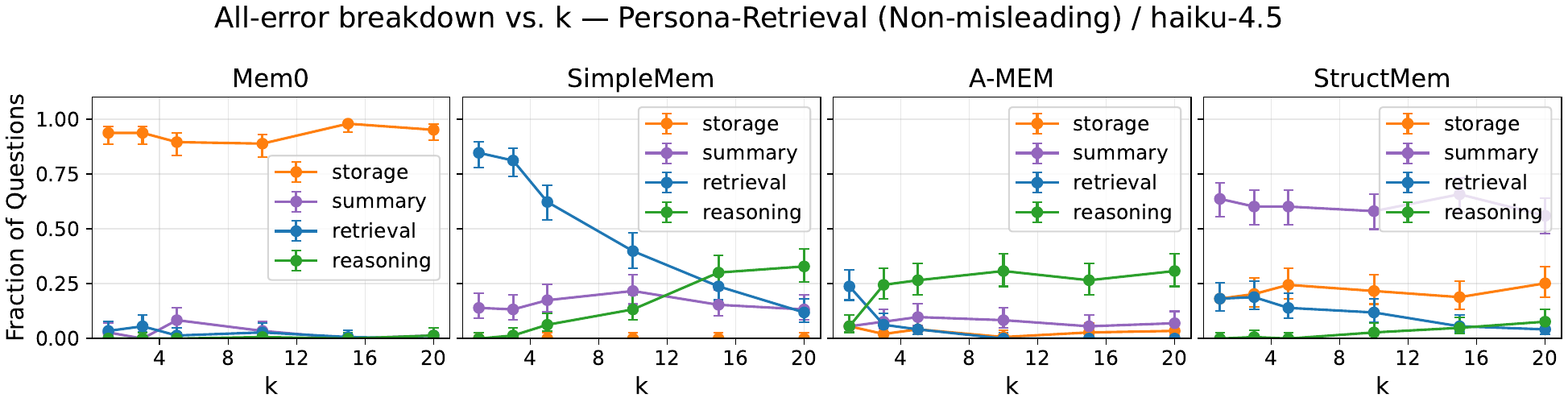}
    \includegraphics[width=0.9\linewidth]{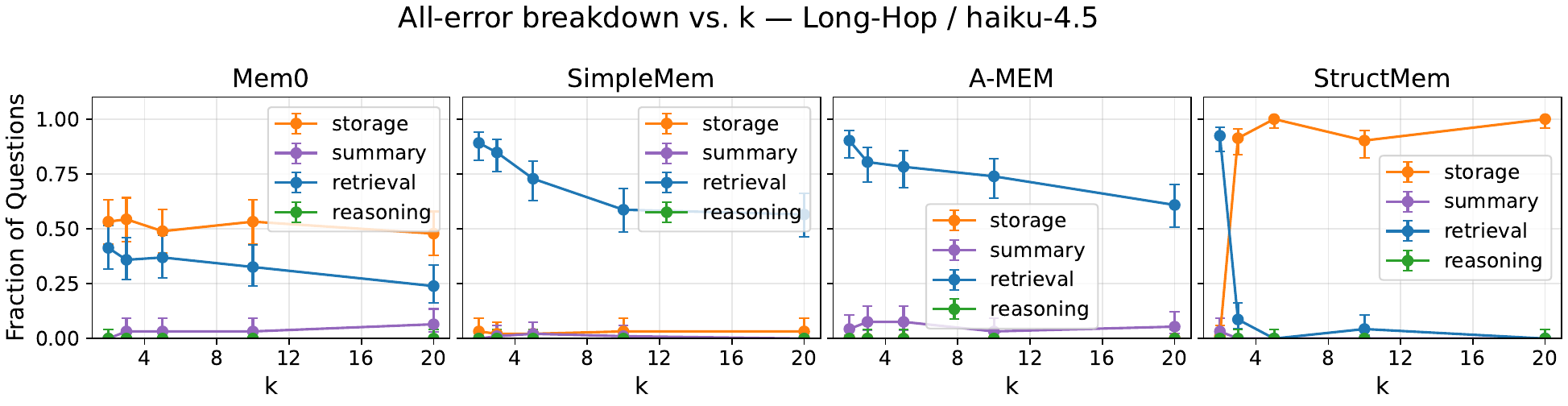}

    \caption{All error classifications for all datasets and systems, using Haiku-4.5, including reasoning errors.}
    \label{fig:app-haiku-err}

\end{figure*}

\begin{figure*}
    \centering
    \includegraphics[width=0.9\linewidth]{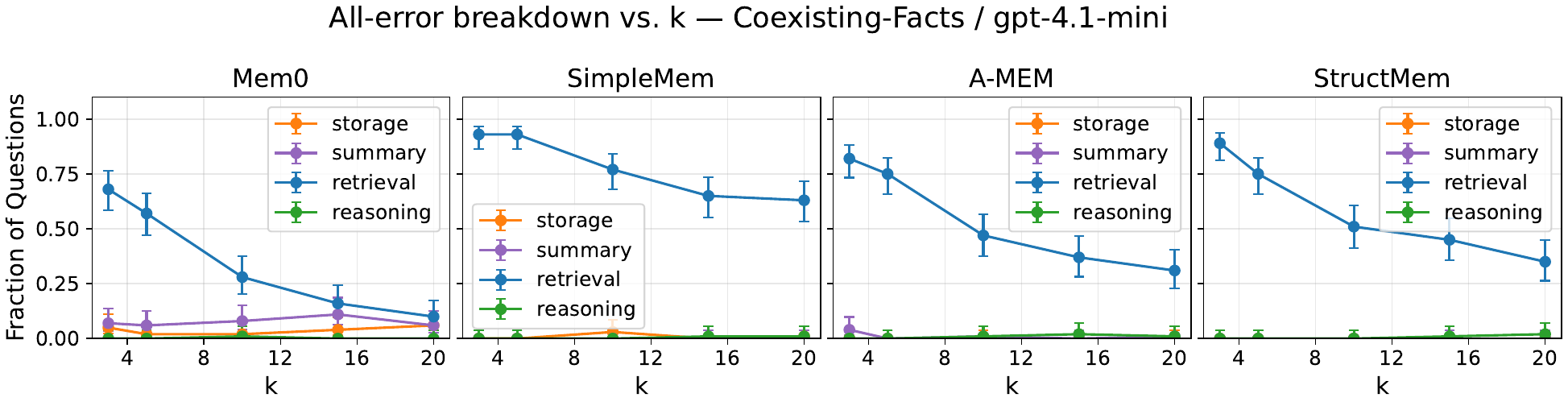}
    \includegraphics[width=0.9\linewidth]{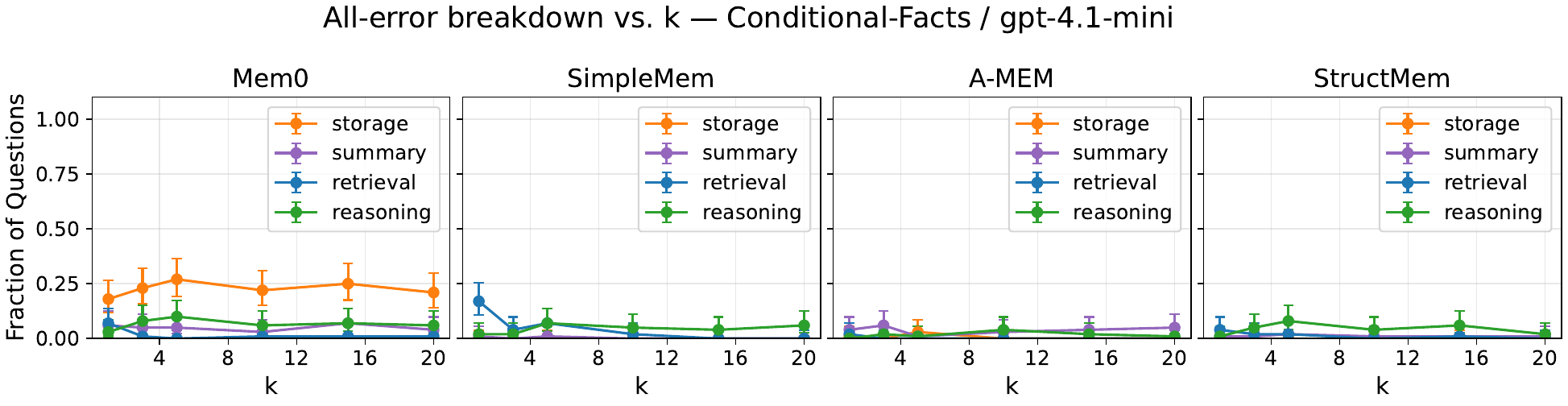}
    \includegraphics[width=0.9\linewidth]{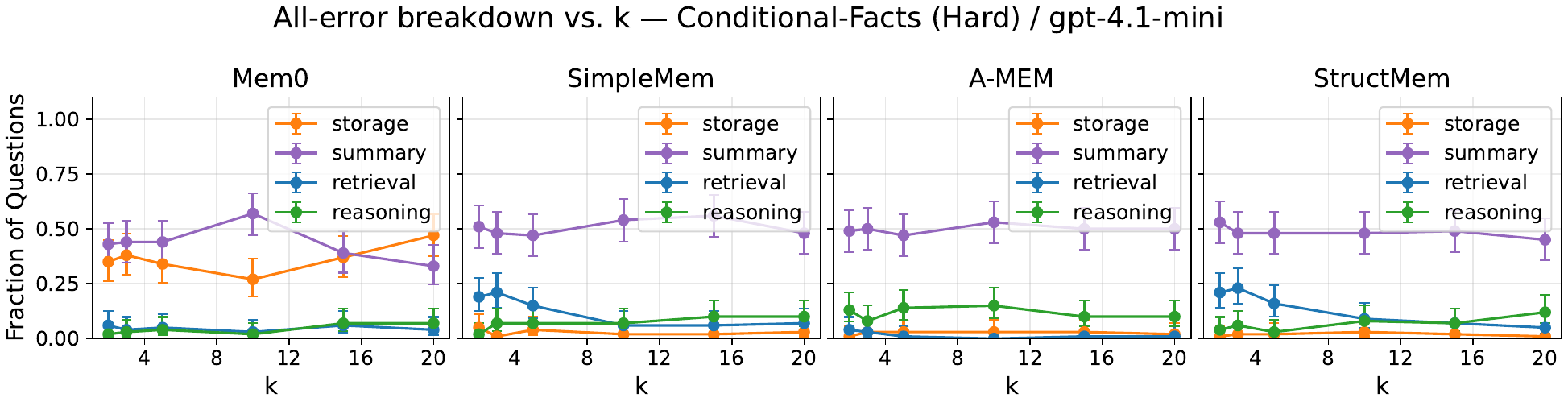}
    \includegraphics[width=0.9\linewidth]{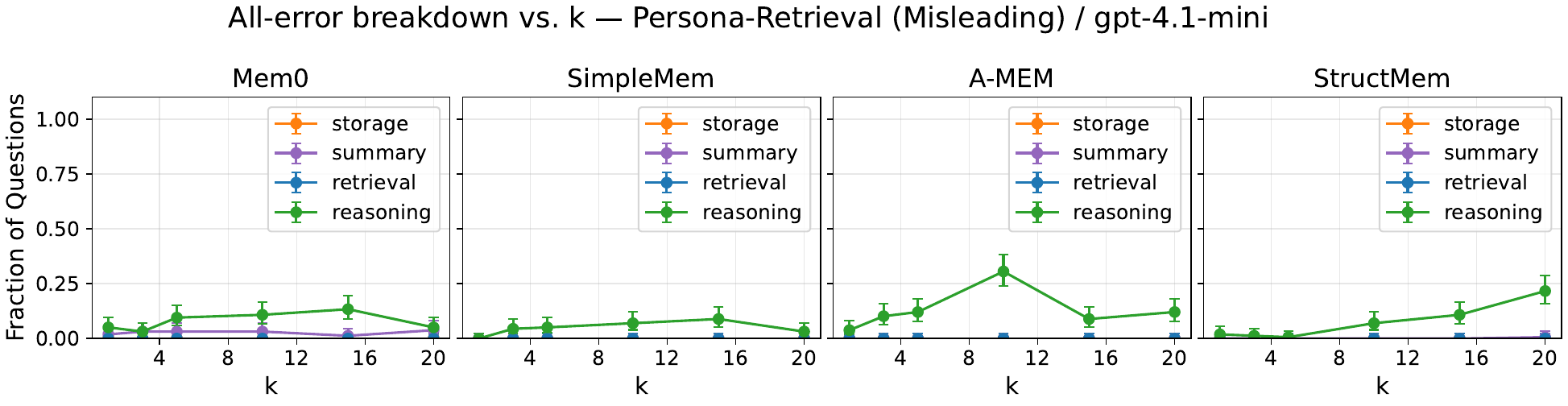}
    \includegraphics[width=0.9\linewidth]{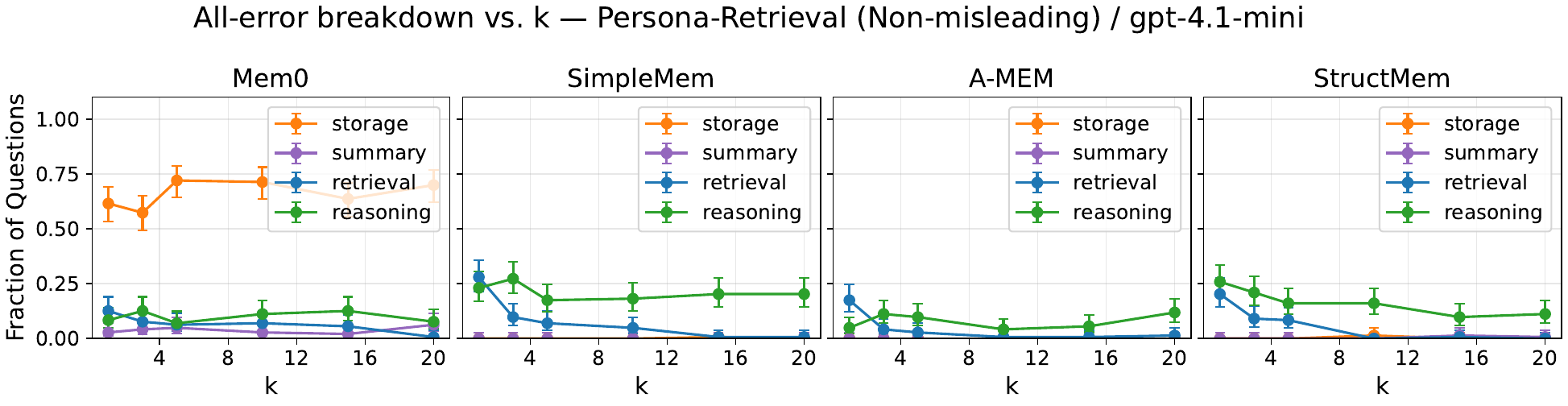}
    \includegraphics[width=0.9\linewidth]{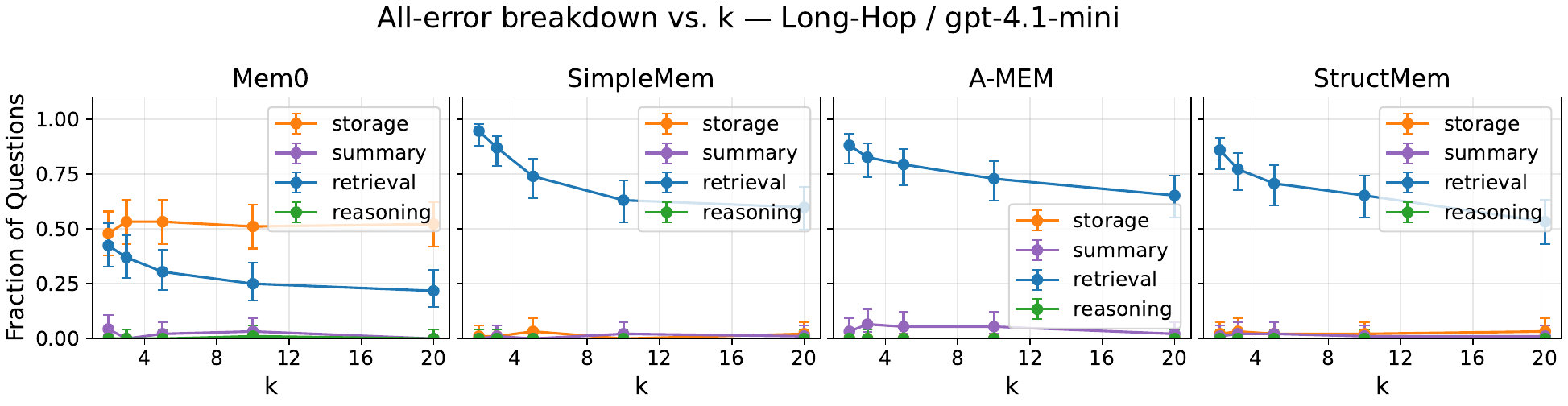}
    \caption{All error classifications for all datasets and systems, using GPT-4.1-mini, including reasoning errors.}
    \label{fig:app-gpt41-err}

\end{figure*}

\begin{figure*}
    \centering
    \includegraphics[width=0.9\linewidth]{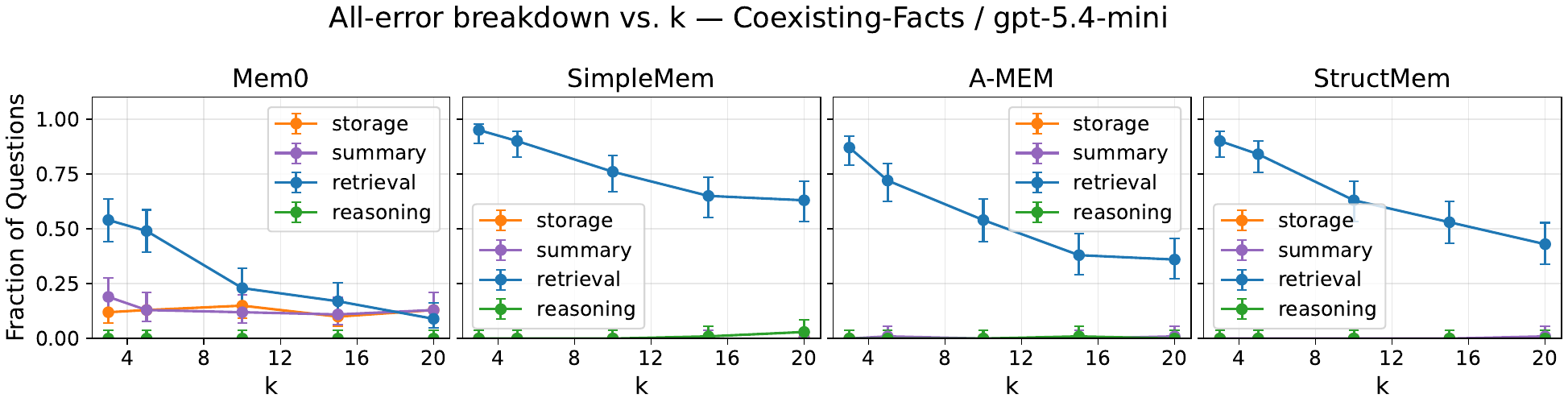}
    \includegraphics[width=0.9\linewidth]{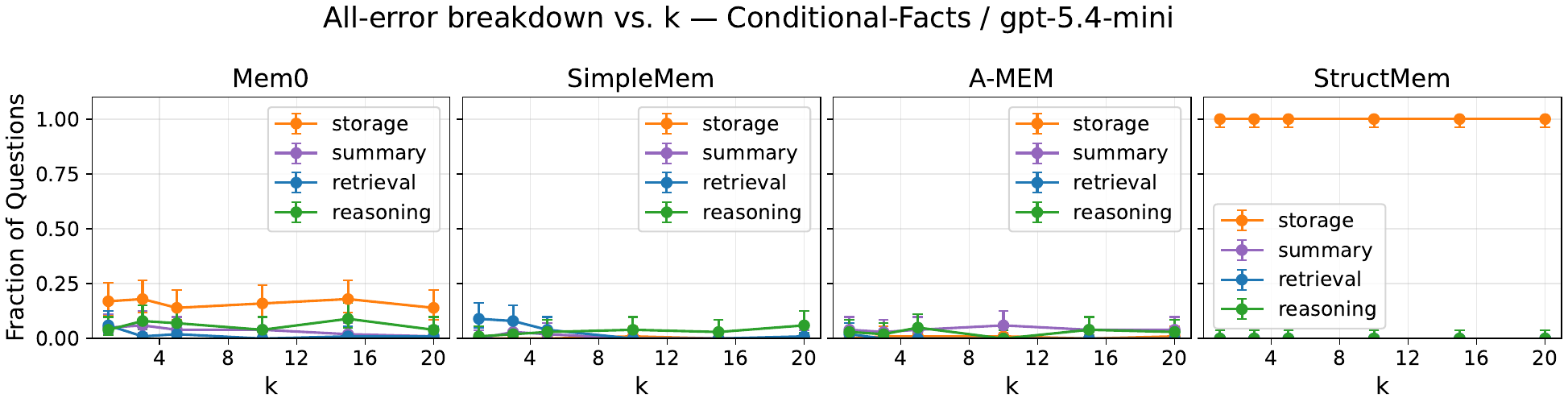}
    \includegraphics[width=0.9\linewidth]{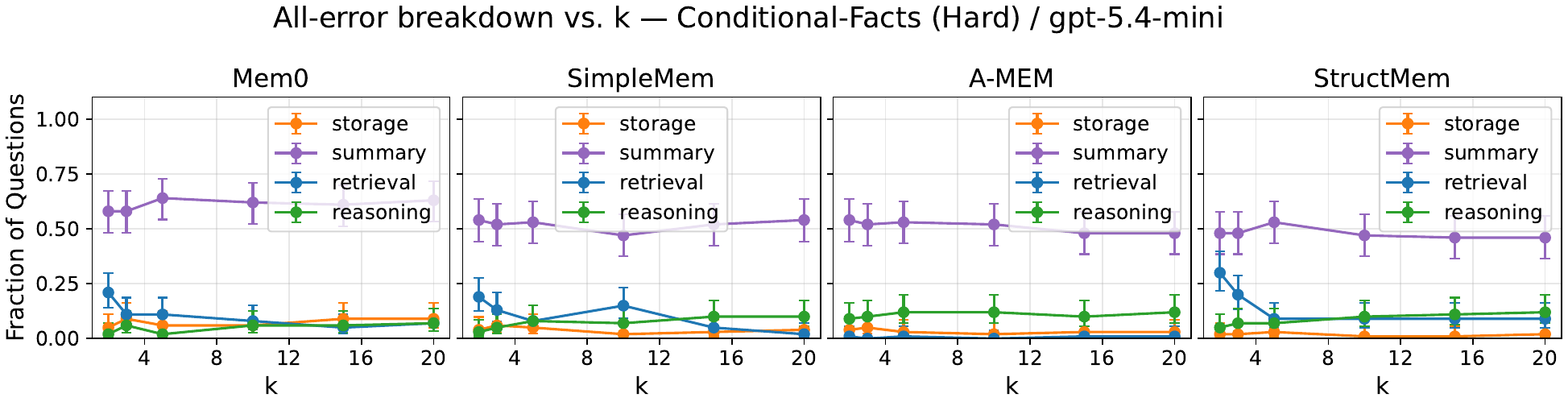}
    \includegraphics[width=0.9\linewidth]{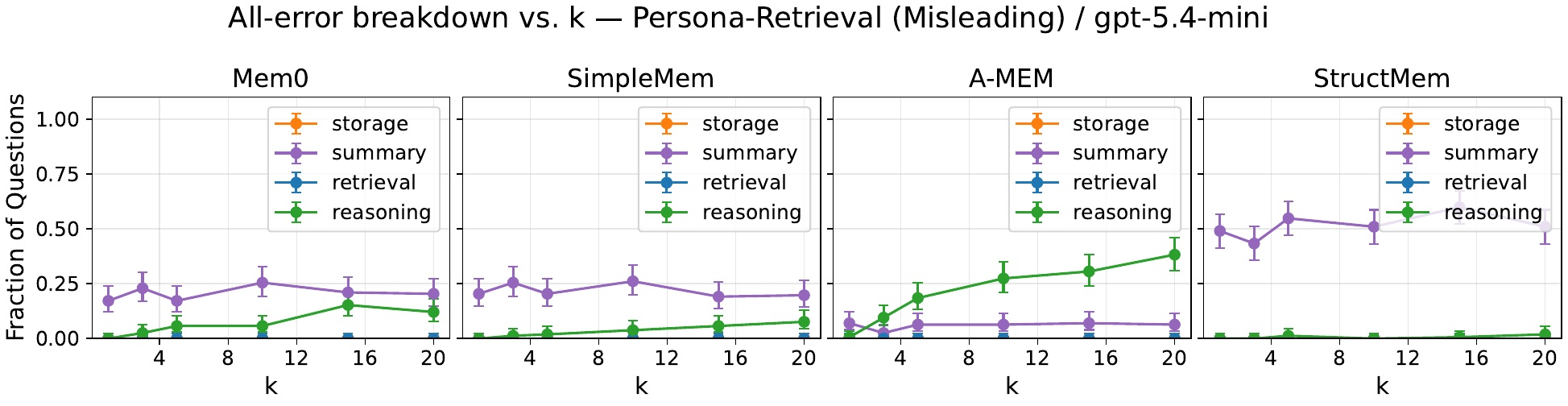}
    \includegraphics[width=0.9\linewidth]{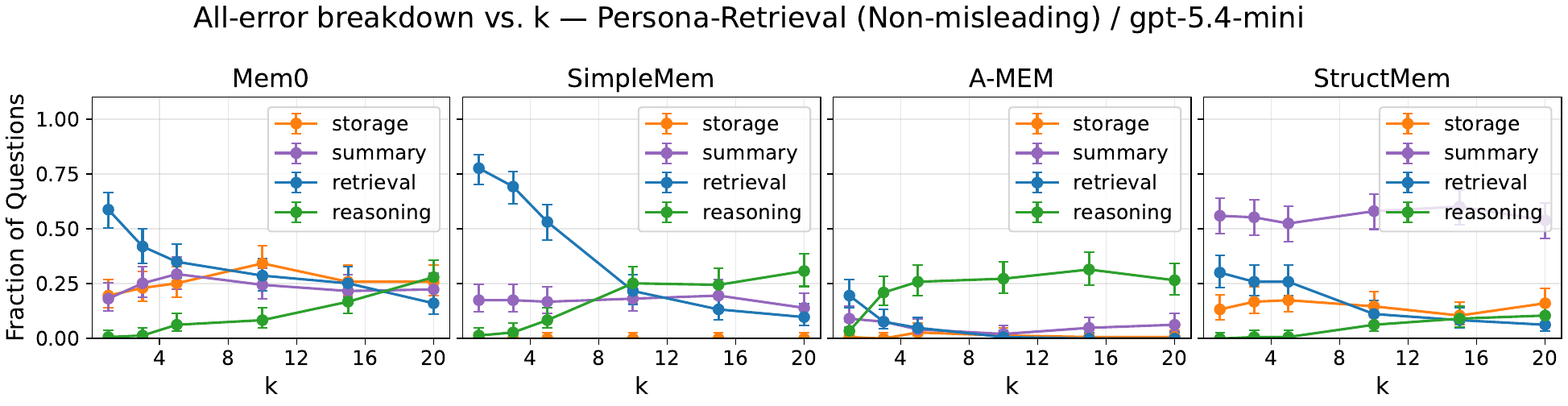}
    \includegraphics[width=0.9\linewidth]{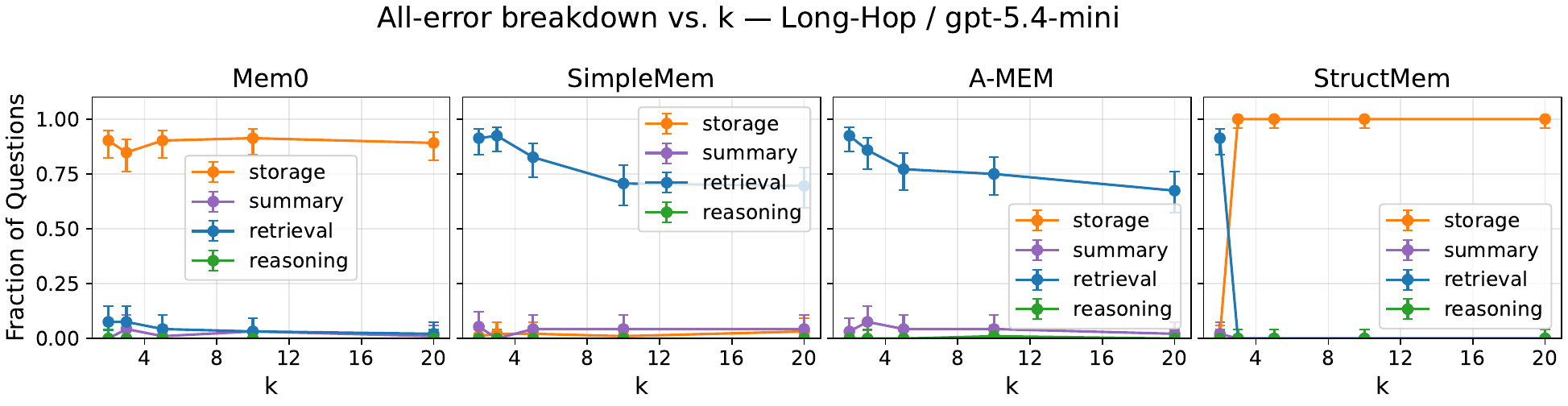}
    \caption{All error classifications for all datasets and systems, using GPT-5.4-mini, including reasoning errors.}
    \label{fig:app-gpt54-err}

\end{figure*}

Figure~\ref{fig:app-perf-v-k} shows success rates as a function of $k$, the number of retrieved memories.
Figure~\ref{fig:app-perf-v-model} shows how performance on \textsc{MemFail} changes across different models.
Figures~\ref{fig:app-gemini-err}, \ref{fig:app-haiku-err}, \ref{fig:app-gpt41-err}, and~\ref{fig:app-gpt54-err} show the per-system error-type breakdowns for each test-taker model.

% =====================================================================
% Appendix: Full Prompt Listings
% =====================================================================
% Preamble requirements: tcolorbox + listings + a `promptbox`
% environment. A reasonable definition (adapt to taste):
%
%   \usepackage{tcolorbox}
%   \usepackage{listings}
%   \tcbuselibrary{breakable, skins}
%
%   \lstdefinestyle{prompt}{
%     basicstyle=\ttfamily\footnotesize,
%     breaklines=true, breakatwhitespace=true,
%     breakindent=0pt, postbreak=\mbox{\textcolor{gray}{$\hookrightarrow$}\space},
%     columns=fullflexible, keepspaces=true,
%     frame=none, aboveskip=0pt, belowskip=0pt,
%     showstringspaces=false,
%   }
%   \lstset{style=prompt}
%
%   \DeclareTColorBox{promptbox}{O{}}{
%     enhanced, breakable,
%     colback=gray!5, colframe=gray!50!black,
%     coltitle=white, colbacktitle=gray!60!black,
%     title=#1, fonttitle=\bfseries\small,
%     left=2mm, right=2mm, top=1mm, bottom=1mm,
%   }
%
% Each prompt below is reproduced verbatim from the corresponding
% generator / evaluator / judge module. Format-string placeholders
% ({entity}, {original_fact}, {all_memories_formatted}, etc.) are shown
% as they appear in source; the surrounding code substitutes them at
% call time.

\section{Full Prompt Listings}
\label{app:prompts}

This appendix reproduces every prompt used in \textsc{MemFail} verbatim:
the dataset-generation prompts (\ref{app:gen-prompts}), the prompts that
are sent to the memory-augmented evaluation model at evaluation time
(\ref{app:eval-prompts}), and the LLM-judge prompts used to grade and
classify errors (\ref{app:grade-prompts}).

% =====================================================================
\subsection{Generation prompts}
\label{app:gen-prompts}

% ---------------------------------------------------------------------
\subsubsection{\textit{Conditional-Facts}}

The Easy variant calls a single datapoint generator (Prompt~\ref{prompt:cond-gen})
followed by an essay wrapper that stitches the conditional fact into a short
casual essay (Prompt~\ref{prompt:cond-essay-easy}). The Hard variant reuses the
same datapoint generator and replaces the essay wrapper with a stricter
decomposition prompt (Prompt~\ref{prompt:cond-essay-hard}).

\begin{promptbox}[Generation: \textit{Conditional-Facts} datapoint generator]
\label{prompt:cond-gen}
\begin{lstlisting}
Generate conditional-facts datapoints. Each datapoint describes an entity (person, pet, or character) who has a CONDITIONAL behavior -- they only do something under a specific condition.

For each spec below, generate:

1. "entity": a realistic first name (for persons/characters) or a pet name (for pets)
   e.g. "Jordan", "Miso", "Captain Rex"

2. "behavior": a short action phrase describing what the entity does conditionally.
   Invent something creative and specific to the entity and condition type -- do NOT default
   to cliches like "goes for a run" or "drinks coffee". The behavior should feel personal
   and idiosyncratic, not universally common.
   The examples below are illustrative ONLY -- do not reuse them:
   e.g. "re-reads old letters", "sketches floor plans", "hums while doing dishes",
        "sends voice notes instead of texts", "reorganizes their bookshelf"

3. "condition": the specific condition under which the behavior occurs
   e.g. "after 5pm", "when it's raining", "when feeling stressed"
   - Must be concrete and testable -- the question will present a specific context
   - Avoid vague conditions like "sometimes" or "often"

4. "entity_facts": a list containing exactly 1 natural statement that directly encodes
   the full conditional fact -- both the behavior AND the condition in a single sentence.
   - Must be a casual, first-person or third-person conversational sentence
   - Must clearly state BOTH what the entity does AND when/under what condition
   - 1-2 sentences max
   - The example below is illustrative ONLY -- do not reuse it:
     ["Alex has a rule: no coffee before 5pm, since it messes with their sleep."]

5. "question": a natural question about whether the entity should do (or would do) the behavior,
   given a SPECIFIC context that may or may not satisfy the condition.

   CRITICAL RULE -- the question MUST be non-inferrable without the entity's specific fact:
   A person with no knowledge of the entity should NOT be able to guess the correct answer
   from common sense or general norms alone. The correct answer must depend on knowing
   THIS entity's specific conditional rule.

   BAD (inferrable from common sense):
     - "It's a bright sunny afternoon. Should Zarek wear his heavy winter cloak?"
       -> Anyone would say no, regardless of any stored fact.
     - "Jordan hasn't slept in 30 hours. Would they want to go clubbing?"
       -> Common sense gives the answer.

   GOOD (requires knowing the entity's rule) -- these examples are illustrative ONLY, do not reuse them:
     - "It's 3pm and I'm meeting Alex -- should I grab them a coffee?"
       -> Without knowing Alex's after-5pm rule, you might reasonably say yes.
     - "It's a quiet Sunday morning. Would Priya want to reorganize her bookshelf?"
       -> Without knowing Priya only does this when stressed, you can't tell.
     - "We're at the park and it's 18C outside. Would Miso eat from the red bowl?"
       -> Without knowing Miso's specific rule, this is genuinely ambiguous.

   The question should present a context where a reasonable person WITHOUT the entity's
   specific rule could plausibly answer either way -- making the stored fact decisive.

6. "question_context": the specific context presented in the question
   e.g. "3pm", "quiet Sunday morning", "18C at the park"

7. "condition_met": "yes" if the question context satisfies the condition, "no" if not
   Think carefully -- if the condition is "after 5pm" and the context is "3pm", it's "no"

8. "ground_truth_answer": a short yes/no answer with a brief reason
   e.g. "No -- it's only 3pm and Alex doesn't drink coffee before 5pm."
   e.g. "Yes -- it's raining, which is exactly when Jordan likes to cook elaborate meals."

Return strict JSON with key "rows", a list of objects:
- row_id (int)
- entity (string)
- entity_category (string: "person", "pet", or "character")
- behavior (string)
- condition_type (string, same as input)
- condition (string)
- entity_facts (list of exactly 1 string)
- question (string)
- question_context (string)
- condition_met (string: "yes" or "no")
- ground_truth_answer (string)

Rules:
1) entity_facts must have exactly 1 statement encoding both the behavior and the condition
2) The condition must be concrete and testable (not vague)
3) The question must present a specific context value that clearly either meets or doesn't meet the condition
4) condition_met must correctly reflect whether the question context satisfies the condition
5) ground_truth_answer must be consistent with condition_met
6) The question MUST be non-inferrable: without knowing the entity's specific rule, a
   reasonable person should be genuinely uncertain about the answer
7) Vary condition_met between "yes" and "no" across the batch
8) Do NOT reuse any entity names, behaviors, conditions, or phrasings from the examples above --
   they exist only to illustrate the format
9) Output ONLY valid JSON

Input specs:
{specs as JSON}
\end{lstlisting}
\end{promptbox}

\begin{promptbox}[Generation: \textit{Conditional-Facts} essay wrapper (Easy)]
\label{prompt:cond-essay-easy}
\begin{lstlisting}
For each item below, write a natural essay (7-10 sentences) about the entity
that embeds the conditional fact into a rich, casual narrative.

Rules:
1. The essay MUST preserve the conditional fact clearly -- both the behavior AND the
   condition must be present. Paraphrase is fine; do not omit either part.
2. All other sentences should describe the entity's background, personality, daily routines,
   relationships, hobbies, quirks, or life context. Every such sentence must be an
   unconditional, factual statement.
3. Do NOT introduce any new conditional statements anywhere in the essay. Forbidden
   constructions: "only when", "unless", "except when", "but only if", "whenever X then Y",
   "only after", "only if", or any other conditional phrasing beyond what was already in the
   original fact.
4. The essay should feel natural -- like an excerpt from a chat conversation, personal blog,
   or journal entry, not a formal report or list.
5. The conditional fact may appear anywhere in the essay, surrounded by unrelated context
   before and after it.
6. 5-8 sentences total.

Return strict JSON with key "rows", a list of:
  row_id (int, same as input), essay (string)

Output ONLY valid JSON.

Input:
{items as JSON}
\end{lstlisting}
\end{promptbox}

\begin{promptbox}[Generation: \textit{Conditional-Facts} essay wrapper (Hard)]
\label{prompt:cond-essay-hard}
\begin{lstlisting}
For each item below, write a natural essay (8-12 sentences) about the entity
that DECOMPOSES the original conditional fact into THREE distributed, non-adjacent sentences.
The reader must COMPOSE the rule from scattered evidence rather than copy it from a single
sentence.

CRITICAL RULE -- DO NOT write any single sentence that explicitly states both the behavior
AND the condition together. The original conditional fact must be split into exactly three
sentences spread across the essay:

  (A) BEHAVIOR sentence: describes the behavior as a tendency, habit, or pattern,
      WITHOUT naming the trigger condition.
  (B) CONDITION sentence: establishes the condition as part of the entity's life context
      or environment, WITHOUT naming the behavior.
  (C) LINK sentence (REQUIRED, not optional): subtly connects the two through timing,
      co-occurrence, or scene-setting -- using language like "It's usually...", "By then...",
      "Around that time...", "Most of the time it happens...", "That's typically when..." --
      but still WITHOUT explicitly stating "X happens when Y" or using any conditional phrasing.

The three sentences (A), (B), and (C) MUST all appear at non-adjacent positions in the
essay. Specifically: between any two of them there must be at least one sentence of
unrelated context. No two of (A), (B), (C) may be next to each other.

Examples of the distributed pattern (illustrative only -- do not reuse):

  Example 1
  Original fact: "Mara reorganizes her sketches when it's after midnight."
  Distributed essay:
    "Mara has lived in the same studio apartment for six years.
     She tends to reorganize her sketches in sudden bursts of focus.    <-- (A) BEHAVIOR
     Her cat Pepper is usually asleep on the windowsill by then.
     She works as a freelance illustrator and keeps irregular hours that stretch deep into the night.   <-- (B) CONDITION
     She drinks tea instead of coffee.
     It's almost always during those late hours that this kind of restless energy hits her.   <-- (C) LINK
     Her sister calls every Sunday afternoon."

  Example 2
  Original fact: "Devon sends long voice notes to his friends after a tough workout."
  Distributed essay:
    "Devon grew up in a small town outside Sacramento and still texts his old high school group chat daily.
     He has a habit of sending sprawling, ten-minute voice notes to his closest friends.   <-- (A) BEHAVIOR
     His apartment is decorated with secondhand furniture and a wall of climbing medals.
     Lately he has been pushing himself hard at the gym, leaving most sessions completely drained and shaky.   <-- (B) CONDITION
     He works remotely as a backend engineer and prefers afternoon meetings.
     Those long, rambling messages tend to come right after he stumbles home from the bouldering wall.   <-- (C) LINK
     His mom still mails him birthday cards a week early."

  Example 3
  Original fact: "Biscuit (a corgi) only eats from the red bowl when there are guests in the house."
  Distributed essay:
    "Biscuit is a five-year-old corgi who lives with the Tanaka family in Portland.
     He has a peculiar habit of eating exclusively from the red ceramic bowl on certain days.   <-- (A) BEHAVIOR
     He sleeps under the dining room table and follows the youngest kid everywhere.
     The Tanakas host frequent dinner parties, and the house is often full of unfamiliar voices and shoes by the door.   <-- (B) CONDITION
     His favorite toy is a chewed-up stuffed carrot.
     The red bowl tends to come out specifically on those crowded, noisy evenings.   <-- (C) LINK
     He gets groomed once a month at a place on Hawthorne."

Other rules:
1. The behavior and condition must BOTH be recoverable by a careful reader who composes
   sentences (A), (B), and (C) -- but NEITHER should appear in the same sentence.
2. Do NOT use explicit conditional phrasing anywhere ("only when", "whenever", "if",
   "unless", "except when", "but only if", "only after", "only if").
3. The link sentence (C) should use timing/scene language, not logical connectives.
4. All remaining sentences should describe the entity's background, personality, daily
   routines, relationships, hobbies, quirks, or life context -- unconditional factual statements.
5. The essay should feel natural -- like an excerpt from a personal blog or journal entry.
6. The correlation between the behavior and condition should be obvious to somebody who has read both sentences. It should NOT be vague or too subtle.
7. 8-12 sentences total.

Return strict JSON with key "rows", a list of:
  row_id (int, same as input), essay (string)

Output ONLY valid JSON.

Input:
{items as JSON}
\end{lstlisting}
\end{promptbox}

% ---------------------------------------------------------------------
\subsubsection{\textit{Coexisting-Facts}}

A single datapoint generator (Prompt~\ref{prompt:coexist-gen}) produces, for
each preference category, $N$ isolated first-person statements plus a holistic
scenario question whose answer requires all $N$.

\begin{promptbox}[Generation: \textit{Coexisting-Facts} datapoint generator]
\label{prompt:coexist-gen}
\begin{lstlisting}
Generate coexisting-facts datapoints. Each datapoint represents a user (first-person "I") with MULTIPLE preferences in the same category. Each preference will be stored as a completely SEPARATE, ISOLATED memory -- so each fact statement must make sense entirely on its own, with no reference to the other preferences.

For each spec below, generate:
1. "preferences": list of exactly num_preferences distinct preferences in the category
   (e.g. for foods: ["pizza", "sushi", "ramen"])

2. "preference_facts": list of exactly num_preferences short, natural first-person statements --
   ONE statement per preference, in the same order as "preferences".
   - Each statement must stand alone as a complete, self-contained fact
   - Each statement must mention ONLY that single preference (not the others)
   - Use varied, natural phrasing -- not a template ("I love X", "I enjoy X", "X is my favorite", etc.)
   - 1-2 sentences max per fact
   - Examples for foods:
       "I love pizza -- it's my default Friday night meal."
       "Sushi is my go-to whenever I want something fresh and light."
       "I'm a huge ramen fan, especially on cold days."

3. "question": a natural first-person scenario question that REQUIRES knowing ALL preferences.
   - Must NOT be a direct "list all my X" request -- make it a realistic scenario
   - Good: "I'm going grocery shopping -- what should I pick up for dinners this week?"
   - Good: "My friend wants to plan an outing I'd enjoy -- what are some solid options?"
   - The question should have a clearly better answer if ALL preferences are known vs. only one

4. "ground_truth_answer": a concise comma-separated list of all preference names
   Example: "pizza, sushi, ramen"

Return strict JSON with key "rows", a list of objects:
- row_id (int)
- preference_category (string, same as input)
- preferences (list of strings)
- preference_facts (list of strings, same length as preferences, one fact per preference)
- question (string)
- ground_truth_answer (string)

Rules:
1) preference_facts must have exactly the same length as preferences
2) Each fact covers exactly ONE preference and stands alone -- no cross-references
3) The question must be a realistic first-person scenario, NOT "list all my X"
4) Ground truth must include every preference, comma-separated
5) Output ONLY valid JSON

Input specs:
{specs as JSON}
\end{lstlisting}
\end{promptbox}

% ---------------------------------------------------------------------
\subsubsection{\textit{Persona-Retrieval}}

A single datapoint generator (Prompt~\ref{prompt:persona-gen}) jointly produces
the third-person essay about $E$ and the three first-person follow-up
questions, with each slot pre-marked as misleading-or-not by the calling code.

\begin{promptbox}[Generation: \textit{Persona-Retrieval} datapoint generator]
\label{prompt:persona-gen}
\begin{lstlisting}
Generate misleading-persona datapoints. Each datapoint is about a SPECIFIC named
person, embedded in a rich essay (10-15 sentences). Each datapoint has exactly 3 questions.
Each question is independently EITHER non-misleading (asks about the entity by name) OR
misleading (asks about a DIFFERENT named person -- the "distractor" -- who has no presence
in the essay).

For each spec below, generate:

1. "essay": a natural personal essay about the entity (10-15 sentences).
   - Written in third person, naming the entity (e.g. "Maya Patel"). Pronouns are fine
     after the first mention.
   - Embed MANY specific, memorable, idiosyncratic facts: daily rituals, unusual hobbies,
     hard constraints (allergies/aversions/rules), strong preferences, quirky possessions,
     rules of thumb. Aim for at least 4-5 distinct facts so different questions can probe
     different details.
   - Tone: casual, like a journal entry or chat message -- not a formal bio.
   - Do NOT use first-person voice ("I", "me", "my", "we").
   - Do NOT mention any of the distractor names anywhere in the essay.

2. "questions": a list of EXACTLY 3 question objects, in the order given by
   spec.question_slots. Each slot specifies whether that question is misleading and, if
   so, the distractor name to use.

   For each slot:

   If is_misleading=false:
     - "text": a first-person question that explicitly names the entity by their full name.
       The asker wants advice or info ABOUT the entity (e.g. what to get them, what to
       avoid, where to take them, whether they'd enjoy something).
     - The question must be answerable from a SPECIFIC detail in the essay -- NOT from
       generic norms.
     - Phrased naturally. Must NOT smuggle the answer into itself as an assumption.
       BAD:  "What apple dessert can I give Maya that won't make her itch?"
             (assumes the asker already knows about the allergy)
       GOOD: "What dessert should I make for Maya?"
             (open; the essay's allergy info is what makes the answer specific)
     - "ground_truth_answer": short answer (1-2 sentences) drawn from specific essay
       details. This is what a memory-aware system should return.
     - "distractor": null.

   If is_misleading=true:
     - "text": a first-person question that names the GIVEN distractor (NOT the entity).
       Same kind of advice/info question as the non-misleading case, but asked about the
       distractor. The entity's name must NOT appear in the question.
     - "ground_truth_answer": "I don't have information about <distractor name>."
       (Correct behavior is to abstain -- the asker has no info about this person.)
     - "distractor": the distractor name from the spec slot.

   The 3 questions should probe DIFFERENT angles -- don't repeat the same wording or topic
   across slots. If multiple slots are non-misleading, each should probe a different fact
   from the essay.

Return strict JSON with key "rows", a list of objects:
- row_id (int)
- entity (string, same as input)
- essay (string)
- questions (list of exactly 3 objects with fields: text, is_misleading, distractor,
  ground_truth_answer)

Rules:
1) The essay is 10-15 sentences, third-person, names the entity, and never mentions any
   distractor name from any slot.
2) Each non-misleading question names the entity exactly and never names any distractor.
3) Each misleading question names that slot's distractor exactly and never names the
   entity.
4) Non-misleading questions must NOT embed their own answers as assumptions.
5) Each non-misleading ground_truth_answer is supported by specific essay details.
6) Each misleading ground_truth_answer indicates the system should abstain.
7) Output ONLY valid JSON.

Input specs:
{specs as JSON}
\end{lstlisting}
\end{promptbox}

% ---------------------------------------------------------------------
\subsubsection{\textit{Long-Hop}}

Long-Hop generation runs in three phases: chain proposal
(Prompts~\ref{prompt:longhop-gen-system}--\ref{prompt:longhop-gen-user}),
cross-chain conflict / similarity audit
(Prompt~\ref{prompt:longhop-conflict}), and per-chain distractor generation
(Prompts~\ref{prompt:longhop-distract-system}--\ref{prompt:longhop-distract-user}).

\begin{promptbox}[Generation: \textit{Long-Hop} chain proposal --- system message]
\label{prompt:longhop-gen-system}
\begin{lstlisting}
You are constructing a benchmark of multi-hop reasoning chains.

Each chain consists of K+1 short factual statements that strictly link K+2
distinct anchors A -> B -> C -> ... in a transitive chain. Statement i must
relate anchor i to anchor i+1 (no other anchors, except the HEAD subject --
see rule 2 -- which may recur as background). The chain must support a single
multi-hop question: starting from A, answering the question requires chaining
through every statement to reach the terminal anchor.

Hard rules -- every chain must satisfy ALL of these:
1. EXACTLY K+1 statements per chain. Each statement is a single declarative
   English sentence, max ~16 words, no commas-separated multi-claims.
2. Statement i mentions anchor i and anchor i+1, plus an explicit relation
   word -- a verb ("loves", "hates", "always picks"), a conditional ("when",
   "whenever", "if"), a causal ("because", "leads to", "makes me"), a temporal
   ("after", "before"), or a preference ("I do X when Y"). MIDDLE and TERMINAL
   anchors (anchors 2 .. K+2) must appear ONLY in the two facts that border
   them -- they must not be named in any other fact.
   The HEAD subject is special and may recur as background in every fact:
     - First-person voice: "I/me/my" is the implicit speaker. The first-person
       speaker is NEVER literally listed in answer_chain -- anchor 1 is instead
       a state/action/object the speaker relates to ("eat apples", "bored").
     - Named-person voice: the chain's subject is a single named person
       ("Diego", "Marisol Vega"). That person IS anchor 1, and the same name
       must appear in every subsequent fact as background. Pronouns
       ("he/she/his/her") are FINE within a single fact when the proper
       noun also appears in that same fact (see rule 2b).
   Pick ONE voice per chain (first-person OR one named person) and stay
   consistent.
2b. EVERY STATEMENT MUST BE SELF-CONTAINED. A reader will encounter each
   statement in isolation, with no access to the other statements in the
   chain, so each statement must be fully interpretable on its own.

   Pronouns INSIDE a statement are FINE -- write natural English. The only
   rule is that every reference must resolve from the statement alone:
     - First-person ("I", "me", "my", "myself") is always fine -- the speaker
       is implicit and shared across the dataset.
     - Third-person pronouns ("he", "she", "it", "they", "them", "his",
       "her", "their", "its", etc.) are fine when their antecedent appears
       LITERALLY in the SAME statement. Examples that are GOOD:
         "Diego loves Korean food because he finds it spicy."   (he <-> Diego)
         "Marisol picks pop music whenever she is alone."       (she <-> Marisol)
         "Watering succulents keeps them healthy."              (them <-> succulents)

   What's forbidden is leaving a statement DEPENDENT on a different statement
   to resolve a reference. Do NOT:
     - Use a pronoun whose only possible antecedent appears in a DIFFERENT
       statement. If the proper noun isn't in this statement, repeat it.
     - Use referential phrases like "the company", "that town", "this book",
       "the same person" that depend on another statement to resolve.
3. K+2 anchors total per chain. Anchors should be SUBJECTIVE / PERSONAL content
   that cannot be looked up in an encyclopedia. Use anchors like:
     - States, moods, feelings ("bored", "anxious", "calm").
     - Actions, habits, routines ("eat apples", "skip lunch", "go for a run").
     - Preferences and opinions ("loves Korean food", "thinks pop music is
       overrated").
     - Concrete personal objects ("my bookshelf", "popcorn", "chamomile tea").
     - A single specific named person introduced as the chain's head subject
       ("Diego", "Marisol Vega") -- only when that person is anchor 1 and the
       rest of the chain is about their preferences / habits / moods.
   Do NOT use impersonal entities (companies, towns, rivers, books,
   institutions, geographic regions). First-person ("I ...") sentences are
   encouraged. Mix first-person and named-person styles freely.

3b. Every fact must be SUBJECTIVE: an opinion, preference, mood, routine,
   habit, or relational claim about a specific person (the speaker or a
   named person). Facts must NOT be impersonal world claims.
     GOOD:
       - "I think Italian food is overrated."           (first-person opinion)
       - "Diego loves Korean food."                     (fact about a person)
       - "Marisol always feels nervous before tests."   (personal trait)
       - "When I'm tired I get grumpy."                 (first-person routine)
     BAD:
       - "Drannot House published the novel."           (impersonal fact)
       - "Yepelmir lies in the province of Korunda."    (geographic claim)
       - "Strophien Atelier operates from Treskellin."  (impersonal corporate)
4. Within a single chain, all K+2 anchors must be distinct (case-insensitive).
5. Vary the relation patterns across the K+1 statements within one chain -- do
   not reuse the same conditional or verb template back-to-back.
6. The graded question must reference anchor 1 (the head) at least once by
   name and ask about the terminal anchor (the last in the chain), without
   ever naming any intermediate anchor. The question should read as a single
   natural English sentence and have a unique correct answer given the K+1
   statements. Natural pronouns are encouraged when they aid flow -- e.g.,
   "What does Diego do when he is bored?" or "When I'm dehydrated, what
   mood do I end up in?" -- provided every pronoun's antecedent is clear
   from the question itself.
7. ground_truth_answer must equal the terminal anchor exactly (or its shortest
   natural form -- e.g. drop a leading "the" only if the canonical phrase has
   no article).
8. Across chains in this batch, AVOID retelling the same narrative as anything
   in PRIOR CHAIN SUMMARIES (provided in the user message). Generic words like
   "sleep" or "bored" may repeat across chains, but a chain that paraphrases
   another chain's storyline must not be produced.

Distractor options are produced in a separate downstream step -- DO NOT include
any distractors / answer choices in your output here.

Output JSON only -- no commentary.
\end{lstlisting}
\end{promptbox}

\begin{promptbox}[Generation: \textit{Long-Hop} chain proposal --- user message (per batch)]
\label{prompt:longhop-gen-user}
\begin{lstlisting}
Produce a JSON object with exactly {batch_size} chains, each of HOP COUNT K = {hop_count}.

Each chain must follow ALL hard rules from the system prompt. To recap for K = {hop_count}:
- Exactly {K+1} statements.
- Exactly {K+2} distinct anchors, listed in answer_chain in chain order
  (head first, terminal last).
- DO NOT USE THE EXAMPLES I GIVE YOU IN YOUR CHAINS.
- Statement i links anchor i and anchor i+1. The HEAD subject (first-person
  "I" implicit, OR a named person at anchor 1) may also recur as background
  in every fact. Middle and terminal anchors must each appear ONLY in their
  two bordering facts.
- Every fact must be SUBJECTIVE -- an opinion / preference / mood / routine /
  habit, either first-person ("I") or about a single named person
  ("Diego", "Marisol"). No encyclopedic / entity-relation facts.
- Every anchor must appear LITERALLY (case-insensitive substring) in the
  fact(s) it belongs to.
- EVERY STATEMENT MUST BE SELF-CONTAINED -- a reader will see each fact in
  isolation, so each fact must be interpretable on its own. Pronouns INSIDE
  a single fact are fine when the antecedent is in the SAME fact ("Diego
  loves Korean food because he finds it spicy" -- "he" <-> Diego). What's
  forbidden is using a pronoun whose only antecedent appears in a DIFFERENT
  fact. No referential phrases ("the company", "that town") that depend on
  another fact to resolve. First-person ("I", "me", "my") is always fine.
  BAD:  "Lina buys a ticket and travels to the coast." / "Traveling to the
        coast means she visits Seabright."   <- "she" needs fact 1 to resolve.
  GOOD: "Lina buys a ticket and travels to the coast." / "Traveling to the
        coast means Lina visits Seabright."
  ALSO GOOD: "Lina buys a ticket because she loves travel."   <- "she" has
        antecedent "Lina" in the same fact.
- The graded_question references anchor 1 (the head) at least once by name
  and asks about the terminal anchor. The question must be unanswerable from
  any single statement alone. Natural pronouns referring to the head anchor
  are encouraged when they aid flow.
- ground_truth_answer is the canonical written form of the terminal anchor.
- DO NOT include any distractors / answer choices -- they are produced in a
  separate downstream step.

In-context examples (mix of styles -- produce a similar mix):

{four worked examples for the requested K, mixing one named-person opinion
chain with three first-person chains (causal / conditional / temporal /
preference)}

PRIOR CHAIN SUMMARIES (avoid retelling these storylines; pick fresh
narratives -- generic words like "sleep" or "tea" may repeat, but the chain
arc must be new):
{rolling list of one-line "head -> ... -> terminal" summaries from
previously accepted chains, capped at 30}

Output schema (JSON object):
{
  "chains": [
    {
      "facts": ["..."],
      "answer_chain": ["..."],
      "graded_question": "...",
      "ground_truth_answer": "..."
    },
    ...  // exactly {batch_size} chains
  ]
}

Generate now. Aim for diversity -- across the {batch_size} chains, vary the
sentence patterns (causal / conditional / temporal / preference / opinion)
and the topic domains (food, mood, weather, routine, work, hobbies, music,
travel, study, exercise, etc.). EVERY chain must be subjective: a first-
person chain (no named subject -- implicit "I") OR a chain about a single
named person ("Diego", "Marisol Vega", "Lina") and that person's
opinions / habits / moods. Mix the two voices freely across the batch. Do
NOT produce encyclopedic / entity-relation chains (no companies, towns,
rivers, books, geographic features).
\end{lstlisting}
\end{promptbox}

\begin{promptbox}[Generation: \textit{Long-Hop} cross-chain conflict / similarity audit]
\label{prompt:longhop-conflict}
\begin{lstlisting}
SYSTEM:
You are auditing a small set of made-up reasoning chains for cross-chain interference.

Two chains "interfere" if storing both into a single shared memory store would
corrupt reasoning for either chain. Interference occurs when ANY of the
following hold between facts in different chains:
- CONTRADICTION: the facts make incompatible claims about the same anchor
  (e.g., one chain says "I eat apples when bored", another says "I never eat
  apples").
- SHARED DISTINCTIVE ANCHOR: a distinctive proper-noun or distinctive composite
  phrase appears in two different chains (paraphrase or near-spelling counts).
  Generic single-word concepts ("sleep", "bored", "tea") repeating across
  chains is fine and does NOT count as interference.
- OVERTLY SIMILAR NARRATIVE: two chains tell essentially the same storyline
  with the same anchors in the same role -- e.g. "X is owned by Y" and
  "Y owns X", or "When I'm bored I sleep / sleep gives me a dream" appearing
  twice with only minor word swaps.

Return the list of chain IDs to DROP to eliminate all interference. When two
chains conflict, drop only ONE of them (your choice). Be precise; do not flag
chains that merely share generic concepts or common verbs.

Output JSON only.

USER:
CHAINS (each has an `id` and a list of `facts`):
{JSON array of {id, facts} for every survivor -- sent once globally and then
again over overlapping sliding windows of size batch_size}

Output JSON:
{
  "to_drop": ["<chain_id>", ...],
  "reason_per_drop": {"<chain_id>": "<one-line reason>", ...}
}
If there are no conflicts, return {"to_drop": [], "reason_per_drop": {}}.
\end{lstlisting}
\end{promptbox}

\begin{promptbox}[Generation: \textit{Long-Hop} distractor generator --- system message]
\label{prompt:longhop-distract-system}
\begin{lstlisting}
You write distractor options for a multi-hop reasoning multiple-choice question.

You receive:
- A list of FACTS forming a transitive reasoning chain (the graded answer
  requires chaining ALL facts).
- The GRADED_QUESTION (references only the head anchor; asks about the
  terminal anchor).
- The correct GROUND_TRUTH_ANSWER.

Produce EXACTLY 4 distractor options. Each distractor must satisfy ALL of the
following rules:

1. SAME-SHAPE PLAUSIBILITY. Match the correct answer in grammatical form,
   length range, and answer category. If the correct answer is a noun phrase
   naming a mood, every distractor is a noun phrase naming a mood. If the
   correct answer is a short verb phrase ("drink water"), every distractor is
   a short verb phrase of similar length and shape. Pronouns and articles
   that flow naturally with the question are fine -- match the voice the
   question uses (e.g., if the question is "What does Diego do when he is
   bored?", distractors phrased as "he reorganizes his closet" or simply
   "reorganizes the closet" are both acceptable, as long as the distractor
   reads as a fluent answer to the question).

2. REALISTIC AND ORDINARY. Each distractor must name something a real person
   could plausibly feel, do, prefer, eat, or experience in everyday life.
   NO absurd, surreal, slapstick, joke, or comically random options. NO
   things almost no one actually does (e.g., "duel a swan", "memorize country
   capitals from memory", "argue with neighbors about constellations"). Pick
   ordinary moods, habits, hobbies, foods, or activities -- the kind of
   answer a thoughtful peer might genuinely guess.

3. UNAMBIGUOUSLY WRONG. Must not be a paraphrase, synonym, sub-phrase,
   near-spelling, or otherwise overlapping with the correct answer or with
   any anchor / relation phrase that appears in any fact.

4. ORTHOGONAL TO EVERY FACT. A reader looking at any single fact in isolation
   must NOT be able to guess the distractor as a plausible "what comes next"
   or "natural consequence" via common-sense world knowledge. Avoid
   distractors that name typical effects, components, properties, or strong
   associations of any concept mentioned in any fact (e.g., if a fact
   mentions popcorn, do NOT pick a distractor about thirst, salt, or movies;
   if a fact mentions a cold shower, do NOT pick a distractor about feeling
   refreshed or shivery). Pick subject matter unrelated to every fact's
   topic.

5. DISTINCT. The four distractors must be distinct from each other
   (case-insensitive) and distinct from the correct answer.

Examples (note: realistic, ordinary, orthogonal):

CHAIN A facts:
  - "I eat apples when I'm bored."
  - "When I'm bored I go to sleep."
  - "When I sleep I have a dream."
  - "Every dream I have leaves me curious about the future."
GRADED_QUESTION: "What does eating apples eventually leave me feeling?"
CORRECT_ANSWER: "curious about the future"
GOOD distractors (ordinary moods/feelings, orthogonal to apples / sleep / dreams):
  - "nostalgic about old friendships"
  - "motivated to clean my apartment"
  - "indifferent toward upcoming holidays"
  - "satisfied with my routine"

CHAIN B facts:
  - "Marisol thinks pop music is overrated."
  - "Whenever pop music is on Marisol leaves the room."
  - "When Marisol leaves the room Marisol ends up in a sour mood."
GRADED_QUESTION: "What mood does Marisol end up in because of the music genre Marisol dislikes?"
CORRECT_ANSWER: "sour mood"
GOOD distractors (ordinary moods, no music / departure associations):
  - "a focused mood"
  - "a contemplative mood"
  - "a generous mood"
  - "a competitive mood"

Output JSON only -- no commentary.
\end{lstlisting}
\end{promptbox}

\begin{promptbox}[Generation: \textit{Long-Hop} distractor generator --- user message]
\label{prompt:longhop-distract-user}
\begin{lstlisting}
FACTS:
- {fact_1}
- {fact_2}
- ...

GRADED_QUESTION: {graded_question}

CORRECT_ANSWER: {ground_truth_answer}

Produce 4 distractor options that satisfy every rule from the system message. Output JSON:
{"incorrect_options": ["...", "...", "...", "..."]}
\end{lstlisting}
\end{promptbox}

% =====================================================================
\subsection{Evaluation prompts (sent to the memory-augmented model)}
\label{app:eval-prompts}

All four tasks share a single \texttt{ConversationHistoryPromptTemplate}
(\texttt{src/prompt\_templates.py}) that decides what is shown to the
evaluation model on each turn based on whether the turn is graded. The
ungraded form (Prompt~\ref{prompt:eval-ungraded}) is used during the storage
phase, where the model only needs to converse so the memory system can
absorb new facts; the graded form (Prompt~\ref{prompt:eval-graded}) is used
when an answer is being scored. \textit{Long-Hop} additionally wraps each
graded question with a strict-JSON MCQ instruction
(Prompt~\ref{prompt:longhop-mcq-wrapper}) so the chosen letter can be parsed
deterministically.

\begin{promptbox}[Evaluation: ungraded turn (storage phase)]
\label{prompt:eval-ungraded}
\begin{lstlisting}
You are a helpful chat assistant. Read the user's message carefully and remember any new personal information, preferences, or facts they share that you feel are important to remember.They may be recalled in future conversations.

Relevant Past Memories:
{memories}

Conversation History:
{history}

User: {query}
\end{lstlisting}
\end{promptbox}

\begin{promptbox}[Evaluation: graded turn]
\label{prompt:eval-graded}
\begin{lstlisting}
You are an intelligent memory assistant tasked with answering questions using information from past conversation memories.

# CONTEXT:
You have access to memories from previous conversations as well as the conversation history that may be helpful in answering the question.

# INSTRUCTIONS:
Answer the user's question. You may use the provided memories if they are helpful. If you use the memories above to answer a question, please EXPLICITLY RESTATE which memories you used below, or state that you used no memories. You should only use and restate those memories if you explicitly used them to draw conclusions from them.
Conversation History:
{history}
Relevant Memories:
{memories}
END of Relevant Memories

User Question:
{query}
Answer:
\end{lstlisting}
\end{promptbox}

\begin{promptbox}[Evaluation: \textit{Long-Hop} MCQ wrapper (appended to each graded question)]
\label{prompt:longhop-mcq-wrapper}
\begin{lstlisting}
{question_with_choices}

Choose exactly one option above. The correct answer is uniquely
determined by chaining the relevant remembered facts together; the
other four options cannot be inferred from those facts.

Respond with a single JSON object on its own line and nothing else,
using this exact schema:
  {"selected_choice": "<one of A, B, C, D, E>"}
\end{lstlisting}
\end{promptbox}

% =====================================================================
\subsection{Grading prompts (LLM judge)}
\label{app:grade-prompts}

Each task is graded by a staged LLM-judge pipeline that classifies failures
into the canonical taxonomy
(\textsf{not\_stored} / \textsf{summary\_error} / \textsf{not\_retrieved} /
\textsf{reasoning\_error} / \textsf{correct}). Every stage is a structured
JSON-schema call; the system message and the user template are reproduced
below. \textit{Long-Hop} grading omits the invocation stage --- the
$5$-way MCQ answer is parsed deterministically from the model's
\verb|{"selected_choice": "..."}| response.

% ---------------------------------------------------------------------
\subsubsection{Conditional-Facts grading}

Four sequential stages, short-circuiting on the first failure: storage,
summary, retrieval, invocation. The Hard variant replaces the storage,
summary, and retrieval prompts with rule-decomposed analogues that allow
the behavior and condition to be \emph{recovered across multiple memories}.

\begin{promptbox}[Grading: \textit{Conditional-Facts} Stage 1 --- Storage check (Easy)]
\label{prompt:cond-storage-easy}
\begin{lstlisting}
SYSTEM:
You are checking whether a specific fact was stored in a memory system.
The fact may have been paraphrased or compressed, but must still convey the same
information -- including any qualifying condition -- to count as present.

USER:
ORIGINAL FACT:
{original_fact}

ALL_MEMORIES (complete memory store):
{all_memories_formatted}

Is the original fact present in ALL_MEMORIES, even if paraphrased, as long as the
qualifying condition is preserved and the meaning is not altered?
\end{lstlisting}
\end{promptbox}

\begin{promptbox}[Grading: \textit{Conditional-Facts} Stage 1 --- Storage check (Hard)]
\label{prompt:cond-storage-hard}
\begin{lstlisting}
SYSTEM:
You are checking whether the COMPONENTS of a conditional rule -- a behavior and a
triggering condition -- were stored in a memory system at all.

This is a permissive presence check, not a faithfulness check. The two components may
appear in a single memory or split across multiple memories, may be paraphrased or
compressed, and may even be attached to the wrong entity or no entity at all -- that
quality concern is judged in a separate, later step. Your only job here is to determine
whether SOME version of the behavior AND SOME version of the condition exist somewhere
in the memory store.

The rule counts as PRESENT if both components -- the behavior and the triggering
condition -- can be located somewhere in ALL_MEMORIES, in any form (paraphrased,
distributed, or even mislabeled).

The rule counts as ABSENT only if one of the components is genuinely missing from the
entire store -- i.e., no memory mentions the behavior at all, OR no memory mentions the
triggering condition at all.

USER:
COMPONENTS TO LOOK FOR:
  Behavior:  {behavior}
  Condition: {condition}
  (For reference -- entity is "{entity}", but entity linkage is NOT required for this check.)

ALL_MEMORIES (complete memory store):
{all_memories_formatted}

Are BOTH components present somewhere in ALL_MEMORIES, even if paraphrased, even if
distributed across multiple memories, and even if attached to the wrong entity or no
entity? Do not penalize entity mislabeling here -- that is judged separately.
\end{lstlisting}
\end{promptbox}

\begin{promptbox}[Grading: \textit{Conditional-Facts} Stage 2 --- Summary check (Easy)]
\label{prompt:cond-summary-easy}
\begin{lstlisting}
SYSTEM:
You are assessing the quality of a stored memory.

The fact has already been confirmed to exist in the memory store. Your job is to evaluate
whether the stored version faithfully preserves the CRITICAL information -- specifically the
qualifying condition and the conditional behavior -- in a way that would support correct
downstream reasoning.

A stored version has a SUMMARY ERROR if ANY of the following apply:
- The qualifying condition was dropped entirely (stored as an unconditional fact)
- The condition was generalized in a way that changes the specific threshold or trigger
  (e.g., "after 5pm" -> "in the evening" loses precision; "when raining" -> "in bad weather" is too vague)
- The conditional relationship was inverted, confused, or made ambiguous
- Critical specifics (time, place, context, trigger) were lost or distorted such that
  a reader could not reliably determine whether a given scenario satisfies the rule

A stored version is FAITHFUL if the condition is clearly and specifically preserved and
a reader could correctly answer whether a given context satisfies the rule.

USER:
ORIGINAL FACT:
{original_fact}

ALL_MEMORIES (the fact IS confirmed present somewhere in here):
{all_memories_formatted}

Find the memory entry corresponding to this fact and assess whether the stored version
faithfully preserves the qualifying condition and behavior, or whether it has a summary error.
\end{lstlisting}
\end{promptbox}

\begin{promptbox}[Grading: \textit{Conditional-Facts} Stage 2 --- Summary check (Hard)]
\label{prompt:cond-summary-hard}
\begin{lstlisting}
SYSTEM:
You are assessing the quality of a stored conditional rule.

The rule has already been confirmed to be recoverable from the memory store (possibly
from a single memory, possibly composed across multiple memories). Your job is to
evaluate whether the recoverable version faithfully preserves the CRITICAL information --
specifically the qualifying condition and the conditional behavior -- in a way that would
support correct downstream reasoning about whether a given context satisfies the rule.

A stored version has a SUMMARY ERROR if ANY of the following apply:
- The qualifying condition was dropped entirely (the behavior is stored, but no condition
  for this entity can be recovered from any combination of memories).
- The condition was generalized in a way that changes the specific threshold or trigger
  (e.g., "after 5pm" -> "in the evening" loses precision; "when raining" -> "in bad weather"
  is too vague).
- The conditional relationship was inverted, confused, or made ambiguous.
- The behavior and the condition appear in the store but cannot be linked to THIS entity
  (e.g., the condition is attached to a different person, or stated as generic context
  with no clear tie back to the entity).
- Critical specifics (time, place, context, trigger) were lost or distorted such that a
  reader could not reliably determine whether a given scenario satisfies the rule.

The stored version is FAITHFUL if a reader, by reading the relevant memory or composing
across multiple memories, could correctly determine whether a given context satisfies the
conditional rule for this specific entity. Composition across memories is acceptable --
do NOT penalize the memory system simply for distributing the behavior and condition
across separate entries.

USER:
COMPOSED CONDITIONAL RULE:
  Entity:    {entity}
  Behavior:  {behavior}
  Condition: {condition}
  In words:  "{entity} {behavior} {condition}." (paraphrase is fine)

ALL_MEMORIES (the rule IS confirmed recoverable somewhere in here, possibly across
multiple entries):
{all_memories_formatted}

Find the memory entry -- or the set of memory entries -- that together encode this rule
for this entity, and assess whether the recoverable version faithfully preserves the
qualifying condition and behavior, or whether it has a summary error.
\end{lstlisting}
\end{promptbox}

\begin{promptbox}[Grading: \textit{Conditional-Facts} Stage 3 --- Retrieval check (Easy)]
\label{prompt:cond-retrieval-easy}
\begin{lstlisting}
SYSTEM:
You are checking whether a specific fact was included in the memories retrieved and
shown to an AI model when it answered a question.

USER:
ORIGINAL QUESTION (used to search the memory store):
{query}

ORIGINAL FACT:
{original_fact}

RETRIEVED_MEMORIES (shown to the model):
{retrieved_memories}

Was the original fact included in RETRIEVED_MEMORIES? It is ok if the fact was
paraphrased or partially rephrased, as long as it was not changed in a way that
omits or alters information necessary to answer the query.
\end{lstlisting}
\end{promptbox}

\begin{promptbox}[Grading: \textit{Conditional-Facts} Stage 3 --- Retrieval check (Hard)]
\label{prompt:cond-retrieval-hard}
\begin{lstlisting}
SYSTEM:
(Same system message as Stage 3 Easy.)

USER:
ORIGINAL QUESTION (used to search the memory store):
{query}

COMPOSED CONDITIONAL RULE TO LOOK FOR:
  Entity:    {entity}
  Behavior:  {behavior}
  Condition: {condition}
  In words:  "{entity} {behavior} {condition}." (paraphrase is fine)

RETRIEVED_MEMORIES (shown to the model):
{retrieved_memories}

Is the composed conditional rule recoverable from RETRIEVED_MEMORIES -- either from a
single retrieved entry or by composing across multiple retrieved entries -- even if
paraphrased? The rule counts as recovered if both the behavior and the condition (linked
to this entity) are present in the retrieved set in some form. The rule counts as NOT
recovered if either the behavior or the condition is missing from what was shown to the
model, or if the link to this entity cannot be made out.
\end{lstlisting}
\end{promptbox}

\begin{promptbox}[Grading: \textit{Conditional-Facts} Stage 4 --- Invocation check]
\label{prompt:cond-invocation}
\begin{lstlisting}
SYSTEM:
You are checking whether an AI model correctly used a retrieved memory to answer
a conditional-fact question.

The model was shown the relevant memory and must answer YES if the condition is met,
or NO while citing the condition if it is not met.

USER:
ORIGINAL FACT (was present in retrieved memories):
{original_fact}

GROUND TRUTH ANSWER:
{ground_truth_answer}

MODEL RESPONSE:
{llm_response}

Did the model correctly invoke the memory and arrive at the right answer?
\end{lstlisting}
\end{promptbox}

% ---------------------------------------------------------------------
\subsubsection{Coexisting-Facts grading}

The first three stages run \emph{per-fact} (one judge call per coexisting
preference), independently classifying each preference as \textsf{not\_stored}
/ \textsf{summary\_error} / \textsf{not\_retrieved} / \textsf{correct}. The
final invocation check (Prompt~\ref{prompt:coexist-invocation}) is run only
when all $N$ facts reach \textsf{correct} and asks whether the model's
response covers all of them.

\begin{promptbox}[Grading: \textit{Coexisting-Facts} Stage 1 --- Per-fact storage check]
\label{prompt:coexist-storage}
\begin{lstlisting}
SYSTEM:
You are checking whether a specific preference was stored in a memory system.
The preference may have been paraphrased or compressed, but must still be clearly identifiable.

USER:
ORIGINAL PREFERENCE: {preference}
ORIGINAL FACT: {original_fact}

ALL_MEMORIES (complete memory store):
{all_memories_formatted}

Is this preference present in ALL_MEMORIES, even if paraphrased or lightly compressed?
It counts as present as long as the specific preference can still be clearly identified.
\end{lstlisting}
\end{promptbox}

\begin{promptbox}[Grading: \textit{Coexisting-Facts} Stage 2 --- Per-fact summary check]
\label{prompt:coexist-summary}
\begin{lstlisting}
SYSTEM:
You are assessing the quality of a stored preference.

The preference has already been confirmed to exist in the memory store. Your job is to
evaluate whether the stored version faithfully preserves the specific identity of this
preference in a way that would support correct downstream reasoning.

A stored version has a SUMMARY ERROR if ANY of the following apply:
- The preference was overgeneralized or merged with others, losing its distinct identity
  (e.g., "sushi" -> "Asian food"; "jazz" -> "music")
- The preference was corrupted or replaced with something different
- Critical identifying details were lost such that the model could not specifically cite
  this preference when answering a question

A stored version is FAITHFUL if the specific preference can still be clearly and
unambiguously identified from the stored memory.

USER:
ORIGINAL PREFERENCE: {preference}
ORIGINAL FACT: {original_fact}

ALL_MEMORIES (the preference IS confirmed present somewhere in here):
{all_memories_formatted}

Find the memory entry for this preference and assess whether the stored version faithfully
preserves the specific preference identity, or whether it has a summary error.
\end{lstlisting}
\end{promptbox}

\begin{promptbox}[Grading: \textit{Coexisting-Facts} Stage 3 --- Per-fact retrieval check]
\label{prompt:coexist-retrieval}
\begin{lstlisting}
SYSTEM:
You are checking whether a specific preference was included in the
memories retrieved and shown to an AI model when it answered a question.

USER:
ORIGINAL PREFERENCE: {preference}
ORIGINAL FACT: {original_fact}

ORIGINAL QUESTION (used to search the memory store):
{query}

RETRIEVED_MEMORIES (shown to the model):
{retrieved_memories}

Was this preference included in RETRIEVED_MEMORIES, even if paraphrased?
It counts as retrieved as long as the specific preference can be clearly identified.
\end{lstlisting}
\end{promptbox}

\begin{promptbox}[Grading: \textit{Coexisting-Facts} Stage 4 --- Invocation check (when all facts correct)]
\label{prompt:coexist-invocation}
\begin{lstlisting}
SYSTEM:
You are checking whether an AI model correctly used all retrieved preferences
to answer a question.

All expected preferences were present in the retrieved memories. The model should
have mentioned all of them in its response.

USER:
EXPECTED_PREFERENCES (all were present in retrieved memories):
{expected_preferences}

GROUND TRUTH ANSWER:
{ground_truth_answer}

MODEL RESPONSE:
{llm_response}

Did the model correctly mention or account for ALL expected preferences in its response?
Synonyms and paraphrases count (e.g. "pasta dishes" covers "spaghetti").
\end{lstlisting}
\end{promptbox}

% ---------------------------------------------------------------------
\subsubsection{Persona-Retrieval grading (batched)}

\textit{Persona-Retrieval} grading is \emph{batched}: $20$ traces share a
single judge call per stage so the (large) memory store is sent only once
per batch. Stage~1--3 are shared across direct and misleading queries; the
final stage branches on the question type --- direct queries are graded by
the invocation check (Prompt~\ref{prompt:persona-invocation}) and misleading
queries by the abstention check (Prompt~\ref{prompt:persona-abstention}).
All five system messages and all five user templates are listed below.

\begin{promptbox}[Grading: \textit{Persona-Retrieval} system messages (5 stages)]
\label{prompt:persona-systems}
\begin{lstlisting}
STORAGE_SYSTEM:
You are checking whether personal essays about named entities are stored in a
memory system. Each essay may have been paraphrased, compressed, or split across multiple
memory entries, but the entity's identity (name) and the substantive details from the
essay must still be recoverable to count as present.

You will receive ALL_MEMORIES (one shared store) and a list of independent ENTRIES, each
identified by an integer id. For each entry, return one verdict in the "results" array,
preserving the id.

SUMMARY_SYSTEM:
You are assessing the quality of stored personal essays.

Each essay has already been confirmed to exist somewhere in the memory store. Your job
is to evaluate whether the stored version preserves IDENTITY and KEY DETAILS faithfully --
in a way that would support correct downstream reasoning.

A stored version has a SUMMARY ERROR if ANY of the following apply:
- The entity's name was dropped or changed, so the stored details cannot be tied back
  to this specific person
- Specific, idiosyncratic details (allergies, possessions, rituals, hard constraints,
  strong preferences) were lost or generalized to the point that a reader could not
  recover the original meaning
- The stored version is so vague or generic that it would not distinguish this entity
  from a generic person with the same role/hobby

The following are NOT summary errors -- treat them as acceptable:
- Multiple distinct entries existing under the same name. The entity may share a name
  with other people described in the store. As long as the entity's own essay is
  recoverable from at least one entry (or set of entries), this is fine -- do not flag
  it as a merged-identity error.
- The store containing memories about other people, whether related to the entity or
  unrelated. Only the entity's own essay needs to be faithfully preserved; the
  presence of additional people is acceptable.
- The essay being split across multiple entries, paraphrased, or compressed, as long
  as the entity's identity and key memorable details are recoverable.

A stored version is FAITHFUL if the entity is clearly named (in at least one entry)
and the key memorable details from the essay can be recovered from the relevant
memory or memories tied to that entity.

You will receive ALL_MEMORIES (one shared store) and a list of ENTRIES, each
identified by an integer id. For each entry, return one verdict in the "results"
array, preserving the id.

RETRIEVAL_SYSTEM:
You are checking whether each entity's essay was included in the memories retrieved
and shown to an AI model when it answered a question. Each entry has its own
RETRIEVED_MEMORIES (the subset shown for that question). Presence requires that the
entity be identifiable AND that the substantive personal details from the essay be
recoverable from what was shown to the model. Return one verdict per entry under
"results", preserving the id.

INVOCATION_SYSTEM:
You are checking whether an AI model correctly used a retrieved essay to answer a
question about the named entity. The model was shown the relevant memory and must
answer using the entity-specific details from the essay (not generic norms). Return
one verdict per entry in "results", preserving the id.

ABSTENTION_SYSTEM:
You are checking whether an AI model correctly ABSTAINED when asked about a person
the model has no information on.

Each entry's question names a DISTRACTOR -- a different person who is NOT the entity in
the retrieved essay. The correct behavior is to acknowledge no information is available
about the distractor (or to ask for clarification). Applying the entity's essay details
to the distractor -- by treating them as if they were the entity, or by silently using
the entity's specifics to answer about the distractor -- is INCORRECT. Note: the model
may legitimately mention information about other people stored in memory if it does not
attribute the entity's essay details to the distractor. Return one verdict per entry in
"results", preserving the id.
\end{lstlisting}
\end{promptbox}

\begin{promptbox}[Grading: \textit{Persona-Retrieval} Stage 1 --- Storage user template]
\label{prompt:persona-storage}
\begin{lstlisting}
ALL_MEMORIES (complete memory store, shared by all entries):
{all_memories_formatted}

For each entry below, determine whether the entity's essay is present in
ALL_MEMORIES, even if paraphrased, compressed, or split across multiple entries.
Presence requires that the entity be identifiable AND that the substantive
personal details from the essay be recoverable.

ENTRIES:
{entries}

Return one verdict per entry under "results", preserving the id (1..{n}).

(Each entry block has the form
   [id=k]
   ENTITY: {entity}
   ESSAY: {essay}
)
\end{lstlisting}
\end{promptbox}

\begin{promptbox}[Grading: \textit{Persona-Retrieval} Stage 2 --- Summary user template]
\label{prompt:persona-summary}
\begin{lstlisting}
ALL_MEMORIES (complete memory store, shared by all entries):
{all_memories_formatted}

For each entry below, the essay IS confirmed present somewhere in ALL_MEMORIES.
Find the memory entry -- or set of entries -- that correspond to the entity's essay
and assess whether the stored version preserves identity and the key details, or
whether it has a summary error.

Reminder: it is acceptable for the store to contain other entries under the same
name (multiple personas) or memories about other people, related or otherwise.
Only judge whether each entity's own essay is faithfully recoverable from the
entries that describe it.

ENTRIES:
{entries}

Return one verdict per entry under "results", preserving the id (1..{n}).
\end{lstlisting}
\end{promptbox}

\begin{promptbox}[Grading: \textit{Persona-Retrieval} Stage 3 --- Retrieval user template]
\label{prompt:persona-retrieval}
\begin{lstlisting}
For each entry, decide whether the entity's essay was included in that entry's
RETRIEVED_MEMORIES, even if paraphrased or split across multiple entries. Presence
requires that the entity be identifiable AND that the substantive personal details
from the essay be recoverable from what was shown to the model.

ENTRIES:
{entries}

Return one verdict per entry under "results", preserving the id (1..{n}).

(Each entry block has the form
   [id=k]
   ENTITY: {entity}
   QUESTION: {question}
   ESSAY: {essay}
   RETRIEVED_MEMORIES:
   {retrieved_memories}
)
\end{lstlisting}
\end{promptbox}

\begin{promptbox}[Grading: \textit{Persona-Retrieval} Stage 4 --- Invocation user template (direct queries)]
\label{prompt:persona-invocation}
\begin{lstlisting}
For each entry, decide whether the model correctly used the essay's specific
details to answer the question. The answer must align with the GROUND_TRUTH and
reflect the essay's specifics, not generic information.

ENTRIES:
{entries}

Return one verdict per entry under "results", preserving the id (1..{n}).

(Each entry block has the form
   [id=k]
   ENTITY: {entity}
   QUESTION: {question}
   ESSAY: {essay}
   GROUND_TRUTH: {ground_truth_answer}
   MODEL_RESPONSE: {llm_response}
)
\end{lstlisting}
\end{promptbox}

\begin{promptbox}[Grading: \textit{Persona-Retrieval} Stage 4 --- Abstention user template (misleading queries)]
\label{prompt:persona-abstention}
\begin{lstlisting}
For each entry, decide whether the model correctly abstained -- acknowledged it has
no information about the DISTRACTOR -- rather than applying the ENTITY's essay
details to the DISTRACTOR. The model abstains correctly even if it generally
answers using common knowledge, as long as it does NOT smuggle in
ENTITY-specific details. Conflating the two people, or using the essay as if it
were about the distractor, is an abstention failure.

ENTRIES:
{entries}

Return one verdict per entry under "results", preserving the id (1..{n}).

(Each entry block has the form
   [id=k]
   ENTITY (essay is about): {entity}
   DISTRACTOR (named in question, NOT the entity): {distractor}
   ESSAY (about the entity): {essay}
   QUESTION (about the distractor): {question}
   MODEL_RESPONSE: {llm_response}
)
\end{lstlisting}
\end{promptbox}

% ---------------------------------------------------------------------
\subsubsection{Long-Hop grading}

\textit{Long-Hop} grading runs three stages \emph{per supporting fact}
(storage, summary, retrieval) and then determines invocation correctness by
deterministic letter parsing of the model's MCQ answer rather than via an
LLM-judge call. Memory systems may legitimately merge several chain-links
into a single memory entry; each per-fact stage explicitly accepts merged
entries as long as the specific link is still recoverable.

\begin{promptbox}[Grading: \textit{Long-Hop} Stage 1 --- Per-fact storage check]
\label{prompt:longhop-storage}
\begin{lstlisting}
SYSTEM:
You are checking whether a specific factual statement (one link in a multi-hop reasoning chain) is preserved in a memory store.

The memory store may have stored the fact verbatim, paraphrased it, OR merged
several chain-links together into one combined memory. Any of these forms
counts as STORED -- as long as the SPECIFIC link asserted by the target fact
can still be unambiguously identified from at least one memory entry. If the
link's two specific entities and the relation between them are recoverable,
mark fact_in_store=true.

USER:
TARGET FACT (one link of a reasoning chain):
{target_message}

ALL_MEMORIES (complete memory store, possibly with merged entries):
{all_memories_formatted}

Is the target fact present in ALL_MEMORIES -- verbatim, paraphrased, or as part
of a merged memory entry that still preserves the specific link between the
target fact's two entities? Mark fact_in_store=true only if the precise
relationship between the two specific entities is unambiguously recoverable.
\end{lstlisting}
\end{promptbox}

\begin{promptbox}[Grading: \textit{Long-Hop} Stage 2 --- Per-fact summary check]
\label{prompt:longhop-summary}
\begin{lstlisting}
SYSTEM:
You are assessing the quality of a stored chain-link.

The link has already been confirmed to exist somewhere in the memory store
(possibly inside a merged memory entry). Your job is to evaluate whether the
stored version faithfully preserves the specific link in a way that would
support correct downstream chain reasoning.

A stored version has a SUMMARY ERROR if ANY of the following apply:
- The link's two entities were collapsed/renamed/swapped, breaking identity.
- The relation between them was corrupted, weakened, or replaced.
- The link was over-merged with unrelated facts so that the specific link can
  no longer be cleanly extracted (e.g. the entities are listed but not in a
  way that preserves which-relates-to-which).
- A critical detail (e.g. direction of the relation) was lost.

A stored version is FAITHFUL if both entities of the link are clearly named in
some memory entry and the specific relation between them is unambiguous --
EVEN IF the memory entry also contains other chain-links from the same chain
(merging is allowed when each individual link is still recoverable).

USER:
TARGET FACT (one link of a reasoning chain):
{target_message}

ALL_MEMORIES (the link IS confirmed present somewhere in here, possibly merged):
{all_memories_formatted}

Find the memory entry (or entries) that cover this link and assess whether the
stored version faithfully preserves the specific relation between the two
entities, or whether it has a summary error.
\end{lstlisting}
\end{promptbox}

\begin{promptbox}[Grading: \textit{Long-Hop} Stage 3 --- Per-fact retrieval check]
\label{prompt:longhop-retrieval}
\begin{lstlisting}
SYSTEM:
You are checking whether a specific chain-link was included in the memories
retrieved and shown to an AI model when it answered a multi-hop question.

The retrieved memories may include verbatim, paraphrased, or merged versions.
A merged memory that still contains the link's specific relation counts as
"retrieved".

USER:
TARGET FACT (one link of a reasoning chain):
{target_message}

ORIGINAL QUESTION (used to query the memory store):
{query}

RETRIEVED_MEMORIES (shown to the model when it answered):
{retrieved_memories}

Was the target fact's specific link included in RETRIEVED_MEMORIES, even if
paraphrased or merged with other chain-links? Mark fact_in_retrieved=true only
if the precise relationship between the link's two entities is unambiguously
recoverable from the retrieved memories.
\end{lstlisting}
\end{promptbox}

\paragraph{Invocation (deterministic, no LLM call).}
After the three per-fact stages succeed for every fact in a chain, the
invocation outcome is read directly from the model's MCQ response. The
parser tries, in order: (i) parsing the entire response as a JSON object
and reading \verb|selected_choice| / \verb|answer| / \verb|choice|;
(ii) hunting for any embedded JSON object inside the response and trying
the same keys; (iii) falling back to the last lone uppercase letter
\texttt{A}--\texttt{E} that appears in the response. The parsed letter is
compared against \texttt{correct\_choice} from the dataset row to decide
\textsf{correct} vs.\ \textsf{reasoning\_error}.

\end{document}